\documentclass{article} 
\usepackage{iclr2027_conference,times}

\usepackage{amsmath,amsfonts,bm}

\def\eqref#1{equation~\ref{#1}}

\def\1{\bm{1}}

\DeclareMathAlphabet{\mathsfit}{\encodingdefault}{\sfdefault}{m}{sl}
\SetMathAlphabet{\mathsfit}{bold}{\encodingdefault}{\sfdefault}{bx}{n}

\usepackage{hyperref,url,amsthm,amssymb,graphicx,booktabs,array,tabularx}
\usepackage{algorithm,algorithmic}
\newcolumntype{L}[1]{>{\raggedright\arraybackslash}p{#1}}
\graphicspath{{figures/}}

\theoremstyle{plain}
\newtheorem{theorem}{Theorem}[section]
\newtheorem{lemma}[theorem]{Lemma}
\newtheorem{proposition}[theorem]{Proposition}
\newtheorem{corollary}[theorem]{Corollary}

\theoremstyle{definition}
\newtheorem{definition}[theorem]{Definition}
\newtheorem{assumption}[theorem]{Assumption}
\theoremstyle{remark}
\newtheorem{remark}[theorem]{Remark}

\title{Provable Transformer Context Compression for In-Context Reinforcement Learning}

\author{%
  \begin{minipage}{0.95\textwidth}
    \centering
    \normalfont\itshape
    Yajie Zhao$^{1, 2}$, Yingjun Mo$^{1}$, Zhiyong Wang$^{3}$, Bo Tang$^{1}$, Chao Xu$^{2}$, Fang Kong$^{1,*}$\\[0.7em]
    {\normalfont\small $^{1}$ Southern University of Science and Technology}\\
    {\normalfont\small $^{2}$ University of Electronic Science and Technology of China}\\
    {\normalfont\small $^{3}$ Harbin Institute of Technology}%
  \end{minipage}%
}

\hypersetup{
  pdftitle={Provable Transformer Context Compression for In-Context Reinforcement Learning},
  pdfauthor={Yajie Zhao, Yingjun Mo, Zhiyong Wang, Bo Tang, Chao Xu, Fang Kong}
}

\iclrfinalcopy

\begin{document}
\maketitle
\begingroup
\renewcommand{\thefootnote}{*}
\footnotetext{Corresponding author: \texttt{kongf@sustech.edu.cn}.}
\endgroup
\fancyhead{}
\renewcommand{\headrulewidth}{0pt}

\begin{abstract}
Transformers can perform in-context reinforcement learning without parameter updates. However, existing theoretical constructions often require context sizes that grow with the number of sampled transitions, leading to increasing memory costs in large-scale problems. This raises a fundamental scalability question: how much context is intrinsically necessary to carry out such learning? We answer this question for fixed-policy TD evaluation. We show that, with a shared linear decoder, the minimum context dimension required to exactly reproduce the first $L$ TD iterates is the Krylov dimension induced by the sampled transitions. When the available context dimension $m$ falls below this threshold, exact reproduction is no longer possible. In this regime, we characterize the optimal approximation error and show that it decays as $\Theta(\gamma^m\max\{1,1/(m(1-\gamma))\})$ where $\gamma$ is the discounting factor. Moving from representation complexity to actual memory cost, we characterize the optimal retained memory for exact continuous representations as $\Theta(nk_L)$\footnote{$k_L$ here denotes Krylov dimension induced by the sampled transitions} and show experimentally that on a sampled task whose exact rank is full but whose numerical rank is small, a context accurate to $10^{-2}$ is $88.7\%$ smaller than the sparse sufficient statistics of TD. Finally, we establish the learnability of the shared executor through pretraining by proving global convergence of an auxiliary objective over effective attention parameters under suitable task diversity. Experiments across diverse TD evaluation problems validate parameter recovery and the predicted context--accuracy tradeoff.
\end{abstract}

\section{Introduction}
Reinforcement learning (RL) studies how an agent learns to make decisions through interaction with an environment to maximize cumulative reward \citep{sutton2018reinforcement}. In conventional RL, this adaptation typically involves updating the parameters of a policy or value network from collected experience. More recently, with the growing ability of sequence models to use long interaction histories, in-context reinforcement learning (ICRL) has emerged as a different mode of adaptation: a pretrained agent can improve on a new task by conditioning on its interaction history while keeping its model parameters fixed \citep{moeini2025survey}. In-context RL has demonstrated strong empirical success across a range of domains including navigation, continuous-control tasks, and open-ended embodied environments \citep{duan2016rl2,bauer2023humantimescale,lu2023structured,grigsby2024amago}. 

The empirical success naturally raises the question of why ICRL can adapt to unseen tasks using context. Recent theoretical work interprets ICRL as the Transformer forward pass executing an implicit RL algorithm, and can be divided into \emph{supervised pretraining} and \emph{reinforcement pretraining}. Under supervised pretraining, theoretical results show that Transformers can implement a range of classical RL algorithms \citep{laskin2023algorithm,lee2023supervised,lin2024transformersdecisionmakers,liang2026transformerspolicyimprovement,chen2025optimaldynamicregret}. Under reinforcement pretraining, both linear and softmax Transformers can provably implement TD-style updates, with the corresponding computation emerging from policy-evaluation pretraining \citep{wang2025transformerstdd,wang2025towardsprovable,xie2026beyondlinear,xie2026convergencecot}.

However, identifying the algorithm implemented in context does not reveal how much information must be retained to reproduce its update process. Existing in-context RL theory typically retains the full interaction history in the prompt. In several representative theoretical constructions, states or state--action pairs along a trajectory are represented using either one-hot embeddings in tabular settings \citep{lin2024transformersdecisionmakers,wang2025towardsprovable}, or more general $d$-dimensional feature maps, such as $\phi:\mathcal S\to\mathbb R^d$ \mbox{\citep{wang2025transformerstdd,liang2026transformerspolicyimprovement,xie2026beyondlinear}}. Consequently, for a trajectory of $N$ transitions $\{(s_t,a_t,r_t,s_{t+1})\}_{t=1}^{N}$, the task-specific memory of these constructions scales as $\Theta(dN)$. Such linear growth with the interaction horizon can become a substantial bottleneck in long-horizon or high-dimensional RL problems, as storing and processing an ever-growing context incurs increasing inference-time memory and computational costs. Existing constructions therefore establish that retaining the full interaction history is sufficient, but leave open whether it is necessary. Taking TD learning for policy evaluation as a canonical example, this raises two natural questions: 

\noindent\textbf{(Q1)} \emph{Necessary context requirement:}
What is the minimum context dimension, and the corresponding memory requirement, needed to exactly reproduce the TD iterates?

\noindent\textbf{(Q2)} \emph{Approximation with limited context:}
When the available context dimension falls below this threshold, what is the best achievable approximation error? 

To answer these questions, we consider a trajectory $\mathcal D_N=\{(s_t,a_t,r_t,s_{t+1})\}_{t=1}^{N}$ collected under a fixed policy, and $n$ is the state number. Then we study the exact context required to preserve these iterates and the approximation error when the available context is limited. Our main contributions are as follows.

\noindent\textbf{1. Context complexity.}
We characterize the intrinsic context dimension required for in-context TD learning. With a shared linear decoder, we show that the minimum context dimension required to exactly reproduce $L$ empirical TD iterates is the Krylov dimension $k_L$~\citep{saad2003iterative}. When the available context dimension $m$  falls below this threshold, we further characterize the best achievable approximation and establish matching upper and lower error bounds of order $\Theta(\gamma^m\max\{1,1/(m(1-\gamma))\})$.

\noindent\textbf{2. Memory complexity.}
We distinguish context dimension, peak memory, and retained task memory. For representations with encoders continuous in empirical model, we characterize the minimum retained memory required to answer all state--depth TD queries exactly as $\Theta(nk_L)$ real registers\footnote{A real register denotes one storage location that can hold a real number exactly.}. On a sampled task of full exact rank but small numerical rank, a context accurate to $10^{-2}$ retains \textbf{88.7\%} less memory than the sparse TD statistics.
We further show that the optimal retained memory for answering all state--depth queries within error $\epsilon$ is
$\Theta_\gamma\!\bigl(n\min\{n,L,\log(1/\epsilon)\}\log(1/\epsilon)\bigr)$ bits.

\noindent\textbf{3. Learnability and experiment.}
Finally, we study whether the compressed TD computation can be learned through pretraining.  
Under suitable task diversity, we show that an auxiliary objective is globally strongly convex in the effective attention parameters and provide provable global convergence to the target TD update. Extensive experiments further validate parameter recovery, the predicted context--accuracy tradeoff, and substantial memory savings.

\begin{table}[t]
\caption{Constructions that execute TD policy evaluation in the forward pass. $N,n,d,L$ denote transitions, states, feature dimension, and TD depth; $k_L=\dim\mathcal K_L(\gamma\widehat P,\widehat r)$. Theory counts task scalars; buffers are float32 array bytes$/1024$ over 100 encodings at $N=2000$, $n=d=24$, $L=30$, $k_L=12$, excluding parameters and forward scratch. Constructions for other in-context RL problems \citep{lin2024transformersdecisionmakers,chen2025optimaldynamicregret,liang2026transformerspolicyimprovement} retain different objects and are listed in Appendix~\ref{app:other-constructions} as not comparable.}
\label{tab:context-comparison}
\label{tab:context-memory-comparison}
\centering
\setlength{\tabcolsep}{4pt}
\renewcommand{\arraystretch}{1.0}
\footnotesize
\begin{tabularx}{\linewidth}{@{}
>{\raggedright\arraybackslash}p{0.19\linewidth}
>{\raggedright\arraybackslash}X
>{\raggedright\arraybackslash}X
>{\centering\arraybackslash}p{0.13\linewidth}
>{\raggedleft\arraybackslash}p{0.09\linewidth}
@{}}
\toprule
\textbf{Work} & \textbf{Problem setting} & \textbf{Retained object} & \textbf{Retained memory} & \textbf{Buffer (KiB)}\\
\midrule
\citet{wang2025transformerstdd} & TD(0), linear features, sampled trajectory & $N$ transition tokens $(\phi_t,\gamma\phi_{t+1},R_{t+1})$ & $\Theta(Nd)$ & 383.00\\
\citet{wang2025towardsprovable} & tabular TD, known dynamics & $n$ state tokens with rows of $\gamma P$ and $r$ & $\Theta(n^2)$ & 4.79\\
\citet{xie2026beyondlinear} & softmax (kernel) TD, sampled trajectory & $N$ state tokens $(x_t,R_{t+1})$ & $\Theta(Nd)$ & 211.04\\
\citet{xie2026convergencecot} & TD(0) with chain of thought, linear features & $N$ transition tokens and $L$ iterate tokens & $\Theta((N+L)d)$ & 579.15\\
\midrule
\textbf{Ours} & tabular TD (Thm.~\ref{thm:main}) or linear features (Thm.~\ref{thm:features}), sampled trajectory & $k_L$ Krylov tokens $(B,Y,g)$ & $\mathbf{O(nk_L+n)}$ & $\mathbf{2.49}$\\
\bottomrule
\end{tabularx}
\end{table}

To the best of our knowledge, we are the first to characterize the context and memory requirements for preserving an in-context RL computation, as well as the approximation error below the exact-representation threshold. We require recovery across all TD depths, rather than only the terminal output, because our goal is to reproduce the algorithmic computation itself rather than merely its final result. Unlike classical compression methods, we realize the compression end-to-end within a fixed-parameter linear-attention network operating directly on raw transitions. Table~\ref{tab:context-comparison} compares our work with existing ICRL constructions.

\section{Related Work}
\label{sec:related_work}

\paragraph{Theoretical Mechanisms of In-Context Reinforcement Learning.}
Recent theory studies ICRL by viewing the Transformer forward pass as implementing an RL algorithm. Under supervised pretraining, \citet{lin2024transformersdecisionmakers} construct Transformers that approximate classical online decision-making algorithms, while \citet{liang2026transformerspolicyimprovement} extend this perspective to SARSA and actor--critic, and \citet{chen2025optimaldynamicregret} analyze strategies for non-stationary RL. Without explicitly supervising the internal update rule, \citet{wang2025transformerstdd} show that linear-attention Transformers can implement TD updates, and \citet{wang2025towardsprovable} characterize parameters realizing such computations as global minimizers of a pretraining objective. \citet{xie2026beyondlinear} extend the mechanism to standard softmax attention, whereas \citet{xie2026convergencecot} interpret chain-of-thought generation as repeated TD computation.

\paragraph{Reinforcement Learning and Traditional Compression.}
Reinforcement learning studies sequential decision making through interaction with an environment, with classical methods including temporal-difference learning, Q-learning, and policy-gradient methods \citep{sutton1988td,watkins1992qlearning,williams1992reinforce}.
For TD policy evaluation, classical compression replaces raw trajectories with sufficient statistics, as in LSTD and iLSTD \citep{bradtke1996lstd,boyan2002lstd,geramifard2006ilstd,katharopoulos2020linear}, and Krylov subspaces also arise in value approximation and model reduction\citep{parr2007bebf,parr2008models,petrik2007analysis}.
Classical model-reduction methods provide a variety of low-dimensional representations for dynamical and Markov systems \citep{ho1966realization,freund2003krylov,antoulas2005approximation,kemeny1976chains,givan2003equivalence}.
Unlike classical compression methods, we characterize the context needed for TD evaluation and realize the compression end-to-end within a fixed-parameter linear-attention network operating directly on raw transitions.


\paragraph{Context Compression.}
Context compression has been widely studied to reduce the computational and memory costs of models. Existing approaches either compress long contexts into compact learned representations \citep{rae2020compressive,mu2023gist,chevalier2023autocompressors,ge2024icae,zhang2025activationbeacon}, or reduce the input through token/segment selection and retrieval compression \citep{li2023selective,jiang2023llmlingua,jiang2024longllmlingua,pan2024llmlingua2,xu2024recomp}. In contrast, we characterize the memory--accuracy limits of in-context TD compression and realize them directly from raw transitions with fixed-parameter attention. 

\section{Preliminaries}
\label{sec:prelim}

\paragraph{Reinforcement Learning Setting.}We consider an infinite-horizon discounted Markov decision process (MDP) with finite\footnote{The analysis can be extended beyond finite state spaces when states admit a finite-dimensional representation $\phi:\mathcal S\to\mathbb R^d$, in which case the corresponding constructions are carried out in the feature space.} state space $\mathcal S=[n]$, action space $\mathcal A$, transition kernel $P_{\mathrm{MDP}}(s' \mid s,a)$, reward function $r_{\mathrm{MDP}}(s,a)$, and discount factor $\gamma\in [0,1)$\citep{puterman1994markov}. For a fixed policy $\pi$, the MDP induces a Markov reward process (MRP)
with transition matrix $P^\pi\in\mathbb R^{n\times n}$ and expected one-step reward vector $r^\pi\in\mathbb R^n$, where $P^\pi_{ij} := \sum_{a\in\mathcal A} \pi(a\mid i)P_{\mathrm{MDP}}(j\mid i,a)$ and $r^\pi_i :=\sum_{a\in\mathcal A}\pi(a\mid i)r_{\mathrm{MDP}}(i,a)$. The policy-evaluation problem is to estimate the value function $v^\pi(i) := \mathbb E_\pi \left[\sum_{t=0}^{\infty}\gamma^t R_{t+1}\,\middle|\, S_0=i\right]$. Temporal-difference (TD) learning estimates this value function by bootstrapping from consecutive transitions~\citep{sutton1988td,sutton2018reinforcement}. In general, a single TD(0) update takes the form $v_{t+1} = v_t + \alpha_t \bigl( R_{t+1} +\gamma v_t(S_{t+1}) -v_t(S_t) \bigr)\nabla_{v_t} v_t(S_t)$. In the remainder of the paper, we therefore work directly with the induced MRP and focus on the corresponding TD-learning dynamics for fixed-policy evaluation. Let $\mathcal D_N=\{(S_t,A_t,R_{t+1},S_{t+1})\}_{t=0}^{N-1}$ contain $N$ transitions collected under $\pi$. We define $N_i:=\sum_{t=0}^{N-1}\mathbf 1\{S_t=i\}$ and $N_{ij}:=\sum_{t=0}^{N-1}\mathbf 1\{S_t=i,S_{t+1}=j\}$, then assume \textbf{full state coverage}, i.e., $N_i>0$ for all $i\in[n]$. Appendix~\ref{app:exact-context} discusses the rationale for this assumption. The empirical transition and reward estimates are $\widehat P_{ij} := \frac{N_{ij}}{N_i}$ and  $\widehat r_i := \frac{1}{N_i}\sum_{t:S_t=i}R_{t+1}$. For a value vector $v\in\mathbb R^n$, define the empirical TD error $\widehat\delta_t(v) := R_{t+1}+\gamma v(S_{t+1})-v(S_t)$ and the empirical state-visitation matrix $\widehat D_N := \operatorname{diag} \left(\frac{N_1}{N},\ldots,\frac{N_n}{N}\right)$. Starting from $\widehat v_0=0$, let $\widehat v_\ell\in\mathbb R^n$ denote the empirical value estimate after $\ell$ iterations of visitation-normalized batch TD(0)~\citep{yao2008preconditioned}:
\begin{equation}
    \widehat v_{\ell+1}
    =
    \widehat v_\ell
    +
    \widehat D_N^{-1}
    \frac1N
    \sum_{t=0}^{N-1}
    \widehat\delta_t(\widehat v_\ell)e_{S_t}
    =
    \sum_{j=0}^{\ell}
    (\gamma \widehat P)^j\widehat r.
\label{eq:empirical-td-sample}
\end{equation}
Therefore, letting $\widehat A:=\gamma\widehat P$, reproducing the first $L$ TD iterates amounts to preserving the successive directions
$\widehat r,\widehat A\widehat r,\ldots,
\widehat A^{L-1}\widehat r$. This motivates the Krylov space
\citep{saad2003iterative}
\[
    \mathcal K_L(\widehat A,\widehat r)
    :=
    \operatorname{span}
    \{
        \widehat r,
        \widehat A\widehat r,
        \ldots,
        \widehat A^{L-1}\widehat r
    \},
    \qquad
    k_L
    :=
    \dim\mathcal K_L(\widehat A,\widehat r).
\]

\section{Context and Memory Complexity of In-Context TD}
\label{sec:memory-theory}
We study the representation and memory required to reproduce the empirical TD dynamics in (\ref{eq:empirical-td-sample}), and the approximation error when the available context falls below the exact requirement. Our construction is end-to-end: an \emph{encoder} compresses the raw transitions $\mathcal D_N$ into a task-specific context within the forward pass, and an \emph{executor} operates on this context to reproduce the queried TD values without parameter updates; linear attention is the canonical realization, and softmax attention is treated in Appendix~\ref{app:softmax}.

\paragraph{Context representation.}
Let $z_{\ell,i}\in\mathbb R^{d_{\mathrm{model}}}$ denote the $i$th active source token after $\ell$ TD updates, and $Z_\ell := [\,z_{\ell,1},\ldots,z_{\ell,m}\,] \in\mathbb R^{d_{\mathrm{model}}\times m}$ denote the query-independent source context. For a state query $q$, let $z_{\ell,\star}(q)$ be a masked query token and define $\overline Z_\ell(q):=[\,Z_\ell,z_{\ell,\star}(q)\,]$. Note that the query token is not counted among the $m$ active sources.
\begin{definition}[$m$-dimensional context representation]
\label{def:context}
We call $\{Z_\ell\}_{\ell=0}^{L-1}$ an $m$-\textbf{dimensional context representation} of a value trajectory $\{\widetilde v_\ell\}_{\ell=0}^{L}$ if there exist a shared linear decoder $D\in\mathbb R^{n\times m}$ and a fixed readout $\rho\in\mathbb R^{d_{\mathrm{model}}}$ with $s_\ell := Z_\ell^\top\rho \in \mathbb R^m$ and
\[
    \widetilde v_\ell
    =
    D\sum_{j=0}^{\ell-1}s_j,
    \qquad 0\le \ell\le L.
\]
The representation is \emph{exact} if $\widetilde v_\ell=\widehat v_\ell$ for every $\ell\le L$. We write $\mathfrak C_m$ for the set of trajectories that admit an $m$-dimensional context representation.
\end{definition}
The $m$ active source tokens carry the reduced TD increments $s_\ell$, and the shared decoder $D$ maps them back to the state space where the value is $\widetilde v_\ell(q)=e_q^\top D\sum_{j<\ell}s_j$. Context dimension is measured in tokens rather than scalar values and does not directly reflect storage or memory requirements. We therefore analyze the memory costs. For a stored object $X$, let $\operatorname{Loc}(X)$ be the set of distinct scalar locations occupied by $X$. Aliased locations are counted once and stored copies separately; both quantities count real registers and exclude shared parameters.
\begin{definition}[Retained and peak task memory]
\label{def:memory}
For one query $q$, the \textbf{retained task memory} is
$M_{\rm ret}:=\bigl|\operatorname{Loc}(\overline Z_\ell(q))\cup\operatorname{Loc}(D)\bigr|$,
the number of locations that must persist after encoding in order to execute the compressed TD computation. Let $M_{\rm enc}$ be the largest number of distinct locations occupied during encoding, and let $W(\ell)$ be the temporary workspace of execution layer $\ell$. The \textbf{peak task memory} is
$M_{\rm peak}:=\max\bigl\{M_{\rm enc},\;\max_{0\le\ell<L}|\operatorname{Loc}(\overline Z_\ell(q))\cup\operatorname{Loc}(D)\cup\operatorname{Loc}(W(\ell))|\bigr\}$.
\end{definition}

\subsection{End-to-End Construction: Encoder and Executor}
\label{sec:exact-context}
The iterates (\ref{eq:empirical-td-sample}) depend on the data only through $(\widehat A,\widehat r)$ and their increments lie in $\mathcal K_L$, which suggests compressing in two stages, $N\to n\to k_L$: both are carried out by attention heads with fixed parameters directly on the raw records, so compression is part of the forward pass.

\paragraph{Selector heads.}
During encoding, let $U=[u_1,\ldots,u_M]\in\mathbb R^{d_{\rm model}\times M}$ collect the active tokens, each partitioned into named coordinate blocks with selectors $E_a$, so that $E_au_i=u_i[a]$. An encoder head is a residual linear-attention update \citep{katharopoulos2020linear,vonoswald2023transformers}
\begin{equation}
 U^+ = U+ PU\bigl[\mathcal R\odot(U^\top Q U)\bigr],
 \qquad
 P=\sigma E_o^\top E_v,\quad Q=E_k^\top E_q,\quad \sigma\in\{\pm1\},
 \label{eq:encoder-head}
\end{equation}
which adds $\sigma\sum_i u_i[v]\,(u_i[k]^\top u_t[q])$ to block $o$ of token $t$ over the source tokens $i$ allowed by the fixed routing mask $\mathcal R\in\{0,1\}^{M\times M}$: \(Q\) selects the key and query blocks that form the score, \(P\) selects the source value block and writes it with sign \(\sigma\), into the output block, and \(\mathcal R\) specifies which source--target pairs may interact. We also allow $\operatorname{recip}_0(x)=1/x$ for $x>0$, $\operatorname{recip}_0(0)=0$, and the release of tokens at stage boundaries.

\paragraph{Stage 1: raw records to statistics ($N\to n$).}
Raw token \(u_t^{\mathrm{raw}}\) carries $(e_{S_t},e_{S_{t+1}},R_{t+1},1)$, and state token $u_i^{\mathrm{state}}$ carries $e_i$ with zero-initialized statistic blocks. Routing every raw token to every state token with the matching score $e_{S_t}^\top e_i=\mathbf 1\{S_t=i\}$, three heads accumulate $N_{i,:}$, $\sum_{t:S_t=i}R_{t+1}$, and $N_i$ in state token $i$; $\operatorname{recip}_0(N_i)$ and the input $\gamma$ then form $\widehat P_{i,:}$, $\widehat r_i$, and $\widehat A_{i,:}$ by self-routed heads. The raw tokens are released, and the $n$ state tokens carry $(\widehat A,\widehat r)$ (Appendix~\ref{app:stage1}).

\paragraph{Stage 2: from $n$ states to $k_L$ Krylov directions ($n\to k_L$).}
The state-level context already reproduces TD exactly with the executor below (Appendix~\ref{app:state-level}), but it uses $n$ tokens although the first $L$ increments span only the corresponding Krylov space $\mathcal K_L$. Let $h=\min\{n,L\}$. Starting from $b_1=\widehat r$ in a work token, fixed heads run $h$ rounds of an unnormalized Arnoldi recurrence in primal--dual form \citep{saad2003iterative}, i.e.,
\begin{equation}
 d_j=b_j^\top b_j,\qquad \omega_j=\operatorname{recip}_0(d_j)\,b_j,\qquad \eta_j=d_j\operatorname{recip}_0(d_j),\qquad
 b_{j+1}=\widehat Ab_j-\textstyle\sum_{i\le j}b_i\,\omega_i^\top\widehat Ab_j ,
 \label{eq:arnoldi}
\end{equation}
The nonzero $b_j$ are orthogonal and span $\mathcal K_L$, and the gate $\eta_j$ is one exactly on them, so token release keeps exactly $k_L$ candidates. With $B=[b_1,\ldots,b_{k_L}]$ and $\Omega=[\omega_1,\ldots,\omega_{k_L}]$, further heads compile
\begin{equation}
 H=\Omega^\top\widehat AB,\qquad g=\Omega^\top\widehat r,\qquad Y=B+\Omega(H-I_{k_L})^\top,\qquad
 Z_0=[\,B;\ Y;\ g^\top\,]\in\mathbb R^{(2n+1)\times k_L}.
 \label{eq:krylov-context}
\end{equation}
Here $B$ provides coordinates in the Krylov subspace, $H$ and $g$ are the reduced operator and reward, and $Y$ stores $H$ inside width-$n$ blocks so that the executor can read it through inner products.

\paragraph{Executor.}
Write the context at depth $\ell$ as $Z_\ell=[B;\,Y;\,s_\ell^\top]$ with $s_0=g$, and the masked query token as $z_{\ell,\star}(q)=(e_q,0_n,c_\ell)$ with $c_0=0$. Execution uses one linear-attention layer at every depth \citep{yang2024looped},
\begin{equation}
 \overline Z_{\ell+1}
 =
 \overline Z_\ell
 +
 P_{\rm TD}\overline Z_\ell M_{k_L}\bigl(\overline Z_\ell^\top Q_{\rm TD}\overline Z_\ell\bigr),
 \qquad
 P_{\rm TD}=\operatorname{diag}(0_{2n},1),\quad
 Q_{\rm TD}=\left[\begin{smallmatrix}-I_n&I_n&0\\0&0&0\\0&0&0\end{smallmatrix}\right],
 \label{eq:executor}
\end{equation}
with the mask $M_{k_L}=\operatorname{diag}(I_{k_L},0)$ letting only the $k_L$ source tokens act as sources. The static blocks $B$, $Y$ never change, the last row evolves as $s_{\ell+1}=Hs_\ell$ and $c_{\ell+1}=c_\ell-e_q^\top Bs_\ell$. Hence $-c_\ell(q)=e_q^\top B\sum_{j<\ell}H^jg=\widehat v_\ell(q)$ by Krylov moment matching, and $\{Z_\ell\}$ is an exact $k_L$-dimensional context representation with $D=B$ and $s_\ell=H^\ell g$.

\begin{theorem}
\label{thm:main}
Let the empirical TD system be covered and $L\ge1$, we have:
\textbf{(a) Context dimension.}
The minimum dimension of an exact context representation is the corresponding Krylov dimension $k_L$.
\textbf{(b) Construction.}
For fixed $n$ and $L$, a task-independent encoder maps every covered raw transitions in its forward pass, to the context~(\ref{eq:krylov-context}) with exactly
$k_L$ tokens, and the executor~(\ref{eq:executor}) exactly outputs
$\widehat v_\ell(q)$ for all $q$ and $\ell$.
\textbf{(c) Retained memory.}
The construction uses $O(nk_L)$ retained task memory and is optimal up to a factor of two among continuous exact encoders, even with arbitrary decoders.
Hence the optimal retained memory is $\Theta(nk_L)$.
\end{theorem}

Part (a) is the exact context requirement: every increment $\widehat A^j\widehat r=\widetilde v_{j+1}-\widetilde v_j=Ds_j$ lies in $\operatorname{col}(D)$, so $m\ge\operatorname{rank}D\ge k_L$, and (\ref{eq:krylov-context}) attains it. Part (b) is the constructive content: the $k_L$ tokens, the decoder, and the reduced operator are produced by a fixed attention network in its forward pass, so compression is not a preprocessing step. What's more, we can get the corresponding upper bounds $M_{\rm ret}=(2n+1)(k_L+1)$ and $M_{\rm peak}=O(Nn+n^2+nh)$ with $h=\min\{n,L\}$.

For Part (c), fix $1\le k\le\min\{n-1,L-1\}$. On any open set\footnote{Here ''open'' means relatively open in the $k_L=k$ model class, i.e., open with respect to the subspace topology induced by $(\widehat A,\widehat r)$.} of covered
systems with $k_L=k$ containing a model with
$\mathbf 1\notin\mathcal K_L$, every encoder continuous in
$(\widehat A,\widehat r)$ and exact for all state--depth queries must retain
at least $nk+k$ real registers, even with an arbitrary nonlinear decoder
(Appendix~\ref{app:memory-proof}); our context retains $(2n+1)k$ for the source tokens, within a factor two, plus one query token. The bound is class-level: on a restricted family with fewer degrees of freedom, such as the block-cycle tasks of Section~\ref{sec:experiments} where $\mathbf 1\in\mathcal K_L$, a family-specific encoder can retain far less; the theorem controls task-uniform encoders such as ours. If $k_L=n$ and $L\ge n+1$, at least $n^2$ registers are necessary, so the statistics are optimal up to $(n+1)/n$. For offline coding with arbitrary encoders and decoders, the optimal memory for all state--depth queries within error $\epsilon$ is $\Theta_\gamma(n\min\{n,L,\log(1/\epsilon)\}\log(1/\epsilon))$ bits (Appendix~\ref{app:bits}). The same context is also executed exactly by one looped layer of standard two-head softmax attention (Appendix~\ref{app:softmax}), and a single-depth query needs $\Theta(\log L)$ layers (Appendix~\ref{app:depth}).

\paragraph{Linear features and existing in-context TD constructions.}
Existing constructions run TD with linear features on $N$ transition tokens and expose the corresponding TD value in every iterate, one per layer \citep{wang2025transformerstdd} or per generated token \citep{xie2026convergencecot}, and a looped network can be read out at any loop \citep{yang2024looped}. Our goal is to replace the original $O(N)$-token contexts with our $k_L$-token contexts and allow a fixed executor to reproduce the TD computations. Let $\phi:\mathcal S\to\mathbb R^d$, $\phi_t=\phi(S_t)$, $\phi'_t=\phi(S_{t+1})$, and consider batch TD(0) in sum form with a public step size \(\alpha\), the update of \citet{wang2025transformerstdd} with $C_\ell=\alpha I$ and their $1/N$ absorbed into $\alpha$:
\begin{equation}
 A_N=\textstyle\sum_t\phi_t(\phi_t-\gamma\phi'_t)^\top,\quad b_N=\sum_t\phi_tR_{t+1},\quad \theta_0=0 \qquad
 \theta_{\ell+1}=\theta_\ell+\alpha\,(b_N-A_N\theta_\ell)
 .
 \label{eq:lfa-td}
\end{equation}
With $M:=I_d-\alpha A_N$ and $s_0:=\alpha b_N$, the iterates are $\theta_\ell=\sum_{j<\ell}M^js_0$, so their increments span $\mathcal K_L(M,s_0)\subseteq\mathbb R^d$. Let $k^\phi_L:=\dim\mathcal K_L(M,s_0)$, and let a context representation of $\{\theta_\ell\}$ have a decoder $D\in\mathbb R^{d\times m}$ with answer $\phi(q)^\top D\sum_{j<\ell}s_j$.
\begin{theorem}[Feature-based TD]
\label{thm:features}
\textbf{(a)} The minimum dimension of an exact context representation of $(\theta_\ell)_{\ell\le L}$ is $k^\phi_L$.
\textbf{(b)} For fixed $d$, $L$ and public $\alpha,\gamma$, a fixed-parameter encoder whose first stage uses no reciprocal (its second stage is that of Theorem~\ref{thm:main}) maps every batch of feature transitions in its forward pass to a context with exactly $k^\phi_L$ tokens, and the executor (\ref{eq:executor}) outputs $\phi(q)^\top\theta_\ell$ for all $q$ and $\ell\le L$. Moreover, $M_{\rm ret}=(2d+1)(k^\phi_L+1)$, $M_{\rm peak}=O(Nd+d^2+dh)$.
\textbf{(c) Retained memory.}
For each $1\le k\le\min\{d,L\}$, any encoder continuous
in $(M,s_0)$ and exact for all inputs with $k_L^\phi=k$
requires at least $dk$ real registers, even with arbitrary
decoders. Hence the optimal retained memory is
$\Theta(dk_L^\phi)$.
\end{theorem}
Stage~1 forms $A_N,b_N$ by two heads (key $\phi_t$, query $e_i$) and needs no $\operatorname{recip}_0$ because (\ref{eq:lfa-td}) is un-normalized, unlike the visitation-normalized tabular statistics. Stage~2, which does use $\operatorname{recip}_0$ as in Section~\ref{sec:exact-context}, and the executor run on $(M,s_0)$, so the context has at most $d$ tokens against $N$ (Appendix~\ref{app:features}). Thus, Theorem~\ref{thm:features} extends the Krylov-context construction and memory guarantees from one-hot state representations to general linear features under the unnormalized batch TD update~(\ref{eq:lfa-td}).

\subsection{Approximation with Limited Context}
\label{sec:approx-context}
When $m<k_L$, exact reproduction through depth $L$ is impossible, and we ask how accurately an $m$-dimensional context can approximate the TD trajectory. With $\|x\|_\mu^2=\sum_i\mu_ix_i^2$, the normalized worst-case error of the depth-$L$ problem of Definition~\ref{def:context} is
\[
\mathfrak E_{m,L}(\gamma):=\sup_{(\widehat P,\widehat r,\mu)}\inf_{\widetilde v\in\mathfrak C_m}\max_{0\le\ell\le L}\|\widehat v_\ell-\widetilde v_\ell\|_\mu/\|\widehat r\|_\mu,
\]
where the outer supremum is over covered empirical systems of any size with a strictly positive stationary distribution $\mu$ and $\widehat r\ne0$, and $\mathfrak C_m$ is the class of Definition~\ref{def:context}. $\mathfrak E_m(\gamma)$ is the depth-free version ($L=\infty$, $\sup_{\ell\ge0}$), which dominates $\mathfrak E_{m,L}(\gamma)$ for every $L$. The law below is stated for $\mathfrak E_m$, and the finite-depth case follows it. We seek to characterize the optimal worst-case error jointly in context dimension $m$ and discount factor $\gamma$, and to construct a context that attains it.
Classical Krylov approximation \citep{freund2003krylov,liesen2013krylov}
does not: it spends the whole budget on the first $m$
increments, so its worst-case error is exactly
$\gamma^m/(1-\gamma)$ (Proposition~\ref{prop:mm-suboptimal}),
which below the horizon barely decays with $m$.

Instead of preserving only the first $m$ TD increments, we
approximate the TD trajectory by storing its values at selected depths and using the most recently stored value between samples.
To choose these depths, let $J=\lfloor 1/(1-\gamma)\rfloor$,
which reflects how slowly the discounted TD increments decay.
We divide the $m$-token budget into two parts, $m=k+j$:
the first $k$ tokens preserve the initial $k$ TD increments
exactly, while the remaining $j$ tokens capture changes at
later depths. Specifically, we set
\[
(k,j)=
\begin{cases}
(m,0), & \gamma<\frac12,\\
(0,m), & \gamma\ge\frac12,\ m\le J,\\
(m-J,J), & \gamma\ge\frac12,\ m>J.
\end{cases}
\]
The first $k$ sampling depths are $1,\ldots,k$. For the
remaining $j$ tokens, we select later depths $k+i_s$, where $i_s=\min\left\{i\ge0:
1-\gamma^i\ge\frac{s}{j+1}\right\}$ for $s \in [j]$.
This approximately divides the remaining discounted weight into equal portions, allowing
larger gaps between samples at later depths. The complete
set of sampled depths is therefore
\[
\mathcal T_m(\gamma)
=\{1,\ldots,k\}\cup\{k+i_s:1\le s\le j\}.
\]
Write the selected depths in increasing order as $\tau_1<\cdots<\tau_m$, with $\tau_0=0$. Each token $u$ stores the difference between two consecutive
sampled iterates, i.e., $x_u=\widehat v_{\tau_u}-\widehat v_{\tau_{u-1}}$. These differences allow the sampled iterates to be recovered by accumulation.
A fixed five-head encoder constructs the tokens by recording
the TD trajectory at the selected depths
(Appendix~\ref{app:sh-encoder}).

To answer a query at depth $\ell$, we use the query accumulator of the looped executor~(\ref{eq:executor}),
but update it only at the selected sampling depths.
A counter tracks the current depth, and the gate $s_{\ell,u}=\mathbf 1\{\ell+1=\tau_u\}$ determines when token $u$ contributes $e_q^\top x_u$ to the accumulator.
Between sampling depths, the output remains unchanged. Consequently,
\[
\widetilde v_\ell(q)
=\sum_{\tau_u\le\ell}e_q^\top x_u
=e_q^\top\widehat v_{\sigma(\ell)},
\qquad
\sigma(\ell)
=\max\bigl(\{0\}\cup\{\tau_u:\tau_u\le\ell\}\bigr),
\]
where $\sigma(\ell)$ is the most recent sampled depth
not exceeding $\ell$. With $D=[x_1,\ldots,x_m]$,
this gives an $m$-dimensional context representation.

\begin{theorem}

\label{thm:gamma-law}
For every $m\ge1$ and $\gamma\in(0,1)$, with $\Psi_m(\gamma)=\gamma^m\max\{1,\tfrac{1}{m(1-\gamma)}\}$,
\[
 \tfrac1{64}\,\Psi_m(\gamma)\ \le\ \mathfrak E_m(\gamma)\ \le\ e^2\,\Psi_m(\gamma).
\]
The upper bound is attained on every covered system by the sample-and-hold context of dimension $m$, consequently $\mathfrak E_m(\gamma)=\Theta\bigl(\gamma^m\max\{1,1/(m(1-\gamma))\}\bigr)$.
\end{theorem}

For \(m<1/(1-\gamma)\), any \(m\)-dimensional context has worst-case error at least \(\Psi_m/64\asymp 1/[m(1-\gamma)]\), regardless of how it is encoded. Thus, in this range, increasing the context dimension cannot guarantee geometric decay in error; the lower bound decreases only as \(1/m\). Since tabular systems are a special case of feature-based systems with \(\phi(s)=e_s\), the same lower bound applies to feature-based contexts. However, Krylov truncation~\citep{freund2003krylov,liesen2013krylov} incurs a multiplicative gap relative to the minimax upper bound, quantified by $\min\{m,1/(1-\gamma)\}/e^2$. The proof of this gap is given in Appendix~\ref{app:mm-proof}, while the minimax upper and lower bounds are proved in Appendix~\ref{app:gamma-law-proof}.

\textbf{Finite depth.}
Definition~\ref{def:context} considers a fixed number \(L\) of iterations. For every \(L\), the optimal error satisfies \(\mathfrak E_{m,L}\le\mathfrak E_m\le e^2\Psi_m\), and it is zero exactly when \(m\ge k_L\). The constructions used to prove the lower bound require at most \(m+4/(1-\gamma)\) iterations. Thus, when \(L\ge m+4/(1-\gamma)\), the same lower bound applies: \(\mathfrak E_{m,L}\ge\Psi_m/64\). For smaller \(L\), the error bounds also depend on the number of iterations. By choosing the sampled depths to divide the total discounted weight over the first \(L\) iterations into approximately equal parts, sample-and-hold achieves error at most \(e^2\gamma^m\max\{1,(1-\gamma^L)/(m(1-\gamma))\}\). In particular, when \(8m\le L\le1/(1-\gamma)\), the optimal error is \(\Theta(L/m)\). The proof is given in Appendix~\ref{app:finite-depth}.

\subsection{Learnability of the TD Update}
\label{sec:learnability}
The construction above uses the fixed parameters $(P_{\mathrm{TD}},Q_{\mathrm{TD}})$ defined in~\eqref{eq:executor} to execute TD on the compressed context, and we now ask whether pretraining recovers them with the encoder kept fixed. 
First, we should note that $(P_{\mathrm{TD}},Q_{\mathrm{TD}})$ is not unique: linear attention is invariant to the reciprocal rescaling, i.e.,
\[
PZM(Z^\top QZ)=(aP)ZM(Z^\top(a^{-1}Q)Z),\quad a\ne0.
\]
So the correct parameters form an equivalence class rather than a point. To remove this ambiguity we fix $P=P_{\mathrm{TD}}$ and parameterize the trainable query--key interaction as $Q_{\mathrm{eff}}(C,W):=\left[\begin{smallmatrix}C&W^\top&0\\0&0&0\\0&0&0\end{smallmatrix}\right]$ with $C,W\in\mathbb R^{n\times n}$, whose target is $C^\star=-I_n$ and $W^\star=I_n$, for which $Q_{\mathrm{eff}}(C^\star,W^\star)=Q_{\mathrm{TD}}$.

Let \(\tau\) denote a pretraining task with empirical quantities \((\widehat A_\tau,\widehat r_\tau)\). Given the state-level context from Stage~1, the executor with \(Q_{\mathrm{eff}}(C,W)\) maintains a vector \(s_{\tau,\ell}\), initialized at \(s_{\tau,0}=\widehat r_\tau\), and updates it at each depth according to
\[
s_{\tau,\ell+1}=F_\tau(C,W)s_{\tau,\ell},
\qquad
F_\tau(C,W):=I_n+C^\top+\widehat A_\tau W.
\]
At each step, the vector \(s_{\tau,\ell}\) produces an output increment \(-C^\top s_{\tau,\ell}\). At the target parameters, \(s_{\tau,\ell}=\widehat A_\tau^\ell\widehat r_\tau\), so these increments exactly match the empirical TD updates. Summing the increments, the executor's output at depth \(\ell\) is $v^\theta_{\tau,\ell}=-C^\top\sum_{j<\ell}s_{\tau,j}$.

\paragraph{Which pretraining signal identifies the executor?}
The endpoint objective $\mathcal J_{\rm end}(C,W)=\tfrac12\mathbb E_\tau\|v^\theta_{\tau,\infty}-\widehat v_{\tau,\infty}\|_2^2$, which supervises only the converged output as in equilibrium-based learning \citep{bai2019deep}, cannot uniquely identify the TD update rule. This non-identifiability is illustrated by the parameter family $(C,W)=(-aI_n,aI_n)$, $0<a\le1$, under which the executor performs TD with step size $a$, while its fixed point remains unchanged. Thus, whenever the iterations converge, different step sizes yield the same final prediction, giving a continuum of minimizers of $\mathcal J_{\rm end}$. In contrast, supervising the entire trajectory with $\mathcal J_{\rm traj}(C,W)=\tfrac12\mathbb E_\tau\sum_{\ell\le L}\|v^\theta_{\tau,\ell}-\widehat v_{\tau,\ell}\|_2^2$ distinguishes these update rules and removes this ambiguity.

\begin{proposition}[Local identifiability]
\label{prop:identify}
Suppose $L\ge2$ and
\[
\mathbb E_\tau[\widehat r_\tau\widehat r_\tau^\top]\succ0,
\qquad
\mathbb E_\tau[(\widehat r_\tau\widehat r_\tau^\top)\otimes(\widehat A_\tau^\top\widehat A_\tau)]\succ0.
\]
At the target parameters $(-I_n,I_n)$, the Jacobian of the trajectory map $(C,W)\mapsto(v^\theta_{\tau,\ell})_{\ell\le L}$ is injective. Consequently, the target is a strict local minimizer of $\mathcal J_{\rm traj}$. In contrast, the Jacobian of the endpoint map $(C,W)\mapsto v^\theta_{\tau,\infty}$ has the nonzero null direction $(-I_n,I_n)$, so endpoint supervision cannot identify the TD step size.
\end{proposition}

The proof in Appendix~\ref{app:learnability} uses only the first two depths: the first identifies $C$, and the second identifies $W$. To obtain a global guarantee, we replace $\widehat r_\tau$ at these depths with a calibration input $u$ satisfying $\mathbb E[uu^\top]=I_n$. This ensures that all input directions are represented, providing the input-side richness needed for global identification. And it yields the following quadratic objective:
\begin{equation}
 \mathcal J(C,W):=\tfrac12\,\mathbb E_{\tau,u}\Bigl[\bigl\|{-C^\top u}-u\bigr\|_2^2+\bigl\|F_\tau(C,W)\,u-\widehat A_\tau u\bigr\|_2^2\Bigr].
 \label{eq:probe}
\end{equation}
The first term identifies the TD increment, while the second identifies the update dynamics. This objective does not require a precomputed value function. Global identification then requires the pretraining tasks to distinguish all operator directions.
\begin{assumption}[Task richness]
\label{ass:richness}
There are constants $0<\kappa\le\beta'$ such that $\kappa I_n\preceq \mathbb E_\tau[\widehat A_\tau^\top\widehat A_\tau] \preceq\beta' I_n$.
The lower bound ensures that the pretraining tasks collectively cover every operator direction, while the upper bound controls the average magnitude of their effects.
\end{assumption}
Under Assumption~\ref{ass:richness}, we apply the full-batch gradient descent procedure in Algorithm~\ref{alg:td-pretraining} to minimize $\mathcal J$ with step size $\eta$. Let $(C_t,W_t)$ denote the resulting parameters after $t$ iterations.

\begin{algorithm}[!htbp]
\begin{algorithmic}[1]
\REQUIRE Pretraining task distribution $\mathcal D_{\rm pre}$, step size $\eta>0$, number of iterations $T$, fixed $P=P_{\mathrm{TD}}$, fixed encoder, initial $C_0,W_0\in\mathbb R^{n\times n}$
\FOR{$i=0,\ldots,T-1$}
    \STATE For each task $\tau$, run the encoder to obtain $(\widehat A_\tau,\widehat r_\tau)$ and form $F_\tau(C_i,W_i)=I_n+C_i^\top+\widehat A_\tau W_i$.
    \STATE Evaluate the full-batch objective $\mathcal J(C_i,W_i)$ in (\ref{eq:probe}) with the calibration input $u$.
    \STATE Update $C_{i+1}=C_i-\eta\nabla_C\mathcal J(C_i,W_i)$ and $W_{i+1}=W_i-\eta\nabla_W\mathcal J(C_i,W_i)$.
\ENDFOR
\RETURN $(P_{\mathrm{TD}},Q_{\mathrm{eff}}(C_T,W_T))$ with $Q_{\mathrm{eff}}$ as in Section~\ref{sec:learnability}.
\end{algorithmic}
\caption{Pretraining procedure for the effective TD parameters $(C,W)$.}
\label{alg:td-pretraining}
\end{algorithm}

\begin{theorem}[Global convergence in the effective parameter space]
\label{thm:learn}
Under Assumption~\ref{ass:richness}, $\mathcal J$ is a globally strongly convex and smooth quadratic with the \textbf{unique minimizer} $(C^\star,W^\star)=(-I_n,I_n)$. With $\lambda_-=\frac{3-\sqrt5}2\min\{1,\kappa\}$, $\lambda_+=\frac{3+\sqrt5}2\max\{1,\beta'\}$, and any $\eta\le1/\lambda_+$,
\[
 \|C_t-C^\star\|_F^2+\|W_t-W^\star\|_F^2
 \le
 (1-\eta\lambda_-)^{2t}\bigl(\|C_0-C^\star\|_F^2+\|W_0-W^\star\|_F^2\bigr).
\]
\end{theorem}
Thus the target is the unique global minimizer, reached from any initialization (Appendix~\ref{app:learnability}); the theorem identifies the executor, not the encoder, and what a jointly trained encoder learns is examined in Section~\ref{sec:experiments}.

\section{Experiments}
\label{sec:experiments}
\begin{figure}[!htbp]
\centering
\includegraphics[width=\linewidth]{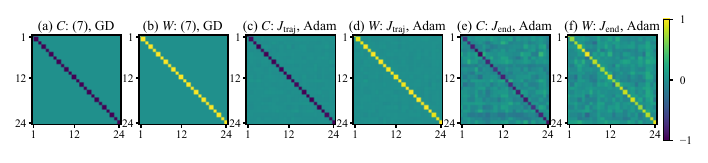}
\caption{\textbf{Parameter identification.} Learned $C, W$: calibration (\ref{eq:probe}) with full-batch gradient descent (a,b), trajectory supervision with Adam (c,d), and the depth-30 endpoint surrogate with Adam (e,f).}
\label{fig:exp-parameters}
\end{figure}

\paragraph{Effective parameter recovery.}
On 2,400 raw-batch tasks ($n=24$, $L=30$, five paired initializations) we compare the three signals of Section~\ref{sec:learnability}. Full-batch gradient descent on the calibration objective (\ref{eq:probe}) recovers $C=-I_{24}$, $W=I_{24}$ to normalized parameter error $1.0\times10^{-15}$ in 3,000 steps, as Theorem~\ref{thm:learn} predicts, and Adam on (\ref{eq:probe}) reaches $3.2\times10^{-16}$; Adam on $\mathcal J_{\rm traj}$ reaches $7.4\times10^{-2}$ in 6,000 steps and $3.3\times10^{-3}$ in 20,000, in line with Proposition~\ref{prop:identify}; Adam on $\mathcal J^{(30)}_{\rm end}$, the depth-30 bootstrapped form of $\mathcal J_{\rm end}$ (Appendix~\ref{app:exp-learning}), stops at $0.51$: its depth-30 output is within $0.04$ of TD on held-out tasks, but its trajectory is off by up to $0.63$, since $\mathcal J_{\rm end}$ constrains only the fixed point (Figure~\ref{fig:exp-parameters}, Appendix~\ref{app:exp-learning}).

\begin{figure}[t]
\centering
\includegraphics[width=\linewidth]{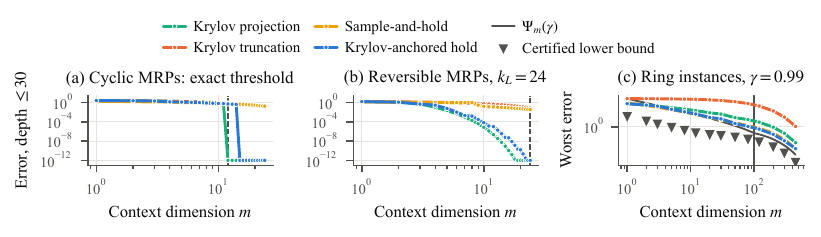}
\caption{\textbf{Context-dimension--accuracy tradeoff.} (a) Cyclic MRPs with $k_L=12$ and (b) reversible MRPs: mean error over 100 tasks per panel ($n=24$, $\gamma=0.9$, $L=30$), interquartile ranges shaded, dashed lines at $k_L$. (c) Worst-case rings at $\gamma=0.99$ against $\Psi_m$ of Theorem~\ref{thm:gamma-law} and a certified lower bound (vertical line: $m=1/(1-\gamma)$).}
\label{fig:context-budget}
\end{figure}

\paragraph{Context--accuracy tradeoff.}
Figure~\ref{fig:context-budget} varies the context dimension $m$ for Krylov projection and truncation, sample-and-hold, and the Krylov-anchored hold (Appendix~\ref{app:anchored}). On 100 covered 12-block cyclic MRPs with $k_L=12$ (panel a), Krylov projection becomes exact at the threshold of Theorem~\ref{thm:main}(a) ($0.86$ at $m=11$, $6\times10^{-16}$ at $m=12$), while truncation and sample-and-hold improve gradually; on reversible MRPs with $k_L=24$ (panel b) Krylov projection still drops from $1.1\times10^{-1}$ at $m=4$ to $2.2\times10^{-7}$ at $m=12$. On worst-case rings at $\gamma=0.99$ (panel c), sample-and-hold and the anchored hold stay within $[0.45,2.7]\times\Psi_m$, with sub-horizon slopes $-0.89$ and $-0.93$ against the predicted $-1$, whereas Krylov truncation has slope $-0.14$ and is up to $37\times$ worse, as Theorem~\ref{thm:gamma-law} predicts.

\paragraph{Memory usage.}
In Table~\ref{tab:exp-memory-sweep}, the block cycles ($k_L=12$ by construction) check the exact threshold: the context retains $(2n+1)(k_L+1)$ values independent of $N$, is not smaller than the statistics at $n=24$, and is $94.6\%$ smaller at $n=480$; the bound $nk_L+k_L$ of Theorem~\ref{thm:main}(c) is a class-level reference, not an instance-level certificate for these tasks, whose trajectories are fixed by the block labels alone (peak memory: Appendix~\ref{app:exp-memory}). The feature rows run TD (\ref{eq:lfa-td}) through the encoder of Theorem~\ref{thm:features}: on Boyan's chain ($d=4$, $k^\phi_L=4$) nothing is saved over the $d^2+d$ LSTD statistics; on mountain car with tile-coded features ($d=512$) the trajectory has full exact rank but at most $10$ singular values above $10^{-2}\sqrt L$, and a context of dimension $32$ is accurate to $1.1\times10^{-2}$ or better with $108$--$117$~KiB, $6$--$7\times$ below the LSTD statistics and independent of $N$. On sampled tabular tasks of full exact rank, contexts accurate to $10^{-2}$ are $9\times$ smaller than sparse TD on a dense kernel but not on a sparse gridworld (Appendix~\ref{app:exp-memory}).

\begin{table}[t]
\caption{\textbf{Retained memory} (KiB, float32). Top: 12-block cycles with $k_L=12$. Bottom: feature TD (\ref{eq:lfa-td}), Statistics $=(A_N,b_N)$, Ours $=$ the exact context (Boyan) or the dimension-$32$ context with error $\le1.1\times10^{-2}$ (mountain car; Raw stored sparsely). Peak as in Definition~\ref{def:memory}.}
\label{tab:exp-memory-sweep}
\centering\small\setlength{\tabcolsep}{4pt}
\begin{tabular*}{\linewidth}{@{\extracolsep{\fill}}lrrrrrrr@{}}
\toprule
Task & $n$ or $d$ & $N$ & Raw & Statistics & Sparse TD & Ours & Peak\\
\midrule
Block cycle & 24 & 960 & 11.44 & 2.44 & \textbf{0.75} & 2.49 & 37.5\\
Block cycle & 480 & 19200 & 228.8 & 903.8 & 102.2 & \textbf{48.8} & 10822.5\\
\midrule
Boyan's chain & 4 & 20000 & 703.1 & \textbf{0.08} & -- & 0.18 & 1.1\\
Mountain car & 452 & 320000 & 21250 & 799.8 & -- & \textbf{116.7} & 9226\\
\bottomrule
\end{tabular*}
\end{table}

\paragraph{Learned bottleneck encoders.}
Finally we train standard softmax Transformers that compress the TD iterates of the fixed first stage into $m$ latent tokens by cross-attention and emit a context for the executors of Appendix~\ref{app:executors}, so every model lies in $\mathfrak C_m$ (Appendix~\ref{app:exp-learned}; three seeds, 8,000 updates per point). The learned encoders beat sample-and-hold on held-out tasks for $m\le8$ ($0.50$--$0.89$ against $1.00$--$1.72$) and the anchored hold for $m\le4$, but their error does not decrease with $m$: it stays at $0.61$--$0.63$ (clocked, $m\ge4$) and $0.50$--$0.54$ (time-invariant, $m\ge8$) up to $m=24$, where the compiled contexts reach $0.10$ and $0.03$; standard bottleneck encoders do not discover the Krylov or clock structure that turns tokens into accuracy, which the compiled encoders supply.

\section{Conclusion}
\label{sec:conclusion}
In this paper, we study how much context a Transformer needs to reproduce TD evaluation without updating its parameters. We show that the minimum context dimension required to reproduce the first \(L\) TD iterates is the Krylov dimension \(k_L\). We construct a fixed-parameter linear-attention network that produces this compressed context directly from raw transitions, with retained memory within a factor of two of the optimum. When the context is too small for exact recovery, we characterize the best achievable worst-case approximation error and show that a simple sample-and-hold method achieves the same rate. We further show that pretraining can recover the shared executor under suitable task diversity, whereas supervising only the final output cannot identify the TD update rule. Our theoretical guarantees focus on TD policy evaluation with explicitly constructed encoders. Whether jointly learned encoders can achieve the same memory and accuracy guarantees and whether these results extend to policy control, remain open questions.
\label{main-text-end}

\clearpage
\subsection*{AI use statement}
In this work, we used ChatGPT Astra and Claude Fable 5.1 as AI-assisted research tools. Their use included polishing the writing and presentation of the paper, simplifying and improving the exposition of several proofs, assisting with the implementation and execution of part of the experiments, and helping extend the analysis from one-hot state representations to general $d$-dimensional feature representations. All AI-assisted mathematical arguments, derivations, and proofs were independently reviewed and verified by the authors. The authors take full responsibility for the correctness of all theoretical claims, proofs, experimental results, and the final content of the paper.

\subsection*{Ethics Statement}
This work does not involve human subjects, personal or sensitive data, or experiments that raise specific ethical concerns. The study is primarily theoretical and computational in nature, and all experiments are conducted using controlled settings without involving real-world participants or private information. We therefore do not identify any significant ethical issues associated with this work.

\subsection*{Reproducibility Statement}
We provide detailed descriptions of the theoretical constructions, experimental settings, and implementation procedures in the appendices. The appendices include the necessary architectural specifications, proof details, experimental configurations, task-generation procedures, and evaluation protocols required to reproduce the reported results. In particular, the complete code for all experiments is included in the supplementary material, together with the corresponding scripts and configurations needed to reproduce the main figures, tables, and quantitative results reported in the paper. Where applicable, random seeds, training budgets, hyperparameters, and model configurations are also provided to facilitate independent verification and reproduction of our experiments.
\bibliography{iclr2027_conference}
\bibliographystyle{iclr2027_conference}
\clearpage
\appendix
\section*{Contents of the Appendices}
\begin{tabular*}{\linewidth}{@{}l@{\extracolsep{\fill}}r@{}}
\hyperref[app:architecture]{A\quad Architecture, head library, and end-to-end exactness} & \pageref{app:architecture}\\
\hyperref[app:softmax]{B\quad Two-head softmax realization of the executors} & \pageref{app:softmax}\\
\hyperref[app:memory-proof]{C\quad Memory lower bounds: real registers and bits} & \pageref{app:memory-proof}\\
\hyperref[app:gamma-law-proof]{D\quad Limited context: proofs, the Krylov-anchored hold, and instance-dependent rates} & \pageref{app:gamma-law-proof}\\
\hyperref[app:depth]{E\quad Execution depth} & \pageref{app:depth}\\
\hyperref[app:learnability]{F\quad Learnability of the shared executor} & \pageref{app:learnability}\\
\hyperref[app:classical]{G\quad Connections to classical statistical and reduced models} & \pageref{app:classical}\\
\hyperref[app:experiments]{H\quad Experimental details and additional results} & \pageref{app:experiments}\\
\hyperref[app:scope]{I\quad Scope and limitations} & \pageref{app:scope}
\end{tabular*}
\clearpage

\section{Architecture, Head Library, and End-to-End Exactness}
\label{app:architecture}
This appendix specifies the encoder of Section~\ref{sec:exact-context} at the level of individual heads and proves Theorem~\ref{thm:main}(a)--(b). All head parameters are fixed for public $n$, $L$, a context dimension $m$ where relevant, and, for the sample-and-hold context, the discount $\gamma$. Raw records, statistics, Krylov directions, duals, reduced matrices, and sampled iterates are activations. Routes may depend on public quantities and on the current stage, never on task values; only the gates computed in the forward pass decide how many candidate tokens survive release.

\subsection{Coverage and Empirical Scope}
\label{app:exact-context}
Full state coverage means $N_i>0$ for every state; it does not require every transition $i\to j$ to be observed. It makes the empirical rows, reward means, and the visitation normalization in (\ref{eq:empirical-td-sample}) well defined, and expanding the sample update gives the closed form $\widehat v_\ell=\sum_{j<\ell}\widehat A^j\widehat r$ without any independence assumption. Coverage neither implies that $\widehat P$ is irreducible nor controls the statistical error of $\widehat v_\ell$ relative to the population trajectory; the statistical error is measured separately in Appendix~\ref{app:exp-accuracy}. The exact results and the memory law of Appendix~\ref{app:memory-proof} require no irreducibility. The stationary-norm statements of Section~\ref{sec:approx-context} add their own positive-stationary-distribution or contraction condition. For data that do not cover every state, a completion or a restricted query domain must be specified before the target is defined; we do not impute missing rows.

\subsection{Token Layout and Selector Heads}
\label{app:parameter-library}
Every encoder token has the same block layout, so that one selector matrix $E_a$ addresses block $a$ in every token. The blocks are, in order: three width-$n$ blocks $\mathrm{id}$, $\mathrm{succ}$, and $\mathrm{slot}$ for the current state, the successor state, and the candidate-position tag; three scalars $\mathrm{one}$, $\gamma$, and $\mathrm{rew}$ for a constant, the input discount, and the observed reward; the statistics $\mathrm{counts}\in\mathbb R^n$, $\mathrm{rsum}$, $\mathrm{visits}$, and $\mathrm{inv}$; the model blocks $\mathrm{Prow},\mathrm{Arow}\in\mathbb R^n$, $r$, and a work vector $\mathrm{Rvec}\in\mathbb R^n$; the Krylov blocks $B,\mathrm{dual},AB,\mathrm{Hcol},\mathrm{Hminus},Y\in\mathbb R^n$ with the scalars $\mathrm{norm}$, $\alpha$, $g$, and $\mathrm{gate}$. The total width is $13n+11$. A head is specified by its key block $k$, query block $q$, value block $v$, output block $o$, and sign $\sigma$, meaning $P=\sigma E_o^\top E_v$ and $Q=E_k^\top E_q$ in (\ref{eq:encoder-head}); equivalently $W_K=E_k$, $W_Q=E_q$, $W_V=E_v$, $W_O=\sigma E_o^\top$, so every nonzero projection entry is $\pm1$. All heads of one layer read the activations before any head writes, which is what makes simultaneous projection subtraction possible; if a block is overwritten, a fixed linear reset clears it first.

\subsection{Stage 1: Raw Records to Statistics}
\label{app:stage1}
Let $E_S=[e_{S_0},\ldots,e_{S_{N-1}}]$, $E_{S^+}=[e_{S_1},\ldots,e_{S_N}]$, and $R_+^\top=[R_1,\ldots,R_N]$. Displaying the transpose of the active-token matrix, with the raw tokens in the first block row and the state tokens in the second,
\begin{equation}
\bigl(U^{(1)}_0\bigr)^\top
=
\begin{array}{c}
\left[
\begin{array}{ccccc}
E_S^\top & E_{S^+}^\top & R_+ & \mathbf 1_N & 0\\
I_n & 0 & 0 & \mathbf 1_n & \gamma\mathbf 1_n
\end{array}
\right]
\\[-2pt]
\begin{array}{ccccc}
\scriptstyle \mathrm{id} & \scriptstyle \mathrm{succ} & \scriptstyle \mathrm{rew} & \scriptstyle \mathrm{one} & \scriptstyle \gamma
\end{array}
\end{array},
\label{eq:stage1-input}
\end{equation}
where the statistic blocks of the state tokens are initialized to zero. To aggregate raw records into state statistics, every raw token is routed to every state token with $Q_{\rm match}[\mathrm{id},\mathrm{id}]=I_n$, so that the score between raw token $t$ and state token $i$ is $e_{S_t}^\top e_i=\mathbf 1\{S_t=i\}$. Keeping this score, three heads with value--output maps $P_{\rm count}[\mathrm{counts},\mathrm{succ}]=I_n$, $P_{\rm reward}[\mathrm{rsum},\mathrm{rew}]=1$, and $P_{\rm visit}[\mathrm{visits},\mathrm{one}]=1$ give
\[
u_i^{\rm state}[\mathrm{counts}]=N_{i,:},\qquad u_i^{\rm state}[\mathrm{rsum}]=\sum_{t:S_t=i}R_{t+1},\qquad u_i^{\rm state}[\mathrm{visits}]=N_i .
\]
Then $u_i^{\rm state}[\mathrm{inv}]=\operatorname{recip}_0(N_i)$. With state self-routing and $Q_{\rm norm}[\mathrm{one},\mathrm{inv}]=1$, the heads $P_{\rm P}[\mathrm{Prow},\mathrm{counts}]=I_n$ and $P_{\rm r}[r,\mathrm{rsum}]=1$ produce $u_i^{\rm state}[\mathrm{Prow}]=N_{i,:}/N_i=\widehat P_{i,:}$ and $u_i^{\rm state}[r]=\widehat r_i$, and a further self-routed head with $Q_\gamma[\mathrm{one},\gamma]=1$ and $P_{\rm A}[\mathrm{Arow},\mathrm{Prow}]=I_n$ gives $u_i^{\rm state}[\mathrm{Arow}]=\gamma\widehat P_{i,:}=\widehat A_{i,:}$. The raw tokens are then released. Because the three aggregation heads are sums over raw tokens with a fixed query per state token, they can be evaluated one chunk of $C$ records at a time, adding each chunk's contribution before the next chunk is loaded, as in the recurrent form of linear attention \citep{katharopoulos2020linear}; the result is identical and the live raw storage is $O(Cn)$ rather than $O(Nn)$. Our implementation uses this chunked evaluation.

\subsection{The State-Level Context}
\label{app:state-level}
Before compressing further, note that the $n$ state tokens already form an exact $n$-dimensional context representation. Take $B=I_n$, $Y=\widehat A^\top$, and $s_0=\widehat r$, i.e.,
\begin{equation}
    \overline Z_0(q)
    =
    \begin{bmatrix}
        I_n & e_q\\
        \widehat A^\top & 0\\
        \widehat r^\top & 0
    \end{bmatrix}
    \in\mathbb R^{(2n+1)\times(n+1)},
\label{eq:linear-full-context}
\end{equation}
so that the $i$th source token contains its state identity $e_i$, its discounted empirical transition row, and its empirical reward. With the executor (\ref{eq:executor}) and mask $M_n=\operatorname{diag}(I_n,0)$, Lemma~\ref{lem:executor} gives $s_{\ell+1}=[I+(\widehat A^\top-I)^\top I]s_\ell=\widehat As_\ell$, hence $s_\ell=\widehat A^\ell\widehat r$ and $-c_\ell=e_q^\top\sum_{j<\ell}\widehat A^j\widehat r=\widehat v_\ell(q)$. The first $2n$ rows never change, so the sampled RL information and the query remain intact throughout the forward pass, and the empirical value is exactly recoverable after compressing $N$ records into $n$ state tokens. This is the context on which the learnability analysis of Section~\ref{sec:learnability} operates.

\subsection{Stage 2: Discovering and Compiling Krylov Directions}
\label{app:stage2}
Let $h=\min\{n,L\}$. Stage~2 retains the $n$ state tokens, introduces one work token, and allocates $h$ candidate tokens with fixed position tags $J_h=[e_1,\ldots,e_h]$ in the block $\mathrm{slot}$:
\begin{equation}
\bigl(U^{(2)}_0\bigr)^\top
=
\begin{array}{c}
\left[
\begin{array}{ccccc}
I_n & \mathbf 1_n & \widehat A & \widehat r & 0\\
0 & 1 & 0 & 0 & 0\\
0 & \mathbf 1_h & 0 & 0 & J_h^\top
\end{array}
\right]
\\[-2pt]
\begin{array}{ccccc}
\scriptstyle \mathrm{id} & \scriptstyle \mathrm{one} & \scriptstyle \mathrm{Arow} & \scriptstyle r & \scriptstyle \mathrm{slot}
\end{array}
\end{array},
\label{eq:stage2-input}
\end{equation}
where the block rows are the state, work, and candidate tokens. Each candidate token $j$ uses $B$ for the current Krylov direction $b_j$, $\mathrm{norm}$ for $d_j=b_j^\top b_j$, $\alpha$ for $\operatorname{recip}_0(d_j)$, $\mathrm{dual}$ for $\omega_j$, $AB$ for $\widehat Ab_j$, $\mathrm{Hcol}$ for a column of the reduced operator, $\mathrm{Hminus}$ for that column minus its tag, $Y$ for the compiled block, $g$ for the initial reduced coefficient, and $\mathrm{gate}$ for the active-direction indicator; these blocks start at zero.

\paragraph{Initialization and one Arnoldi round.} A state-to-work head with $Q_{\rm gather}[r,\mathrm{one}]=1$ and $P_{\rm gather}[\mathrm{Rvec},\mathrm{id}]=I_n$ forms $u^{\rm work}[\mathrm{Rvec}]=\sum_ie_i\widehat r_i=\widehat r$, which a work-to-candidate copy ($Q[\mathrm{one},\mathrm{one}]=1$, $P[B,\mathrm{Rvec}]=I_n$) writes into the first candidate as $b_1=\widehat r$. At round $j$, candidate self-attention forms the squared norm ($Q[B,B]=I_n$, $P[\mathrm{norm},\mathrm{one}]=1$), then $\alpha_j=\operatorname{recip}_0(d_j)$, the dual $\omega_j=\alpha_jb_j$ ($Q[\mathrm{one},\alpha]=1$, $P[\mathrm{dual},B]=I_n$), and the gate $\eta_j=d_j\alpha_j$ ($Q[\mathrm{norm},\alpha]=1$, $P[\mathrm{gate},\mathrm{one}]=1$). The operator application and the projection use
\[
Q_{\rm op}[\mathrm{Arow},B]=I_n,\quad P_{\rm op}[AB,\mathrm{id}]=I_n,\qquad
Q_{\rm proj}[\mathrm{dual},B]=I_n,\quad P_{\rm proj}[B,B]=-I_n,
\]
routed from the states to candidate $j$ and from candidates $1,\ldots,j$ to candidate $j+1$ respectively, after a copy of $AB$ from candidate $j$ into the block $B$ of candidate $j+1$. Thus $u_j^{\rm cand}[AB]=\sum_ie_i(\widehat A_{i,:}b_j)=\widehat Ab_j$ and
$b_{j+1}=\widehat Ab_j-\sum_{i\le j}b_i\,\omega_i^\top\widehat Ab_j$,
which is (\ref{eq:arnoldi}). Since $\eta_j=\mathbf 1\{b_j\ne0\}$ and breakdown is permanent (Lemma~\ref{lem:basis}), $(\eta_1,\ldots,\eta_h)=(1^{k_L},0^{h-k_L})$, and token release keeps exactly the first $k_L$ candidates.

\paragraph{Compilation.} With $B=[b_1,\ldots,b_{k_L}]$ and $\Omega=[\omega_1,\ldots,\omega_{k_L}]$, heads over all candidate pairs form the reduced operator column by column with $Q_H[\mathrm{dual},AB]=I_n$ and $P_H[\mathrm{Hcol},\mathrm{slot}]=I_n$, so that candidate $t$ receives $\sum_ie_i(\omega_i^\top\widehat Ab_t)$, the $t$th column of $H=\Omega^\top\widehat AB$; a copy and a subtraction of the tag form $\mathrm{Hminus}=\mathrm{Hcol}-\mathrm{slot}$; a copy of $B$ into $Y$ followed by $Q_Y[\mathrm{Hminus},\mathrm{slot}]=I_n$, $P_Y[Y,\mathrm{dual}]=I_n$ adds $\sum_i\omega_i(H_{ti}-\delta_{ti})$ to candidate $t$, giving $Y=B+\Omega(H-I)^\top$; and a work-to-candidate head with $Q_g[\mathrm{Rvec},\mathrm{dual}]=I_n$, $P_g[g,\mathrm{one}]=1$ gives $g_t=\omega_t^\top\widehat r$. The retained context is (\ref{eq:krylov-context}). In total the exact schedule uses $6h+12$ head calls, two reciprocal overwrites, and explicit release. Seven further heads implement the terminal correction used by Krylov truncation in Appendix~\ref{app:mm-proof}: they gather $g$ into the work vector, advance it through $m$ reduced multiplications by $H$ (key $\mathrm{slot}$, query $\mathrm{Rvec}$, value $\mathrm{Hcol}$ into $AB$, then copied back into $\mathrm{Rvec}$), read the leading coefficient, and subtract $H^mg$ from the last column of $H$ through a gate and a scaled inverse.

\paragraph{Balancing.} The implementation used in the experiments additionally rescales each active candidate to unit $\ell_1$ norm before it is used, with $\|b_j\|_1$ formed by a ReLU absolute value and a summation head, and applies the projection subtraction twice. Balancing multiplies $b_j$ and divides $\omega_j$ by the same positive scalar, so $B\Omega^\top$, $H$, $g$, and the executed trajectory are unchanged (Lemma~\ref{lem:normalized}), while the dual entries stay bounded (Appendix~\ref{app:exp-stability}).

\subsection{Linear Features: Proof of Theorem~\ref{thm:features}}
\label{app:features}
\paragraph{Stage 1 for features.} Raw token $t$ carries $(\phi_t,\ \phi_t-\gamma\phi'_t,\ R_{t+1})$, where the difference block is a fixed linear map of the input $(\phi_t,\phi'_t)$ and the public $\gamma$; coordinate token $i\in[d]$ carries $e_i$ with zero-initialized blocks $\mathrm{Acol}\in\mathbb R^d$ and $b$. Two heads routed from every raw token to every coordinate token, with key $\phi$ and query $\mathrm{id}$, so that the score between raw token $t$ and coordinate token $i$ is $\phi_i(S_t)$, and with values $\phi_t-\gamma\phi'_t$ and $R_{t+1}$, give
\[
 u_i[\mathrm{Acol}]=\textstyle\sum_t\phi_i(S_t)(\phi_t-\gamma\phi'_t)=(A_N)_{i,:},\qquad u_i[b]=\sum_t\phi_i(S_t)R_{t+1}=(b_N)_i .
\]
A self-routed head with key $\mathrm{one}$, query the public $\alpha$, value $\mathrm{Acol}$, output $\mathrm{Arow}$, and $\sigma=-1$, followed by adding $\mathrm{id}$, forms $\mathrm{Arow}=e_i-\alpha(A_N)_{i,:}=M_{i,:}$, and the same head on $b$ forms $s_0=\alpha b_N$. No reciprocal is used in this stage: in the tabular case the visitation normalization $1/N_i$ is what requires $\operatorname{recip}_0$, whereas (\ref{eq:lfa-td}) is un-normalized. As in Appendix~\ref{app:stage1}, the sums can be streamed one chunk at a time. Stage~2 (Appendix~\ref{app:stage2}) runs unchanged on $(M,s_0)$ in place of $(\widehat A,\widehat r)$, including its reciprocals $\operatorname{recip}_0(d_j)$ in $\omega_j$ and $\eta_j$; the reciprocal-free claim of Theorem~\ref{thm:features}(b) concerns Stage~1 only. Nothing in Lemmas~\ref{lem:basis} and~\ref{lem:compile} uses stochasticity of the operator.

\begin{proof}[Proof of Theorem~\ref{thm:features}]
\emph{(a)} The increments $\theta_{j+1}-\theta_j=M^js_0=Ds_j$ lie in $\operatorname{col}(D)$ for $j<L$, so $m\ge\operatorname{rank}D\ge k^\phi_L$; the Krylov context of (b) attains it.
\emph{(b)} By the heads above, the coordinate tokens carry $(M,s_0)$ exactly. Lemmas~\ref{lem:basis} and~\ref{lem:compile} applied to $(M,s_0)$ give the context $[B;Y;g^\top]$ with exactly $k^\phi_L$ tokens, and the argument of Theorem~\ref{thm:main}(b) gives $s_\ell=H^\ell g$ and $-c_\ell(q)=\phi(q)^\top B\sum_{j<\ell}H^jg=\phi(q)^\top\sum_{j<\ell}M^js_0=\phi(q)^\top\theta_\ell$, where the query token carries $\phi(q)$ in its first block. The memory counts are those of Appendix~\ref{app:memory-accounting} with $d$ for $n$; Stage~1 keeps a chunk of raw tokens of width $2d+1$ and $d$ coordinate tokens of width $O(d)$.
\emph{(c)} Fix a base pair $(M_0,s_0)$ with $\dim\mathcal K_L(M_0,s_0)=k$ in the open set, put $\Delta_0=[s_0,M_0s_0,\ldots,M_0^{k-1}s_0]$ and $M_0^ks_0=\Delta_0\alpha_0$. For $\theta=(\Delta,\alpha)\in\mathbb R^{d\times k}\times\mathbb R^k$ near $\theta_0$ define, with the companion matrix $C(\alpha)$ of Lemma~\ref{lem:family},
$M(\theta)=M_0+(\Delta C(\alpha)-M_0\Delta)\Delta^+$ and $s(\theta)=\Delta e_1$. Near $\theta_0$, $\Delta$ has full column rank, so $\Delta^+$ is continuous, $M(\theta)\Delta=\Delta C(\alpha)$, and hence $M(\theta)^js(\theta)=\Delta C(\alpha)^je_1$: the Krylov dimension is $k$ and the pair stays in the open set for $\theta$ close to $\theta_0$. The trajectory determines $\Delta$ (the first $k$ increments) and then $\alpha$ (the $(k+1)$st increment, using $L\ge k+1$), so $\mathrm{Enc}$ composed with $\theta\mapsto(M(\theta),s(\theta))$ is a continuous injection of an open subset of $\mathbb R^{dk+k}$, and invariance of domain gives $M\ge dk+k$ exactly as in the proof of Theorem~\ref{thm:memory-law}(1). The context of (b) retains $(2d+1)k$ registers, within the factor $(2d+1)/(d+1)<2$.
\end{proof}

\subsection{Executors}
\label{app:executors}
\begin{definition}[Time-invariant and depth-counter executors]
\label{def:executors}
An $m$-dimensional context has static source blocks $X,Y\in\mathbb R^{n\times m}$, source scalars $s_0\in\mathbb R^m$, and a query token $(e_q,0_n,c_0=0)$, so that $\overline Z_0(q)=[X,e_q;\,Y,0_n;\,s_0^\top,0]$. The \textbf{time-invariant} executor is the looped layer (\ref{eq:executor}) with mask $M_m=\operatorname{diag}(I_m,0)$; its output is $-c_\ell$, and it gives $s_{\ell+1}=[I+(Y-X)^\top X]s_\ell$ and $c_{\ell+1}=c_\ell-e_q^\top Xs_\ell$. The \textbf{depth-counter} executor replaces the source recurrence by fixed public time stamps $\tau_u$ and a counter $t_\ell=\ell$: the gate $s_{\ell,u}=1-\delta^2\operatorname{recip}_0(\delta^2)$ with $\delta=t_\ell+1-\tau_u$ equals $\mathbf 1\{\ell+1=\tau_u\}$, and the same query head gives $-c_\ell=\sum_{u:\tau_u\le\ell}e_q^\top x_u$.
\end{definition}
In both cases $-c_\ell(q)=e_q^\top X\sum_{j<\ell}s_j$, so every output is an $m$-dimensional context representation in the sense of Definition~\ref{def:context} with $D=X$. The time-invariant executor realizes the recurrences $s_{\ell+1}=Ks_\ell$, which is how looped attention without timestep information behaves \citep{yang2024looped}; the depth-counter executor adds a timestep signal in the sense of \citet{xu2025looped}. The two can share a looped block and write into one query scalar.
\begin{lemma}[Executor identities]
\label{lem:executor}
For any query-independent $X,Y,s_0$, the time-invariant executor satisfies $s_{\ell+1}=[I+(Y-X)^\top X]s_\ell$ and $c_{\ell+1}-c_\ell=-e_q^\top Xs_\ell$. For the depth-counter executor, $c_{\ell+1}-c_\ell=-\sum_u\mathbf 1\{\ell+1=\tau_u\}e_q^\top x_u$.
\end{lemma}
\begin{proof}
The only nonzero entry of $P_{\rm TD}$ is its last diagonal entry, so static rows never change. The score between source tokens $i$ and $t$ is $x_i^\top(y_t-x_t)$ and the source--query score is $-x_i^\top e_q$. Transposing the last-row update gives both identities. The masked query never acts as a source, so the source scalars do not depend on $q$. For the depth-counter executor, $\delta^2\operatorname{recip}_0(\delta^2)=\mathbf 1\{\delta\ne0\}$ because $\delta^2\ge0$, and the query head is identical.
\end{proof}

\subsection{Exactness and Minimality}
\label{app:exactness}
\begin{lemma}[Primal--dual orthogonality and saturation]
\label{lem:basis}
In exact arithmetic, the nonzero vectors produced by (\ref{eq:arnoldi}) are mutually orthogonal, their duals satisfy $\omega_i^\top b_j=\delta_{ij}$, and a zero candidate is followed only by zeros. After $h=\min\{n,L\}$ rounds the nonzero columns span $\mathcal K_L$, and their number is $k_L$.
\end{lemma}
\begin{proof}
If $b_j\ne0$, then $d_j>0$ and $\omega_j=b_j/d_j$. If the preceding columns are nonzero and orthogonal, then for $p\le j$,
$b_p^\top b_{j+1}=b_p^\top\widehat Ab_j-\sum_{i\le j}b_p^\top b_i\,\omega_i^\top\widehat Ab_j=0$.
A zero $b_j$ gives $d_j=0$, $\omega_j=0$, $\widehat Ab_j=0$, and hence $b_{j+1}=0$. Before breakdown, induction gives $b_j=\widehat A^{j-1}\widehat r+u_j$ with $u_j\in\mathcal K_{j-1}$, so the first $j$ columns span $\mathcal K_j$. If $b_{k+1}=0$ after $k$ nonzero columns, the rearranged recurrence $\widehat Ab_i=b_{i+1}+\sum_{p\le i}b_p\omega_p^\top\widehat Ab_i$ shows that the first $k$ columns span an invariant subspace containing $\widehat r$. If no breakdown occurs, the $h$ columns span $\mathcal K_h=\mathcal K_L$ when $L\le n$, and $\mathbb R^n$ otherwise. The case $\widehat r=0$ gives the zero sequence.
\end{proof}

\begin{lemma}[Forward compilation]
\label{lem:compile}
The heads of Appendix~\ref{app:stage2} form (\ref{eq:krylov-context}), and every inactive candidate carries a zero payload.
\end{lemma}
\begin{proof}
With candidate tags $p_i$, the reduced-column head produces $h_j=\sum_ip_i(\omega_i^\top\widehat Ab_j)$, whose coordinates form column $j$ of the padded $H$. The scalar head gives $g_j=\omega_j^\top\widehat r$. The compilation head writes $y_j=b_j+\sum_i\omega_i(h_i-p_i)^\top p_j=b_j+\sum_i\omega_i(H_{ji}-\delta_{ji})$, which on active columns is $B+\Omega(H-I)^\top$. For $j>k_L$ all of $b_j,\omega_j,\widehat Ab_j$ vanish, so the corresponding row and column of $H$ and the entries $g_j,y_j$ are zero. The gate $d_j\operatorname{recip}_0(d_j)$ is one exactly on nonzero columns, and by Lemma~\ref{lem:basis} the gate pattern is a prefix.
\end{proof}

\begin{proof}[Proof of Theorem~\ref{thm:main}(a) and (b)]
\emph{(a)} Let $\{Z_\ell\}$ be an exact $m$-dimensional context representation with decoder $D$. Each increment $\widehat A^j\widehat r=\widehat v_{j+1}-\widehat v_j=Ds_j$ lies in $\operatorname{col}(D)$ for $j<L$, so $\mathcal K_L\subseteq\operatorname{col}(D)$ and $m\ge\operatorname{rank}D\ge k_L$. The context (\ref{eq:krylov-context}) attains $m=k_L$ by part (b).

\emph{(b)} Stage~1 produces $(N_{i,:},\sum_tR_{t+1}\mathbf 1\{S_t=i\},N_i)$ in state token $i$, since $\sum_t[e_{S_{t+1}};R_{t+1};1](e_{S_t}^\top e_i)$ equals these quantities; multiplying by $\operatorname{recip}_0(N_i)=1/N_i$ and then by $\gamma$ yields $(\widehat A_{i,:},\widehat r_i)$. Lemmas~\ref{lem:basis} and~\ref{lem:compile} give (\ref{eq:krylov-context}) with exactly $k_L$ retained tokens. Let $\Pi=B\Omega^\top$, which is an orthogonal projection onto $\mathcal K_L$ because $\Omega=B(B^\top B)^{-1}$ for orthogonal columns. Then $Bg=\Pi\widehat r=\widehat r$, and if $BH^jg=\widehat A^j\widehat r$ with $j+1<L$, then $BH^{j+1}g=\Pi\widehat A^{j+1}\widehat r=\widehat A^{j+1}\widehat r$. Lemma~\ref{lem:executor} with $(Y-B)^\top B=(H-I)\Omega^\top B=H-I$ gives $s_\ell=H^\ell g$ and $-c_\ell=e_q^\top\sum_{j<\ell}BH^jg=\widehat v_\ell(q)$ for $\ell\le L$. If $k_L<L$, breakdown implies invariance of $\mathcal K_L$, and exactness holds at every depth. The memory counts are established in Appendix~\ref{app:memory-accounting}.
\end{proof}

\subsection{Continuity within Rank Strata}
\label{app:continuity}
Let $\Sigma_k=\{(\widehat A,\widehat r):\dim\mathcal K_L(\widehat A,\widehat r)=k\}$. By Lemma~\ref{lem:basis}, the recurrence (\ref{eq:arnoldi}) depends only on $(\widehat A,\widehat r)$ and uses neither stochasticity nor rationality, so the second stage is exact for every real input.
\begin{lemma}[Continuity within each rank stratum]
\label{lem:continuity}
Restricted to $\Sigma_k$, the second-stage encoder is a rational function whose denominators do not vanish on $\Sigma_k$. It is therefore continuous on $\Sigma_k$. The same holds with positive $\ell_1$ balancing, which replaces $b$ by $b/\|b\|_1$ with $\|b\|_1$ computed by ReLU absolute values.
\end{lemma}
\begin{proof}
On $\Sigma_k$ the gates are identically $(1^k,0^{h-k})$. For $j\le k$, $b_j$ is a rational function whose denominators are products of $d_1,\ldots,d_{j-1}>0$, so $\operatorname{recip}_0(d_j)=1/d_j$ is continuous there. For $j>k$ every quantity is identically zero. The retained quantities $H,g,Y$ are polynomials in these. Balancing multiplies each active $b_j$ by $1/\|b_j\|_1$, which is continuous away from $b_j=0$.
\end{proof}

\subsection{Rank Discontinuity across Strata}
\label{app:numerical-margin}
For $P=\left[\begin{smallmatrix}0&1\\1&0\end{smallmatrix}\right]$ and $r_\varepsilon=(1,1+\varepsilon)^\top$, $\det[r_\varepsilon,\gamma Pr_\varepsilon]=-\gamma\varepsilon(2+\varepsilon)$. Hence the exact rank is $1$ at $\varepsilon=0$ and $2$ for $0<|\varepsilon|<1$. No finite composition of continuous maps can output the corresponding binary gate on this family, which is why $\operatorname{recip}_0$ and release are declared primitives. This is consistent with the memory law of Appendix~\ref{app:memory-proof}: an exact rank-adaptive encoder must be discontinuous across strata, yet it can be continuous within each stratum (Lemma~\ref{lem:continuity}). On an input class whose nonzero reciprocal arguments are separated from zero by a margin $\Delta>0$, the reciprocal can be replaced by a ReLU interpolant plus plateau gates with arbitrarily small output error and identical prefix decisions. Without a margin, tiny residuals yield large duals; positive $\ell_1$ balancing removes this growth in our experiments (Appendix~\ref{app:exp-stability}), but a tolerance-based rank is then a numerical policy, not the algebraic rank. The numerical-rank bound of Proposition~\ref{prop:instance}(b) and the threshold certificate of Appendix~\ref{app:threshold} give the certified replacements.

\subsection{Visit-Weighted Inner Product}
\label{app:weighted}
Replacing $b^\top b$ by $\langle b,b\rangle_w=b^\top D_wb$ with $w_i=N_i$ requires two more blocks, $\mathrm{wid}=N_ie_i$ in the state tokens and $\mathrm{DB}=D_wb_j$ in the candidates, and four heads: forming $\mathrm{wid}$ (key $\mathrm{one}$, query $\mathrm{visits}$, value $\mathrm{id}$, state self-route), weighting $B$ (key $\mathrm{id}$, query $B$, value $\mathrm{wid}$, states to candidate $j$), the squared norm (key $B$, query $\mathrm{DB}$, value $\mathrm{one}$, self-route), and the dual (key $\mathrm{one}$, query $\alpha$, value $\mathrm{DB}$, self-route). The duals $\omega_j=D_wb_j/\langle b_j,b_j\rangle_w$ are invariant to rescaling $w$, so no division by $N$ is needed. Lemma~\ref{lem:basis} holds for every positive definite inner product, so exactness and Lemma~\ref{lem:continuity} are unchanged. A \emph{budget-$m$} encoder runs $m$ rounds instead of $h$.

\subsection{Sample-and-Hold Encoder and Depth-Counter Executor}
\label{app:sh-encoder}
The schedule $\mathcal T=\mathcal T_m(\gamma)\cap[1,L]$ is public. Besides the state tokens, the encoder uses one trajectory token with blocks $\mathrm{cur}$ (initially $0$) and $\mathrm{inc}$, and $m'=|\mathcal T|\le m$ sample tokens with a block $y$. Five heads suffice: an initialization head (key $r$, query $\mathrm{one}$, value $\mathrm{id}$, states to trajectory, first layer only) sets $\mathrm{inc}=\widehat r$; in every layer an accumulation head (key $\mathrm{one}$, query $\mathrm{one}$, value $\mathrm{inc}$, output $\mathrm{cur}$, self-route), a step head (key $\mathrm{Arow}$, query $\mathrm{inc}$, value $\mathrm{id}$, output $\mathrm{inc}$, states to trajectory), and a clearing head (same as the accumulation head with $\sigma=-1$ and output $\mathrm{inc}$) perform $\mathrm{cur}\leftarrow\mathrm{cur}+\mathrm{inc}$ and $\mathrm{inc}\leftarrow\widehat A\,\mathrm{inc}$, because heads read synchronously; and at layer $\tau_u$ a latch head (key $\mathrm{one}$, query $\mathrm{one}$, value $\mathrm{cur}$, output $y$, trajectory to sample $u$) stores $y_u=\widehat v_{\tau_u}$. By induction, layer $t$ begins with $\mathrm{cur}=\widehat v_t$ and $\mathrm{inc}=\widehat A^t\widehat r$. A difference head (route $u-1\to u$, $\sigma=-1$) forms $x_u=y_u-y_{u-1}$. The encoder has $\max\mathcal T+1\le L+1$ layers and peak $O(n^2+nm)$ live scalars; after release the context retains $m'(n+1)$ scalars, $x_u$ and one time stamp per token. Using repeated squaring as in Appendix~\ref{app:doubling}, the depth can be reduced to $O(\log L)$ at the same peak.

The depth-counter executor of Definition~\ref{def:executors} uses the following fixed sublayers: a token-wise affine map $\delta\leftarrow t+1-\tau$; a self-routed head computing $\delta^2$; $\operatorname{recip}_0$; a self-routed head computing $s=1-\delta^2\operatorname{recip}_0(\delta^2)$; the query head (key $x$, query $e_q$, value $s$, $\sigma=-1$, source tokens to query), which has the form of the TD executor's query head; and the counter increment. If one prefers to avoid $\operatorname{recip}_0$, an unrolled version whose layer $\ell$ routes only the tokens with $\tau_u=\ell+1$ uses pure linear attention. When $\gamma$ is an input rather than public, the schedule can be computed in the network by threshold gates on $1-\gamma^i$ maintained in a work token; we do not use this variant.

\subsection{Krylov-Anchored Encoder}
\label{app:anchored-encoder}
The anchored encoder of Appendix~\ref{app:anchored} first runs the budget-$k$ visit-weighted Krylov encoder and packs its $k$ tokens. A work token then accumulates two trajectories in parallel: the TD trajectory, with the heads of Appendix~\ref{app:sh-encoder}, and the projected trajectory $\widetilde v^{\rm G}_t=B\sum_{i<t}s_i$, using the TD executor's source heads on the packed Krylov tokens together with a head with key $s$, query $\mathrm{one}$, value $B$, and target $\mathrm{gcur}$. At layer $k+i_s$ a fixed linear map forms $\mathrm{cur}-\mathrm{gcur}=e_{k+i_s}$, and a latch head stores it; a difference head forms the increments. Execution uses one looped block containing both the TD source and query heads (for the Krylov tokens) and the depth-counter heads (for the residual tokens); all heads write the same query scalar, so the output is $\widetilde v^{\rm G}_\ell+e_{\sigma(\ell)}$. The implementation matches this closed form to $2.8\times10^{-15}$ (Appendix~\ref{app:exp-checks}).

\subsection{Retained, Peak, and Shared Memory}
\label{app:memory-accounting}
The stored objects of the exact construction and their treatment are as follows. Raw transition tokens occupy $O(Cn)$ locations when streamed and $O(Nn)$ in a batch, and are released after Stage~1. State tokens and statistics occupy $O(n^2)$ locations and are released after the context is packed. Candidates, duals, images, and the columns of $H$ occupy $O(nh+h^2)$ locations at fixed capacity $h$ (or $m$) and are released after packing. The retained context $B,Y,g$ occupies $(2n+1)k_L$ locations, of which $B$ is the decoder; the sample-and-hold tokens $x_u,\tau_u$ occupy $m(n+1)$; the query token and its accumulator occupy $2n+1$. Scores and workspace occupy $O(n^2+nh)$ temporary locations and are charged at peak. The fixed selectors and routes are shared parameters and are counted separately.
\begin{proof}[Memory bounds in Theorem~\ref{thm:main}(b)]
After packing, the context blocks occupy $nk_L+nk_L+k_L+(2n+1)=(2n+1)(k_L+1)$ locations, which is $M_{\rm ret}$ in Definition~\ref{def:memory}: the decoder $D=B$ aliases the first block and is counted once, and $c_\ell$ accumulates the queried value, so no cumulative coefficient vector is stored. During Stage~1, the state-matching heads need score blocks of size $C\times n$ (or $N\times n$) and output blocks of size $n\times n$, never an all-token square score matrix. After release, the model, candidates, reduced columns, and routed scores fit in $O(n^2+nh)$ locations. Execution scores have $k_L(k_L+1)$ entries. Hence $M_{\rm peak}=O(Nn+n^2+nh)$ in the batch form and $O(Cn+n^2+nh)$ in the streamed form. The measured peaks of our implementation appear in Appendix~\ref{app:exp-memory}.
\end{proof}
\section{Two-Head Softmax Realization of the Executors}
\label{app:softmax}
This appendix realizes both executors of Definition~\ref{def:executors} with one looped layer of standard masked softmax attention with two heads. The construction is a \emph{compilation}: the encoder writes log-probability codes into the tokens, and the output map applies a gain that is fixed for the whole task family. It shows that nothing in Sections~\ref{sec:exact-context}--\ref{app:depth} depends on the attention being linear, and it changes the retained-memory counts by at most a constant factor. It says nothing about what a trained softmax model does; the models trained in this paper are the executor parameters of Appendix~\ref{app:exp-learning} and the bottleneck encoders of Appendix~\ref{app:exp-learned}. Earlier softmax compilations of TD are of the same kind \citep{xie2026beyondlinear}.

\paragraph{Reduced systems.} For this appendix an $m$-dimensional context is described by a triple $(X,H,g)$ with $X\in\mathbb R^{n\times m}$, $H\in\mathbb R^{m\times m}$, and $g\in\mathbb R^m$; the executor runs $s_{\ell+1}=Hs_\ell$ from $s_0=g$ and outputs $\widetilde v_\ell(q)=e_q^\top X\sum_{j<\ell}s_j$. The Krylov context (\ref{eq:krylov-context}) is $(B,\Omega^\top\widehat AB,\Omega^\top\widehat r)$, with $Y$ no longer needed because $H$ is stored in the codes; the state-level prompt is $(I_n,\widehat A,\widehat r)$. A family of reduced systems is \emph{admissible for the gain} $\Lambda>0$ if
\begin{equation}
 m\,\max\{\|H-I_m\|_{\max},\ \|X\|_{\max}\}<\Lambda\qquad\text{for every member},
 \label{eq:softmax-admissible}
\end{equation}
where $\|\cdot\|_{\max}$ is the largest absolute entry.

\paragraph{Codes and tokens.} Fix the target types $\mathcal T=\{0\}\cup[m]\cup\{(\star,q):q\in[n]\}$, of size $d_k=1+m+n$, and a bijection $\iota:\mathcal T\to[d_k]$ that depends only on $n$ and $m$. Put $b^{(0)}=0$, $b^{(j)}_i=(H-I_m)_{ji}$ for $j\in[m]$, and $b^{(\star,q)}_i=X_{qi}$. For $i\in[m]$ and $t\in\mathcal T$ define the probabilities
\begin{equation}
 \pi^\pm_{i\mid t}=\frac1{2m}\pm\frac{b^{(t)}_i}{2\Lambda},\qquad
 \pi^\pm_{0\mid t}=\frac12\mp\frac{\mathbf 1^\top b^{(t)}}{2\Lambda}.
 \label{eq:softmax-probabilities}
\end{equation}
Source token $i\in\{0,\ldots,m\}$ (index $0$ is a null source) carries two key codes $\kappa_i^\pm\in\mathbb R^{d_k}$ with $(\kappa_i^\pm)_{\iota(t)}=\log\pi^\pm_{i\mid t}$, a tag $\chi_i=\sqrt{d_k}\,e_{\iota(i)}$, and a scalar $u_i$; the query token for state $q$ carries the tag $\chi_{(\star,q)}$ and a scalar $u_\star$. The projections are fixed block selectors, $W_Q^\pm z=\chi$, $W_K^\pm z=\kappa^\pm$, $W_V^\pm z=u$, and the output map writes $\Lambda(a^+-a^-)$ into the scalar of the target, where $a^\pm$ are the two head outputs. The query is masked as a source; all heads read synchronously. Initialize $u_i=g_i$ for $i\in[m]$ and $u_0=u_\star=0$.

\begin{lemma}[Two-head softmax executor]
\label{lem:softmax-executor}
For a family admissible for $\Lambda$, the single set of projections above, which depends only on $n$, $m$, and $\Lambda$, satisfies $(u_{\ell,1},\ldots,u_{\ell,m})=H^\ell g$ and $u_{\ell,\star}(q)=e_q^\top X\sum_{j<\ell}H^jg$ for all $\ell$. Its context retains $(m+2)\bigl(3(n+m+1)+1\bigr)$ reals, at most five times the $(2n+1)(m+1)$ reals of the linear prompt.
\end{lemma}
\begin{proof}
By (\ref{eq:softmax-admissible}), $|b^{(t)}_i|<\Lambda/m$ and $|\mathbf 1^\top b^{(t)}|<\Lambda$, so every probability in (\ref{eq:softmax-probabilities}) is positive, and for each $t$ they sum to one over $i\in\{0,\ldots,m\}$. The scaled score of source $i$ at target $t$ is $(\kappa_i^\pm)^\top\chi_t/\sqrt{d_k}=\log\pi^\pm_{i\mid t}$, so the masked softmax returns exactly $\pi^\pm_{i\mid t}$, and the layer adds $\Lambda\sum_i(\pi^+_{i\mid t}-\pi^-_{i\mid t})u_i=\sum_{i\ge1}b^{(t)}_iu_i$ to the scalar of target $t$, using $u_0=0$. At the null target this is zero, at source target $j$ it is $((H-I)s)_j$, so $s\leftarrow Hs$, and at the query it is $e_q^\top Xs$. Induction and summation give the identities. Each of the $m+1$ source tokens and the query stores two codes, a tag, and a scalar, which is $3d_k+1$ reals; since $m\le n$, $(m+2)(3(n+m+1)+1)\le(m+2)(6n+4)\le5\,(2n+1)(m+1)$, using $(m+2)/(m+1)\le3/2$ and $(6n+4)/(2n+1)\le10/3$.
\end{proof}

\begin{corollary}[Task-uniform gains]
\label{cor:softmax-gains}
\textbf{(a)} For the $\ell_1$-balanced Euclidean Krylov encoder with any budget $m\le n$, on every covered $n$-state batch, every reward, and every $\gamma\in(0,1)$, the family is admissible for $\Lambda_n=2n(n+1)$. The Galerkin contexts are budget-$m$ instances of the same encoder. For the visit-weighted encoder the bound holds with $\Lambda_n$ multiplied by $\sqrt{\max_iN_i/\min_iN_i}$.
\textbf{(b)} The clocked executor is the same layer with $H=I_m$, with the gates $\mathbf 1\{\ell+1=\tau_u\}$ of Definition~\ref{def:executors} as the source scalars $u_u$ (computed by $\operatorname{recip}_0$ as before), and with $X$ the block of held increments. For sample-and-hold with $\|\widehat r\|_\infty\le R$ the family is admissible for $\Lambda_{m,\gamma}=2mR/(1-\gamma)$, which is public when $\gamma$ is. The anchored hold uses both heads in one looped block, writing the same query scalar.
\end{corollary}
\begin{proof}
(a) Balancing gives $\|b_j\|_1=1$, so $\|B\|_{\max}\le1$ and $\|Pb_j\|_\infty\le1$, hence $\|\widehat Ab_j\|_2\le\gamma\sqrt n$. The dual is $\omega_i=b_i/\|b_i\|_2^2$ with $\|b_i\|_2\ge\|b_i\|_1/\sqrt n=1/\sqrt n$, so $|H_{ij}|=|\omega_i^\top\widehat Ab_j|\le\|\widehat Ab_j\|_2/\|b_i\|_2\le\gamma n<n$ and $\|H-I\|_{\max}<n+1$. Then $m\max\{n+1,1\}\le n(n+1)<\Lambda_n$. With weights $w=N$, $\omega_i=D_wb_i/\langle b_i,b_i\rangle_w$ and Cauchy--Schwarz give $|H_{ij}|\le\|D_w^{1/2}\widehat Ab_j\|_2/\|D_w^{1/2}b_i\|_2\le\gamma n\sqrt{\max_iN_i/\min_iN_i}$. (b) With $H=I$ all source targets receive zero, and the query receives $\sum_uu_u\,e_q^\top x_u$, which is the depth-counter update. Each held increment satisfies $\|x_u\|_\infty\le\sum_{t=\tau_{u-1}}^{\tau_u-1}\gamma^t\|\widehat r\|_\infty\le R/(1-\gamma)$, so $m\|X\|_{\max}\le mR/(1-\gamma)<\Lambda_{m,\gamma}$.
\end{proof}

\paragraph{What the construction does and does not show.} The codes $\log\pi^\pm_{i\mid t}$ are written by the encoder, which therefore needs a logarithm as a declared primitive in addition to those of Appendix~\ref{app:scope}. The softmax weights are prescribed by the codes rather than computed from a similarity between learned representations, exactly as in earlier softmax compilations of TD; this is why the main text uses linear attention, whose heads compute the same recurrence from the raw statistics without codes. The gain $\Lambda$ multiplies the rounding error of the probabilities, so the realization is exact in real arithmetic but loses about $\log_{10}\Lambda$ digits in floating point, which Table~\ref{tab:softmax-checks} shows.

\begin{table}[h]
\caption{\textbf{Softmax realization checks} on 60 random covered batches ($4\le n\le24$, $\gamma\in\{0.5,0.8,0.9,0.95\}$, $10\le L\le60$, $\|\widehat r\|_\infty=1$). Errors are maximum absolute deviations divided by $\max_{\ell,s}|\widehat v_\ell(s)|$. The gain ratio is the family quantity in (\ref{eq:softmax-admissible}) divided by the task-uniform gain of Corollary~\ref{cor:softmax-gains}; all 60 batches were admissible.}
\label{tab:softmax-checks}
\centering\small
\begin{tabularx}{\linewidth}{@{}X r@{}}
\toprule
Check & Max\\
\midrule
Exact Krylov context, softmax executor with $\Lambda_n=2n(n+1)$ vs.\ direct TD & $5.3\times10^{-13}$\\
\quad same vs.\ the linear-attention executor on the same context & $5.3\times10^{-13}$\\
Budget-$m$ Galerkin context, softmax vs.\ linear executor & $8.9\times10^{-13}$\\
Sample-and-hold, softmax clocked head with $\Lambda_{m,\gamma}=2m/(1-\gamma)$ vs.\ $\widehat v_{\sigma(\ell)}$ & $6.0\times10^{-14}$\\
Krylov-anchored hold, softmax vs.\ linear executor (family gain from the context) & $1.5\times10^{-14}$\\
Largest gain ratio, Krylov / clocked & $0.11$ / $0.25$\\
\bottomrule
\end{tabularx}
\end{table}
The linear executor reaches $4\times10^{-15}$ on the same contexts (Table~\ref{tab:checks}); the difference is the gain. These checks compare the two-head softmax construction with the linear-attention executor and the reference TD or sample-and-hold trajectory on the same encoded contexts.
\section{Memory Lower Bounds: Real Registers and Bits}
\label{app:memory-proof}
This appendix proves Theorem~\ref{thm:main}(c). The register statement is the following theorem, of which parts (1)--(4) give, respectively, the lower bound $nk+k$, its tightness, the factor-two statement for our context, and the full-rank case. The bit statement is Theorem~\ref{thm:bits} in Appendix~\ref{app:bits}.

Let $\Sigma_k^+$ denote the set of models $(\gamma\widehat P,\widehat r)$ with strictly positive stochastic $\widehat P$ and $k_L=k$. A \textbf{continuous exact representation of size $M$} on a set $\mathcal O\subseteq\Sigma_k^+$ consists of a continuous map $\mathrm{Enc}:\mathcal O\to\mathbb R^M$ and an \emph{arbitrary} map $\mathrm{Dec}$ with $\mathrm{Dec}(\mathrm{Enc}(\widehat A,\widehat r),s,\ell)=\widehat v_\ell(s)$ for all $s\in[n]$ and $\ell\le L$.
\begin{theorem}[Continuous-encoding memory law]
\label{thm:memory-law}
Let $n\ge2$, $1\le k\le\min\{n-1,L-1\}$, and let $\mathcal O$ be relatively open in $\Sigma_k^+$ and contain a model with $\mathbf 1\notin\mathcal K_L$ (such models exist for every such $k$).
\begin{enumerate}
\item \textbf{(Lower bound)} Every continuous exact representation on $\mathcal O$ has $M\ge nk+k$.
\item \textbf{(Tightness)} A continuous exact representation of size exactly $nk+k$ exists on all of $\Sigma_k^+$.
\item \textbf{(Attention)} On each rank stratum, the encoder of Theorem~\ref{thm:main}(b) is continuous in $(\widehat A,\widehat r)$, and its $(2n+1)k$ context reals are within the factor $(2n+1)/(n+1)<2$ of optimal.
\item \textbf{(Full rank)} If $L\ge n+1$, every continuous exact representation on a nonempty relatively open subset of $\Sigma_n^+$ has $M\ge n^2$. The statistics $(\widehat A,\widehat r)$ are therefore optimal up to a factor $(n+1)/n$, and the second stage helps only when $k_L<n$.
\end{enumerate}
\end{theorem}
\emph{Proof idea.} Near a base chain whose Krylov space avoids $\mathbf 1$, we build an $(nk+k)$-parameter family: the increments $\Delta=[\widehat r,\ldots,\widehat A^{k-1}\widehat r]$ and the coefficients $\alpha$ of $\widehat A^k\widehat r$ in this basis vary freely. A stochastic matrix is continuously completed via $P(\theta)=P_0+(T-P_0W)W^+$ with $W=[\Delta,\mathbf 1]$. The trajectory determines $(\Delta,\alpha)$, so $\mathrm{Enc}$ restricted to the family is a continuous injection from an open subset of $\mathbb R^{nk+k}$, and Brouwer's invariance of domain \citep{hatcher2002algebraic} forces $M\ge nk+k$. The remainder of this appendix carries this out.

\paragraph{Scope of the bound.}
In Theorem~\ref{thm:main}(c) an ``open set of covered systems'' means a relatively open subset of $\Sigma_k^+$, as in Theorem~\ref{thm:memory-law}. The bound concerns encoders defined on such a set: it says that no continuous encoder can do with fewer than $nk+k$ registers on \emph{all} models of the set, because the set contains an $(nk+k)$-parameter family whose trajectories are distinct. It is not an instance-level statement, and it does not apply to restricted families on which the trajectory has fewer degrees of freedom. Two examples make the quantifier explicit. If the rewards are restricted to $\widehat r=c\mathbf 1$, then $k_L=1$ and $\widehat v_\ell=c\,\frac{1-\gamma^\ell}{1-\gamma}\mathbf 1$, so a single register holding $c$ suffices when $n$ and $\gamma$ are public. For the block-cycle tasks of Section~\ref{sec:experiments}, every row of $\widehat P$ moves to the next block and the reward is the indicator of block $0$, so $\widehat v_\ell=\sum_{j<\ell}\gamma^j\mathbf 1_{B_{-j}}$ depends on the model only through the block partition; here $\mathbf 1\in\mathcal K_L$ and $\widehat P$ has zero entries, so the family lies outside the hypotheses, and an encoder for this family alone needs only the $n$ block labels. What the theorem does control is a task-uniform encoder: the Krylov context of Theorem~\ref{thm:main}(b) retains $(2n+1)k_L$ registers on every rank-$k_L$ system, and no continuous encoder that works on all dense rank-$k_L$ systems can retain fewer than $nk_L+k_L$ on all of them. This is the sense in which the lower-bound column of Table~\ref{tab:memory} is to be read (Appendix~\ref{app:exp-memory}). Remark~\ref{rem:raw-interface} adds the corresponding caveat on the interface: the bound is stated at the level of the real-valued empirical model, not of a fixed finite batch of records.

\subsection{Proof of the Memory Law}
Throughout, $\gamma\in(0,1)$, $L$, and $n$ are fixed. We write $\Sigma_k^+$ for the models $(\gamma P,r)$ with $P$ row stochastic and entrywise positive and with $\dim\mathcal K_L(\gamma P,r)=k$. Positive $P$ are irreducible and have strictly positive stationary distributions, so they belong to the system class of Section~\ref{sec:approx-context}. The theorem concerns model-level encoders, which is the interface at which the second-stage compression operates (Remark~\ref{rem:raw-interface}).

\begin{lemma}[A base point]
\label{lem:basepoint}
Let $\eta\in(0,1)$ and let $Q$ be the reflecting random walk on a path, with $Q_{i,i\pm1}=\frac12$ and $Q_{11}=Q_{nn}=\frac12$. Put $P_0=(1-\eta)\mathbf 1\mathbf 1^\top/n+\eta Q$ and $\psi_j(i)=\cos(\pi j(i-\frac12)/n)$. Then $P_0$ is entrywise at least $(1-\eta)/n$, $P_0\mathbf 1=\mathbf 1$, and $P_0\psi_j=\eta\cos(\pi j/n)\psi_j$ with $\psi_j\perp\mathbf 1$ for $1\le j\le n-1$. For $1\le k\le n-1$ and $r_0=\sum_{j=1}^k\psi_j$, we have $\mathcal K_L(\gamma P_0,r_0)=\operatorname{span}\{\psi_1,\ldots,\psi_k\}$ whenever $L\ge k$. This space has dimension $k$ and does not contain $\mathbf 1$.
\end{lemma}
\begin{proof}
For interior $i$, the sum-to-product identity gives $(Q\psi_j)(i)=\cos(\pi j/n)\psi_j(i)$. At the two boundaries the reflection gives the same identity, using $\cos(\pi j-x)=(-1)^j\cos x$. The matrix $Q$ is symmetric with distinct eigenvalues $\cos(\pi j/n)$, so $\psi_j\perp\psi_0=\mathbf 1$ and $\mathbf 1\mathbf 1^\top\psi_j=0$. The Krylov matrix is $[\psi_1,\ldots,\psi_k]V$, where $V$ is the Vandermonde matrix of the distinct nodes $\gamma\eta\cos(\pi j/n)$ and is therefore invertible. The resulting space is orthogonal to $\mathbf 1\neq0$.
\end{proof}

\begin{lemma}[An $(nk+k)$-parameter family]
\label{lem:family}
Let $(\gamma P_0,r_0)\in\Sigma_k^+$ with $\mathbf 1\notin\mathcal K_L$ and $k\le L-1$. Put $A_0=\gamma P_0$ and $\Delta_0=[r_0,A_0r_0,\ldots,A_0^{k-1}r_0]$, and choose $\alpha_0$ with $A_0^kr_0=\Delta_0\alpha_0$; it exists because $\mathcal K_L=\mathcal K_k$ is invariant. For $\theta=(\Delta,\alpha)\in\mathbb R^{n\times k}\times\mathbb R^k$, let $C(\alpha)$ be the companion matrix ($Ce_i=e_{i+1}$ for $i<k$, $Ce_k=\alpha$) and set
\[
 W=[\Delta,\mathbf 1],\qquad T=[\gamma^{-1}\Delta C(\alpha),\,\mathbf 1],\qquad P(\theta)=P_0+(T-P_0W)W^+,\qquad r(\theta)=\Delta e_1 .
\]
There is a neighborhood $\mathcal N$ of $\theta_0=(\Delta_0,\alpha_0)$ on which the following hold:
(a) $P(\theta)$ is continuous in $\theta$, entrywise positive, and row stochastic;
(b) $\gamma P(\theta)\Delta=\Delta C(\alpha)$;
(c) $\widehat A^jr=\Delta C(\alpha)^je_1$ for all $j$, so in particular $\widehat A^jr=\Delta e_{j+1}$ for $j<k$ and $\widehat A^kr=\Delta\alpha$;
(d) $(\gamma P(\theta),r(\theta))\in\Sigma_k^+$.
\end{lemma}
\begin{proof}
At $\theta_0$, $W_0$ has full column rank because $\Delta_0$ does and $\mathbf 1\notin\operatorname{col}\Delta_0$. Full column rank is an open condition, so near $\theta_0$ the map $W^+=(W^\top W)^{-1}W^\top$ is continuous and $W^+W=I$. Hence $P(\theta)W=T$, which gives (b) and $P\mathbf 1=\mathbf 1$. At $\theta_0$, $A_0\Delta_0=\Delta_0C(\alpha_0)$ and $P_0\mathbf 1=\mathbf 1$, so $T_0=P_0W_0$ and $P(\theta_0)=P_0>0$; positivity persists nearby. Part (c) follows from (b) by induction. For (d), $\mathcal K_L\subseteq\operatorname{col}\Delta$ by (c), and $\mathcal K_L$ contains $\Delta e_1,\ldots,\Delta e_k$ because $k\le L$; $\Delta$ keeps full column rank nearby.
\end{proof}

\begin{proof}[Proof of Theorem~\ref{thm:memory-law}]
\textbf{(1)} Take a model in $\mathcal O$ with $\mathbf 1\notin\mathcal K_L$ and let $\iota(\theta)=(\gamma P(\theta),r(\theta))$ be the family of Lemma~\ref{lem:family}. Since $\iota$ is continuous, $\iota(\theta_0)\in\mathcal O$, $\iota(\mathcal N)\subseteq\Sigma_k^+$, and $\mathcal O$ is relatively open, we may shrink $\mathcal N$ so that $\iota(\mathcal N)\subseteq\mathcal O$. By Lemma~\ref{lem:family}(c), the trajectory $\Phi(\theta)=(\widehat v_\ell(s))_{s,\ell\le L}$ determines the increments $\widehat v_{j+1}-\widehat v_j=\Delta e_{j+1}$ for $j<k$ and $\widehat v_{k+1}-\widehat v_k=\Delta\alpha$ (using $L\ge k+1$). Hence it determines $\Delta$ and, since $\Delta$ has full column rank, $\alpha$. So $\Phi$ is injective on $\mathcal N$. Because $\mathrm{Dec}(\mathrm{Enc}(\iota(\theta)),\cdot,\cdot)=\Phi(\theta)$, the map $F=\mathrm{Enc}\circ\iota$ is a continuous injection of the open set $\mathcal N\subseteq\mathbb R^{nk+k}$ into $\mathbb R^M$. If $M<nk+k$, compose with the inclusion $x\mapsto(x,0)\in\mathbb R^{nk+k}$. The result is a continuous injection between open subsets of the same dimension, so its image is open by invariance of domain \citep{hatcher2002algebraic}. But the image lies in the hyperplane $\mathbb R^M\times\{0\}$, which has empty interior, a contradiction.

\textbf{(2)} On $\Sigma_k^+$ set $\mathrm{Enc}(\widehat A,\widehat r)=(\Delta,\alpha)$ with $\Delta=[\widehat r,\ldots,\widehat A^{k-1}\widehat r]$ and $\alpha=(\Delta^\top\Delta)^{-1}\Delta^\top\widehat A^k\widehat r$. Before saturation the Krylov spaces grow strictly, so $\Delta$ has full column rank on $\Sigma_k$ and $\alpha$ is continuous. Since $L\ge k+1$, $\mathcal K_{k+1}=\mathcal K_k$, so $\widehat A^k\widehat r=\Delta\alpha$ and $\widehat A\Delta=\Delta C(\alpha)$. Then $\mathrm{Dec}(\cdot,s,\ell)=e_s^\top\Delta\sum_{j<\ell}C(\alpha)^je_1$ is exact.

\textbf{(3)} Continuity is Lemma~\ref{lem:continuity}, and exactness is Theorem~\ref{thm:main}. Each source token stores $b_j,y_j,g_j$, i.e., $2n+1$ reals, and $(2n+1)k/(nk+k)=(2n+1)/(n+1)$.

\textbf{(4)} The stratum $\Sigma_n^+$ is nonempty: take $P_0$ from Lemma~\ref{lem:basepoint} and $r_0=\sum_{j=0}^{n-1}\psi_j$, whose $n$ eigenvalues are distinct. Fix any model $(\gamma P_1,r_1)$ in the given open set. Parameterize positive stochastic matrices near $P_1$ by their first $n-1$ columns ($n(n-1)$ parameters) and rewards near $r_1$ ($n$ parameters). The rank-$n$ condition is open. The trajectory determines $r=\widehat v_1$ and $\widehat A=[\delta_1,\ldots,\delta_n][\delta_0,\ldots,\delta_{n-1}]^{-1}$ with $\delta_j=\widehat v_{j+1}-\widehat v_j$, so it is injective in these $n^2$ parameters. The argument of (1) gives $M\ge n^2$.
\end{proof}

\begin{remark}[The case $L=k$ and gauge counting]
If $L=k$, fix $\alpha=\alpha_0$; the same argument gives $M\ge nk$, and storing $\Delta$ attains it. A numerical realization $(B,H,g)$ has $nk+k^2+k$ entries but carries a $\mathrm{GL}(k)$ gauge freedom $(B,H,g)\mapsto(BG,G^{-1}HG,G^{-1}g)$. Removing its $k^2$ dimensions leaves exactly $nk+k$, so the bound is tight modulo gauge.
\end{remark}

\begin{remark}[Why continuity, and the raw-data interface]
\label{rem:raw-interface}
Without continuity, a set-theoretic bijection $\mathbb R^d\to\mathbb R$ (e.g., interleaving decimal digits) gives $M=1$. Linear decoders exclude this pathology by fiat, whereas continuity of the encoder excludes it naturally and permits arbitrary decoders. Empirical models from finite batches have rational entries, so continuity is stated at the model level, where the second stage operates. If one instead fixes the observed transitions and lets only rewards vary continuously, the same argument gives only $M\ge k$. Lower bounds for record-level interfaces need counting arguments such as the bit law of Appendix~\ref{app:bits}, which is consistent with Theorem~\ref{thm:memory-law}: each real carries $\Theta(\log(1/\epsilon))$ bits.
\end{remark}

\subsection{Offline Retained-Bit Complexity}
\label{app:bits}
This subsection concerns unrestricted offline codes, not the context of Theorem~\ref{thm:main}(b). It complements Theorem~\ref{thm:memory-law}: there, reals are counted under continuity; here, bits are counted with arbitrary encoders and decoders.
\begin{definition}[Offline coding problem]
For public $n$, $L\ge1$, $\epsilon>0$, and $\gamma\in(0,1)$, let $\mathfrak M_n=\{(P,r):P\ge0,\ P\mathbf 1=\mathbf 1,\ \|r\|_\infty\le1\}$. A $B$-bit code consists of $\operatorname{Enc}:\mathfrak M_n\to\{0,1\}^B$ and a decoder satisfying $\max_s\sup_{\ell\le L}|\operatorname{Dec}(\operatorname{Enc}(P,r),s,\ell)-e_s^\top\sum_{j<\ell}(\gamma P)^jr|\le\epsilon$. Let $B^*_{n,L}(\epsilon,\gamma)$ be the minimum such $B$.
\end{definition}
\begin{theorem}[Retained-bit law]
\label{thm:bits}
For fixed $\gamma$ and sufficiently small $\epsilon$, uniformly in $n$ and $L$ (including $L=\infty$), $B^*_{n,L}(\epsilon,\gamma)=\Theta_\gamma(n\min\{n,L,\ln(1/\epsilon)\}\ln(1/\epsilon))$.
\end{theorem}
\begin{proof}[Proof sketch with explicit $\gamma$-dependence]
Let $t=\ln(1/\epsilon)$, $a=\ln(1/\gamma)$, and $V=(1-\gamma)^{-1}$.
\emph{Upper bound.} Let $J_0=\lceil\ln(4V/\epsilon)/a\rceil$ and $J=\min\{L,J_0\}$. The \emph{prefix code} quantizes $\widehat v_1,\ldots,\widehat v_J$ to accuracy $\epsilon/4$ with $\lceil\log_2(3+4V/\epsilon)\rceil$ bits per coordinate and holds $\widehat v_J$ afterwards, since the tail satisfies $V\gamma^J\le\epsilon/4$. The \emph{model code} quantizes rewards to accuracy $\epsilon(1-\gamma)/4$ and each transition row to $\ell_1$-accuracy $\epsilon(1-\gamma)^2/(4\gamma)$ using integer rows summing to $M$, since $\|p-k/M\|_1\le n/M$. The perturbation bound $\sup_\ell\|\widetilde v_\ell-v_\ell\|_\infty\le\delta_r/(1-\gamma)+\gamma\delta_P/(1-\gamma)^2$ then gives accuracy $\epsilon/2$. Using the prefix code if $J<n$ and the model code otherwise yields $O(n\min\{n,J\}\ln(V/\epsilon))$ bits.
\emph{Lower bound.} Packing rewards with $P=I$ gives $n\log_2(1/(4\epsilon))$ bits. For the delayed-reward family, use $p=n-h-1$ source states, a deterministic delay chain $c_h\to\cdots\to c_1\to c_0$ rewarded at $c_1$, and transition probabilities $\theta_{ij}\in[0,1/(2h)]$ from source $i$ into $c_j$. This family satisfies $\widehat v_{j+1}(u_i)-\widehat v_j(u_i)=\gamma^j\theta_{ij}$. Choose $h=\min\{\lfloor(n-1)/2\rfloor,L-1,\lfloor t/(8(1+a))\rfloor\}$ and pack each $\theta_{ij}$ on a grid of spacing $8\epsilon\gamma^{-j}$. For $\epsilon\le\exp(-\max\{16,8(1+a)\})$, this gives $B^*\ge n\min\{n,L,t\}t/(96(1+a)\ln2)$.
\end{proof}
\begin{remark}[Dependence on $\gamma$]
The constants are explicit. The upper bound's effective depth is $J_0\approx t/(1-\gamma)$ as $\gamma\to1$, whereas the lower bound's delay length is $h\approx t/8$. For $\ln(1/\epsilon)<\min\{n,L\}$, the two therefore differ by a factor of order $1/(1-\gamma)$ in the depth term. Closing this factor, the analogue of Theorem~\ref{thm:gamma-law} for bits, is open. The law ranges over all real stochastic models; a fixed finite batch has finite information, which can saturate as $\epsilon\downarrow0$. For comparison, coding only the fixed point $(I-\gamma P)^{-1}r$ needs $\Theta_\gamma(n\ln(1/\epsilon))$ bits.
\end{remark}

\section{Limited Context: Proofs, the Krylov-Anchored Hold, and Instance-Dependent Rates}
\label{app:gamma-law-proof}
This appendix proves Theorem~\ref{thm:gamma-law} and states and proves the results referred to in Section~\ref{sec:approx-context}: the suboptimality of Krylov moment matching, the Krylov-anchored hold, and instance-dependent rates. Throughout, an $m$-dimensional context is one of Definition~\ref{def:context}, and $\mathfrak C_m$ is the corresponding class of trajectories.

\subsection{Statements}
\label{app:anchored}
\begin{proposition}[Moment matching is not minimax]
\label{prop:mm-suboptimal}
\textbf{(a)} Krylov truncation, i.e., the terminal-corrected context that equals TD stopped after $m$ steps, has worst-case error exactly $\gamma^m/(1-\gamma)$ over systems with $k_L>m$. This exceeds $e^2\Psi_m$ by the factor $\min\{m,1/(1-\gamma)\}/e^2$.
\textbf{(b)} Krylov projection (the Galerkin context $\widetilde v_\ell=B\sum_{j<\ell}H^jg$ obtained from the budget-$m$ encoder with Euclidean or visit-weighted Arnoldi) has error at least $0.657\,\gamma^m/\sqrt{1-\gamma}\ge 0.089\,e^2\Psi_m/\sqrt{1-\gamma}$ on a directed ring with one rewarded state, whenever $\gamma\ge\frac12$ and $m\ge1/(1-\gamma)$.
\end{proposition}
Both encoders spend their whole budget on the \emph{first} $m$ increments, whereas the optimal context must also represent late ones. The loss comes from moment matching rather than from time invariance: on the ring of (b), numerically optimized time-invariant recurrences come within a factor $1.8$ of the certified lower bound, closer than sample-and-hold itself (Appendix~\ref{app:ti-probe}); whether a uniformly computable time-invariant encoder attains $\Theta(\Psi_m)$ remains open. Moment matching does, however, adapt to spectral structure that sample-and-hold ignores, which motivates the following hybrid.

\paragraph{Krylov-anchored hold.}
Split $m=k+j$ with $j=0$ if $\gamma<\frac12$ and $j=\min\{J,\lfloor m/4\rfloor\}$ otherwise, where $J=\lfloor1/(1-\gamma)\rfloor$. Run $k$ rounds of visit-weighted Arnoldi (Appendix~\ref{app:weighted}), and let $\widetilde v^{\rm G}$ be the resulting projected trajectory. With the depth-counter executor, hold the residual $e_\ell=\widehat v_\ell-\widetilde v^{\rm G}_\ell$ at the $j$ depths $k+i_s$ of $\mathcal T_j(\gamma)$. The output is $\widetilde v_\ell=\widetilde v^{\rm G}_\ell+e_{\sigma(\ell)}$, with $e_{\sigma(\ell)}=0$ before the first sample. One looped block contains both the TD heads and the depth-counter heads (Appendix~\ref{app:anchored-encoder}).
\begin{theorem}[Minimax optimality and instance adaptivity]
\label{thm:anchored}
Suppose the encoder weights $w$ satisfy $\|\widehat Px\|_w\le\|x\|_w$ (e.g., $w=\mu$, or visit counts of a balanced batch). Let $\mathcal E^{\rm G}_k=\sup_\ell\|\widehat v_\ell-\widetilde v^{\rm G}_\ell\|_w/\|\widehat r\|_w$. Then the Krylov-anchored hold is an $m$-dimensional context with $k+j=m$ and satisfies
\[
 \sup_{\ell\ge0}\frac{\|\widehat v_\ell-\widetilde v_\ell\|_w}{\|\widehat r\|_w}\ \le\ \min\bigl\{2e^2\,\Psi_m(\gamma),\ 2\,\mathcal E^{\rm G}_{k}\bigr\},\qquad k\ge\lceil 3m/4\rceil .
\]
Its retained memory is $(2n+1)(k+1)+j(n+1)\le(2n+1)(m+1)$, and it is exact whenever $k_L\le k$.
\end{theorem}
\begin{proposition}[Instance-dependent rates and robustness]
\label{prop:instance}
\textbf{(a) Reversible chains.} If $N_{ij}=N_{ji}$, then visit-weighted Krylov projection with context dimension $m$ satisfies $\mathcal E^{\rm G}_m\le 4\beta^m/((1-\beta)\sqrt{1-\gamma^2})\le 8\beta^m/(1-\gamma^2)$ with $\beta=\gamma/(1+\sqrt{1-\gamma^2})$. The anchored hold is therefore also at most $16\beta^{\lceil3m/4\rceil}/(1-\gamma^2)$.
\textbf{(b) Numerical rank.} Let $\sigma_i$ be the singular values of $V_L=D_w^{1/2}[\widehat v_1,\ldots,\widehat v_L]$. Every $m$-dimensional context in $\mathfrak C_m$ has $\sup_{\ell\le L}\|\widehat v_\ell-\widetilde v_\ell\|_w\ge\sigma_{m+1}/\sqrt L$, while the best $m$-dimensional subspace achieves $\sigma_{m+1}$. Moreover, $|\sigma_i(V_L)-\sigma_i(V'_L)|\le\sqrt L\sup_\ell\|\widehat v_\ell-\widehat v'_\ell\|_w$, and the latter is $O\bigl(\delta_r/(1-\gamma)+\delta_A\|\widehat r\|_w/(1-\gamma)^2\bigr)$ under model perturbations of size $(\delta_A,\delta_r)$.
\end{proposition}
Part (a) is a square-root acceleration: the effective horizon becomes $1/\sqrt{1-\gamma}$ instead of $1/(1-\gamma)$, because the Chebyshev coefficients of every partial sum $\sum_{j<\ell}x^j$ on $[-\gamma,\gamma]$ are nonnegative and dominated by those of $1/(1-x)$. Part (b) replaces the discontinuous exact rank by the stable $\varepsilon$-rank $\#\{i:\sigma_i>\varepsilon\sqrt L\}$, a lower bound on the context dimension that \emph{any} method needs for accuracy $\varepsilon$; it is what makes compression meaningful on sampled tasks whose exact rank is full (Appendix~\ref{app:exp-accuracy}). A threshold gate on Arnoldi residuals gives a certified adaptive rank as well (Appendix~\ref{app:threshold}).

\subsection{Three Tools}
\begin{lemma}[Contraction]
\label{lem:contraction}
If $P$ is row stochastic and $\mu>0$ satisfies $\mu^\top P=\mu^\top$, then $\|Px\|_\mu\le\|x\|_\mu$. For every row-stochastic $P$, $\|Px\|_\infty\le\|x\|_\infty$.
\end{lemma}
\begin{proof}
By Jensen's inequality, $(\sum_jP_{ij}x_j)^2\le\sum_jP_{ij}x_j^2$. Weighting by $\mu_i$ and summing gives $\sum_j\mu_jx_j^2$. The sup-norm statement is immediate.
\end{proof}

\begin{lemma}[Error controls block-increment width]
\label{lem:block-width}
If $\widetilde v\in\mathfrak C_m$ and $\sup_\ell\|\widehat v_\ell-\widetilde v_\ell\|\le\varepsilon$, then $U=\operatorname{col}(D)$ has $\dim U\le m$ and satisfies $\operatorname{dist}(\widehat v_b-\widehat v_a,U)\le2\varepsilon$ for all $0\le a<b$.
\end{lemma}
\begin{proof}
$\widetilde v_b-\widetilde v_a=D\sum_{a\le j<b}s_j\in U$, and the triangle inequality gives $\|(\widehat v_b-\widehat v_a)-(\widetilde v_b-\widetilde v_a)\|\le2\varepsilon$.
\end{proof}

\begin{lemma}[Width of orthogonal geometric families]
\label{lem:waterfill}
Let $x_0,\ldots,x_{T-1}$ be orthogonal in an inner-product norm with $\|x_t\|=a_0\rho^t$, $\rho\in(0,1)$, and $T\ge m+1+1/(1-\rho^2)$. Then every subspace $U$ of dimension at most $m$ satisfies $\max_t\operatorname{dist}(x_t,U)\ge a_0\rho^m\max\{\sqrt{1-\rho^2},\rho^{1/(1-\rho^2)}\}\ge a_0\rho^m/2$.
\end{lemma}
\begin{proof}
Let $\Pi$ be the orthogonal projection onto $U$, and suppose all distances are at most $\varepsilon$. Put $\eta=\varepsilon/a_0$. The unit vectors $\hat x_t$ are orthonormal, so $\sum_t\|\Pi\hat x_t\|^2\le\operatorname{tr}\Pi\le m$. Each term also satisfies $\|\Pi\hat x_t\|^2=1-\operatorname{dist}(x_t,U)^2/\|x_t\|^2\ge1-\eta^2\rho^{-2t}$. Summing over $t\le m$ gives $m\ge m+1-\eta^2\sum_{t\le m}\rho^{-2t}$, hence $\eta\ge\rho^m\sqrt{1-\rho^2}$. Next let $a=1/(1-\rho^2)$ and $T'=\lceil m+a\rceil\le T$. If $\eta\le\rho^{T'-1}$, then
$\sum_{t<T'}(1-\eta^2\rho^{-2t})=T'-\eta^2\frac{\rho^{-2T'}-1}{\rho^{-2}-1}>T'-\eta^2\rho^{-2(T'-1)}a\ge T'-a\ge m$,
a contradiction. Thus $\eta>\rho^{T'-1}\ge\rho^{m+a}$. For the final inequality: if $\rho^2\le3/4$ then $\sqrt{1-\rho^2}\ge1/2$. Otherwise $u=1-\rho^2<1/4$, and $\ln(1-u)\ge-u/(1-u)$ gives $\rho^{1/u}=\exp(\ln(1-u)/(2u))\ge e^{-2/3}>1/2$.
\end{proof}

\subsection{The Lower Bound $\mathfrak E_m\ge\Psi_m/64$}
Both instances are directed rings, $(Pv)(x)=v(x+1)$ modulo $n$. The empirical model is realized exactly by a covered batch containing one transition per state, and the stationary distribution $\mu$ is uniform.

\paragraph{Instance I (single rewarded state).} Let $r=\sqrt n\,e_0$, so $\|r\|_\mu=1$. The increments $\widehat A^tr=\gamma^t\sqrt n\,e_{-t}$ ($t<n$) are orthogonal with norms $\gamma^t$. If $n\ge m+1+1/(1-\gamma^2)$, Lemmas~\ref{lem:block-width} (with $b=a+1$) and~\ref{lem:waterfill} give $2\varepsilon\ge\gamma^m/2$, i.e., every $\widetilde v\in\mathfrak C_m$ has error at least $\gamma^m/4$.

\paragraph{Instance II (rewarded arc).} For integers $w,T\ge1$, let $n=(2T+2)w$ and $r=\sqrt{n/w}\,\mathbf 1_{[0,w)}$, so $\|r\|_\mu=1$. The block increments $\tau_i=\widehat v_{(i+1)w}-\widehat v_{iw}=\sum_{j=iw}^{(i+1)w-1}\gamma^jP^jr$ are supported on arcs of length $2w-1$, because $P^jr$ is supported on $[-j,-j+w)$. The even blocks $\tau_{2t}$, $t<T$, therefore have pairwise disjoint supports inside the ring and are orthogonal. Moreover, $\tau_{2t}$ is a shifted copy of $\gamma^{2tw}\tau_0$. With $S_w=\sum_{j<w}\gamma^j$ and $\mathcal S=\operatorname{supp}\tau_0$, Cauchy--Schwarz gives
\[
 \|\tau_0\|_\mu\ \ge\ \frac{\langle\tau_0,\mathbf 1_{\mathcal S}\rangle_\mu}{\|\mathbf 1_{\mathcal S}\|_\mu}
 =\frac{n^{-1}\sqrt{n/w}\;wS_w}{\sqrt{(2w-1)/n}}=S_w\sqrt{\frac{w}{2w-1}}\ \ge\ \frac{S_w}{\sqrt2}.
\]
Applying Lemma~\ref{lem:waterfill} with $a_0=\|\tau_0\|_\mu$, $\rho=\gamma^{2w}$, and $T\ge m+1+1/(1-\gamma^{4w})$, together with Lemma~\ref{lem:block-width} for $(a,b)=(2tw,(2t+1)w)$, shows that every $\widetilde v\in\mathfrak C_m$ has error at least $S_w\gamma^{2wm}/(4\sqrt2)$.

\begin{proof}[Proof of the lower bound in Theorem~\ref{thm:gamma-law}]
Let $q=1/(m(1-\gamma))$, so that $\Psi_m=\gamma^m\max\{1,q\}\le\max\{1,q\}$.
If $q\le16$, then $\Psi_m\le16\gamma^m$, and Instance~I gives error at least $\gamma^m/4\ge\Psi_m/64$.
If $q>16$, then $1-\gamma<1/16$. Take $w=\lfloor q/4\rfloor\ge4$, so that $w>q/4-1>3q/16$, $w(1-\gamma)\le1/(4m)\le1/4$, and $2wm(1-\gamma)\le1/2$. Bernoulli's inequality gives $S_w\ge\sum_{j<w}(1-j(1-\gamma))\ge w(1-w(1-\gamma)/2)\ge7w/8$. From $\ln\gamma\ge-(1-\gamma)/\gamma\ge-\frac{16}{15}(1-\gamma)$ we get $\gamma^{2wm}\ge e^{-8/15}$. Instance~II then gives error at least
\[
 \frac{1}{4\sqrt2}\cdot\frac78\cdot\frac{3q}{16}\cdot e^{-8/15}\ \ge\ \frac{q}{59}\ \ge\ \frac{\Psi_m}{59}.
\]
Both cases give at least $\Psi_m/64$. Taking the infimum over $\widetilde v\in\mathfrak C_m$ and the supremum over systems proves the claim. Both instances have $k_L>m$, so the bound also holds for systems below the exact threshold.
\end{proof}

\subsection{The Upper Bound and the Sample-and-Hold Context}
\begin{lemma}[Sample-and-hold error]
\label{lem:sh}
Let $\|\cdot\|$ be any norm with $\|\widehat Px\|\le\|x\|$. For the schedule with parameters $(k,j)$, every $\ell\ge0$ satisfies $\|\widehat v_\ell-\widehat v_{\sigma(\ell)}\|<\gamma^k\|\widehat r\|/((j+1)(1-\gamma))$.
\end{lemma}
\begin{proof}
For $\ell\le k$ the error is zero. Let $\ell=k+i$ and $s=\max\{s:i_s\le i\}$ (with $i_0=0$), so that $\sigma(\ell)=k+i_s$. Then $\|\widehat v_\ell-\widehat v_{\sigma(\ell)}\|\le\sum_{t=k+i_s}^{k+i-1}\gamma^t\|\widehat r\|=\gamma^k(\gamma^{i_s}-\gamma^i)\|\widehat r\|/(1-\gamma)$. By definition $1-\gamma^{i_s}\ge s/(j+1)$. If $s<j$, then $i<i_{s+1}$ gives $1-\gamma^i<(s+1)/(j+1)$; if $s=j$, then $1-\gamma^i<1$. In both cases $\gamma^{i_s}-\gamma^i<1/(j+1)$.
\end{proof}

\begin{proof}[Proof of the upper bound in Theorem~\ref{thm:gamma-law}]
The increments $x_u$ span at most $m$ dimensions and the executor gives $D=X$, so the sample-and-hold output lies in $\mathfrak C_m$. It remains to bound $F=\gamma^k/((j+1)(1-\gamma))$ by $e^2\Psi_m$; Lemma~\ref{lem:contraction} supplies the required norms. If $\gamma<\frac12$, then $F=\gamma^m/(1-\gamma)\le2\gamma^m\le2\Psi_m$. If $\gamma\ge\frac12$ and $m\le J$, then $q\ge1$, and $F<1/(m(1-\gamma))=\gamma^{-m}\Psi_m\le\gamma^{-1/(1-\gamma)}\Psi_m\le e^2\Psi_m$, using $-\ln\gamma\le(1-\gamma)/\gamma\le2(1-\gamma)$. Otherwise $m\ge J+1>1/(1-\gamma)$, so $\Psi_m=\gamma^m$, and $F<\gamma^{m-J}\le\gamma^{-J}\Psi_m\le e^2\Psi_m$.
\end{proof}

\begin{remark}[The schedule must depend on $\gamma$]
\label{rem:dyadic}
For constant reward $\widehat r=\mathbf 1$, $\widehat v_\ell=\frac{1-\gamma^\ell}{1-\gamma}\mathbf 1$, and any schedule $\mathcal T$ has error at least $\gamma^{\max\mathcal T}/(1-\gamma)$ as $\ell\to\infty$. If $\mathcal T$ is fixed independently of $\gamma$ and $\gamma\ge\frac12$ is close enough to one that $\max\mathcal T\le1/(1-\gamma)$, this is at least $e^{-2}/(1-\gamma)\ge e^{-2}m\Psi_m$. Dyadic schedules $\{1,2,4,\ldots\}$ and uniform schedules are therefore off by a factor $\Omega(m)$. The quantile schedule places samples where the remaining discounted mass decreases by equal fractions.
\end{remark}

\subsection{Finite Depth}
\label{app:finite-depth}
This subsection states and proves the finite-depth version of Theorem~\ref{thm:gamma-law} summarized in Section~\ref{sec:approx-context}. Throughout, $\mathfrak C_m$ is the depth-$L$ class of Definition~\ref{def:context} and the error of a context is $\max_{0\le\ell\le L}\|\widehat v_\ell-\widetilde v_\ell\|_\mu/\|\widehat r\|_\mu$, so that $\mathfrak E_{m,L}(\gamma)$ is the worst case of its infimum over $\mathfrak C_m$. Lemma~\ref{lem:block-width} applies verbatim to block increments with $b\le L$, since its proof uses only the errors at depths $a$ and $b$.
\begin{proposition}[Finite depth]
\label{prop:finite-depth}
For all $m\ge1$, $\gamma\in(0,1)$ and $L\ge1$, with $\Psi_{m,L}(\gamma):=\gamma^m\max\{1,(1-\gamma^L)/(m(1-\gamma))\}\le\Psi_m(\gamma)$:
\textbf{(a)} $\mathfrak E_{m,L}(\gamma)\le\mathfrak E_m(\gamma)$, and $\mathfrak E_{m,L}(\gamma)=0$ if and only if $m\ge L$.
\textbf{(b)} $\mathfrak E_{m,L}(\gamma)\le e^2\Psi_{m,L}(\gamma)$, attained by sample-and-hold with sampled depths at the quantiles of the discounted mass through depth $L$.
\textbf{(c)} $\mathfrak E_{m,L}(\gamma)\ge\gamma^m/(2\sqrt{m+1})$ if $L\ge m+1$, and $\mathfrak E_{m,L}(\gamma)\ge\Psi_m(\gamma)/64$ if $L\ge m+4/(1-\gamma)$.
\textbf{(d)} If $\gamma\ge\tfrac12$ and $8m\le L\le1/(1-\gamma)$, then $L/(400m)\le\mathfrak E_{m,L}(\gamma)\le e^2L/m$.
\end{proposition}
\begin{proof}
\emph{(a)} Restricting a context of depth $L'\ge L$, or of unbounded depth, to the depths $\ell\le L$ gives a context of depth $L$ with the same decoder, and a maximum over fewer depths is smaller; hence $\mathfrak E_{m,L}$ is nondecreasing in $L$ and $\mathfrak E_{m,L}\le\mathfrak E_m$. If $m\ge L$, the Krylov context of Theorem~\ref{thm:main}(b) is exact with $k_L\le L\le m$ tokens (padded with zero tokens), so $\mathfrak E_{m,L}=0$. If $m<L$, part (c) gives $\mathfrak E_{m,L}\ge\gamma^m/(2\sqrt{m+1})>0$.

\emph{(b)} Keep the split $(k,j)$ of Section~\ref{sec:approx-context}; if $k\ge L$ the context is exact. Otherwise replace the sampled depths by the quantiles of the discounted mass through depth $L$, $\mathcal T_{m,L}(\gamma)=\{1,\ldots,k\}\cup\{k+i_s:1\le s\le j\}$ with $i_s=\min\{i\ge0:1-\gamma^i\ge\frac{s}{j+1}(1-\gamma^{L-k})\}$, so that $i_j\le L-k$; the encoder and executor are those of Appendix~\ref{app:sh-encoder} with these stamps, and the output lies in $\mathfrak C_m$ as in the proof of Theorem~\ref{thm:gamma-law}. For $\ell\le k$ the error is zero. For $\ell=k+i\le L$ let $s=\max\{s:i_s\le i\}$; as in Lemma~\ref{lem:sh}, $\|\widehat v_\ell-\widehat v_{\sigma(\ell)}\|_\mu\le\gamma^k(\gamma^{i_s}-\gamma^i)\|\widehat r\|_\mu/(1-\gamma)$, and $\gamma^{i_s}-\gamma^i=(1-\gamma^i)-(1-\gamma^{i_s})\le(1-\gamma^{L-k})/(j+1)$: for $s<j$ because $1-\gamma^i<\frac{s+1}{j+1}(1-\gamma^{L-k})$, and for $s=j$ because $i\le L-k$. Hence the error is at most $F_L:=\gamma^k(1-\gamma^{L-k})/((j+1)(1-\gamma))\le\gamma^k(1-\gamma^L)/((j+1)(1-\gamma))$, and the three cases of the proof of Theorem~\ref{thm:gamma-law} give $F_L\le e^2\Psi_{m,L}$: for $\gamma<\frac12$, $F_L\le\gamma^m/(1-\gamma)\le2\gamma^m\le2\Psi_{m,L}$; for $\gamma\ge\frac12$ and $m\le J$, $F_L<\frac{1-\gamma^L}{m(1-\gamma)}=\gamma^{-m}\cdot\gamma^m\frac{1-\gamma^L}{m(1-\gamma)}\le\gamma^{-J}\Psi_{m,L}\le e^2\Psi_{m,L}$; and for $m\ge J+1$, $F_L\le\gamma^{m-J}/((J+1)(1-\gamma))<\gamma^{-J}\gamma^m\le e^2\Psi_{m,L}$.

\emph{(c)} Let $L\ge m+1$ and take Instance~I with $n\ge m+2$ states. The single-step increments $x_t=\widehat A^tr=\gamma^t\sqrt n\,e_{-t}$, $t=0,\ldots,m$, are orthogonal with $\|x_t\|_\mu=\gamma^t$, and each is a block increment with $b=t+1\le L$. Let a context have error $\varepsilon$, let $U=\operatorname{col}(D)$ with $\dim U\le m$, let $\Pi$ be the orthogonal projection onto $U$, and let $\hat x_t$ be the normalized vectors. Then $\sum_{t\le m}\|\Pi\hat x_t\|^2\le\operatorname{tr}\Pi\le m$, so
\[
 \sum_{t\le m}\operatorname{dist}(x_t,U)^2=\sum_{t\le m}\gamma^{2t}\bigl(1-\|\Pi\hat x_t\|^2\bigr)\ \ge\ \gamma^{2m}\sum_{t\le m}\bigl(1-\|\Pi\hat x_t\|^2\bigr)\ \ge\ \gamma^{2m},
\]
hence $\max_t\operatorname{dist}(x_t,U)\ge\gamma^m/\sqrt{m+1}$, and Lemma~\ref{lem:block-width} gives $2\varepsilon\ge\gamma^m/\sqrt{m+1}$.
For the second claim, the proof of the lower bound of Theorem~\ref{thm:gamma-law} uses only block increments $\widehat v_b-\widehat v_a$ with $b\le L_0$: for Instance~I, $L_0=n=\lceil m+1+1/(1-\gamma^2)\rceil\le m+2+1/(1-\gamma)$; for Instance~II, $L_0=(2T-1)w$ with $T=\lceil m+1+1/(1-\gamma^{4w})\rceil$, $w=\lfloor q/4\rfloor$ and $q=1/(m(1-\gamma))>16$, which forces $\gamma>\frac12$. Since $x:=4w(1-\gamma)\le q(1-\gamma)=1/m\le1$ and $\gamma^{4w}\le e^{-x}$, we have $1-\gamma^{4w}\ge1-e^{-x}\ge x/2$, so $T\le m+2+1/(2w(1-\gamma))$ and
\[
 (2T-1)w\ \le\ (2m+3)w+\frac1{1-\gamma}\ \le\ \frac{2m+3}{4m(1-\gamma)}+\frac1{1-\gamma}\ \le\ \frac{9}{4(1-\gamma)}.
\]
Both depths are at most $m+4/(1-\gamma)$, so for $L\ge m+4/(1-\gamma)$ every context of depth $L$ incurs error at least $\Psi_m/64$ on one of the two instances, exactly as in that proof.

\emph{(d)} Let $\gamma\ge\frac12$ and $8m\le L\le1/(1-\gamma)$. Take Instance~II with $w=\lfloor L/(4m)\rfloor\ge2$ and $T=2m$; its even blocks $u_t=\tau_{2t}$, $t<T$, have $b=(2t+1)w\le(4m-1)w<L$, are orthogonal, and satisfy $\|u_t\|_\mu=\gamma^{2tw}\|\tau_0\|_\mu\ge a:=\gamma^{4mw}S_w/\sqrt2$. For a context with error $\varepsilon$ and $U=\operatorname{col}(D)$, the argument of part (c) gives $\sum_{t<T}\operatorname{dist}(u_t,U)^2\ge a^2(T-m)=a^2m$, so $\max_t\operatorname{dist}(u_t,U)\ge a/\sqrt2$ and, by Lemma~\ref{lem:block-width}, $\varepsilon\ge a/(2\sqrt2)\ge\gamma^{4mw}S_w/4$. Now $S_w\ge w\gamma^{w}$, $(4m+1)w\le\frac{4m+1}{4m}L\le\frac{5}{4(1-\gamma)}$, and $-\ln\gamma\le2(1-\gamma)$ for $\gamma\ge\frac12$, so $\gamma^{(4m+1)w}\ge e^{-5/2}$; with $w\ge L/(4m)-1\ge L/(8m)$ this gives $\varepsilon\ge e^{-5/2}L/(32m)\ge L/(400m)$. The upper bound is part (b) with $\Psi_{m,L}\le\gamma^m\max\{1,L/m\}\le L/m$, because $(1-\gamma^L)/(1-\gamma)\le L$ and $L\ge m$.
\end{proof}

\subsection{Proof of Proposition~\ref{prop:mm-suboptimal}}
\label{app:mm-proof}
\paragraph{Terminal correction (Krylov truncation).}
For $m<k_L$ the budget-$m$ encoder gives an upper-Hessenberg $H_m$ with unit subdiagonal and $g_m=e_1$. Replacing $H_m$ by $T_m=H_m-(H_m^mg_m)e_m^\top$ preserves $T_m^jg_m=H_m^jg_m$ for $j<m$ and gives $T_m^mg_m=0$, since $e_m^\top H_m^jg_m=0$ for $j<m-1$ and $e_m^\top H_m^{m-1}g_m=1$. Seven further heads compute $T_m$ by $m$ reduced multiplications (Appendix~\ref{app:stage2}), and the output is exactly $\widehat v_{\min\{\ell,m\}}$, i.e., TD stopped after $m$ steps.

\begin{proof}[Proof of (a)]
By Lemma~\ref{lem:contraction}, $\|\widehat v_\ell-\widehat v_m\|_\mu\le\sum_{i\ge m}\gamma^i\|\widehat r\|_\mu=\gamma^m\|\widehat r\|_\mu/(1-\gamma)$. For the matching lower bound, take the ring with $n=m+2$, $L\ge m+1$, and $r_\varepsilon=\mathbf 1+\varepsilon\sqrt n\,e_0$. The Krylov vectors $\gamma^j(\mathbf 1+\varepsilon\sqrt n\,e_{-j})$ form, up to a column permutation, the matrix $\mathbf 1\mathbf 1^\top+\varepsilon\sqrt nI$, which is nonsingular; hence $k_L>m$. As $\ell\to\infty$ the error tends to $\|\widehat A^mv_\infty\|_\mu$ with $v_\infty=\frac{\mathbf 1}{1-\gamma}+\varepsilon\sqrt n(I-\widehat A)^{-1}e_0$, which is at least $\gamma^m(1-\varepsilon)/(1-\gamma)$, while $\|r_\varepsilon\|_\mu\le1+\varepsilon$. Letting $\varepsilon\to0$ gives the supremum $\gamma^m/(1-\gamma)$, and $\frac{\gamma^m/(1-\gamma)}{\Psi_m}=\min\{m,\frac1{1-\gamma}\}$.
\end{proof}

\begin{proof}[Proof of (b)]
Take the ring with $n=m+p$ states, $p=\lceil1/(1-\gamma)\rceil$, $r=\sqrt n\,e_0$, and $L\ge n$. Then $\varphi_t=P^tr$ ($t<n$) are orthonormal in $\|\cdot\|_\mu$, and all visit counts are equal, so Euclidean and visit-weighted Arnoldi coincide. The encoder gives $b_j\propto\varphi_{j-1}$ for $j\le m$, and $\widehat Ab_m\propto\varphi_m\perp\operatorname{span}(b_1,\ldots,b_m)$, so the last column of $H$ vanishes and $H$ is nilpotent. Galerkin execution therefore outputs $\sum_{t<\min\{\ell,m\}}\gamma^t\varphi_t$, and at $\ell=n$ its error is $\|\sum_{t=m}^{n-1}\gamma^t\varphi_t\|_\mu=\gamma^m\sqrt{(1-\gamma^{2p})/(1-\gamma^2)}$. Since $\gamma^{2p}\le e^{-2}$ and $1-\gamma^2\le2(1-\gamma)$, this is at least $\sqrt{(1-e^{-2})/2}\,\gamma^m/\sqrt{1-\gamma}\ge0.657\,\gamma^m/\sqrt{1-\gamma}$. For $m\ge1/(1-\gamma)$ we have $\Psi_m=\gamma^m$, which gives the second form. The instance has $k_L=\min\{n,L\}>m$.
\end{proof}

\subsection{Proof of Theorem~\ref{thm:anchored}}
\begin{lemma}[Normalized Galerkin form]
\label{lem:normalized}
Suppose the budget-$k$ $w$-Arnoldi encoder does not break down. Let $Q$ be its $w$-orthonormalized basis, $Q^*=Q^\top D_w$, and $\widehat H=Q^*\widehat AQ$. Then the executed trajectory is $\widetilde v^{\rm G}_\ell=\sum_{i<\ell}Q\widehat H^iQ^*\widehat r$, the matrix satisfies $\|\widehat H\|_2\le\|\widehat A\|_w\le\gamma$, and $Q\widehat H^iQ^*\widehat r=\widehat A^i\widehat r$ for $i<k$. Diagonal rescalings of the basis, including $\ell_1$ balancing, leave the trajectory unchanged.
\end{lemma}
\begin{proof}
With $D_g=\operatorname{diag}(\|b_i\|_w)$ and $Q=BD_g^{-1}$, we have $\Omega=D_wBD_g^{-2}$, $H=D_g^{-1}\widehat HD_g$, and $g=D_g^{-1}Q^*\widehat r$, so $BH^ig=Q\widehat H^iQ^*\widehat r$. $Q$ is an isometry from $(\mathbb R^k,\ell_2)$ into $(\mathbb R^n,\|\cdot\|_w)$, which gives the norm bound. Moment matching for $i<k$ follows as in Theorem~\ref{thm:main}, with the $w$-orthogonal projection $QQ^*$.
\end{proof}
\begin{proof}[Proof of Theorem~\ref{thm:anchored}]
If breakdown occurs within $k$ rounds, $\mathcal K_L$ is invariant, all residuals vanish, and the output is exact. Otherwise let $e_\ell=\widehat v_\ell-\widetilde v^{\rm G}_\ell$. By Lemma~\ref{lem:normalized}, $e_\ell=0$ for $\ell\le k$, and for $k\le a\le b$,
\[
 \|e_b-e_a\|_w\le\sum_{i=a}^{b-1}\bigl(\|\widehat A^i\widehat r\|_w+\|Q\widehat H^iQ^*\widehat r\|_w\bigr)\le2\sum_{i=a}^{b-1}\gamma^i\|\widehat r\|_w .
\]
The anchored output error is $\|e_\ell-e_{\sigma(\ell)}\|_w$, where $\sigma(\ell)\in\{k\}\cup\{k+i_s\}$ and $e_k=0$. The argument of Lemma~\ref{lem:sh} bounds it by $2\gamma^k\|\widehat r\|_w/((j+1)(1-\gamma))$, and trivially by $2\sup_\ell\|e_\ell\|_w=2\mathcal E^{\rm G}_k\|\widehat r\|_w$. The increments lie in $\operatorname{col}(B)$ plus the span of $j$ held residual increments, so the output is in $\mathfrak C_m$.

It remains to show $F:=2\gamma^k/((j+1)(1-\gamma))\le2e^2\Psi_m$. If $\gamma<\frac12$, then $j=0$ and $F\le4\gamma^m\le4\Psi_m$. Let $\gamma\ge\frac12$, $J=\lfloor1/(1-\gamma)\rfloor\ge2$, and $x=m(1-\gamma)$. If $m\le3$, then $j=0$ and $F/\Psi_m=2\min\{m,1/(1-\gamma)\}\le6$. If $j=J<\lfloor m/4\rfloor$, then $J+1>1/(1-\gamma)$ gives $F<2\gamma^{-J}\gamma^m\le2e^2\Psi_m$. Otherwise $m\ge4$ and $j=\lfloor m/4\rfloor\le J$. Then $j+1\ge(m+1)/4$ gives $F<8\gamma^{-j}\gamma^m/x$, hence $F/\Psi_m<8\gamma^{-j}\min\{1,1/x\}$. Here $\gamma^{-j}\le\min\{\gamma^{-m/4},\gamma^{-J}\}\le\min\{e^{x/2},e^2\}$. The function $e^{x/2}/x$ is convex on $[1,4]$, so its maximum there is at an endpoint. The bound is therefore at most $8e^{1/2}$ for $x\le1$, at most $\max\{8e^{1/2},2e^2\}=2e^2$ for $1<x\le4$, and at most $8e^2/x<2e^2$ for $x>4$. Finally, $k=m-j\ge m-\lfloor m/4\rfloor=\lceil3m/4\rceil$, and each residual token stores $n$ values and one time stamp.
\end{proof}

\subsection{Proof of Proposition~\ref{prop:instance}}
\begin{proof}[Proof of (a)]
If $N_{ij}=N_{ji}$, then $\sum_iN_{ij}=N_j$, so $w=(N_i)$ is stationary; moreover $w_i\widehat P_{ij}=N_{ij}$ is symmetric, so $\widehat A$ is $w$-self-adjoint with spectrum in $[-\gamma,\gamma]$. $\widehat H$ is then symmetric with spectrum in $[-\gamma,\gamma]$. If $m\ge k_L$ the Galerkin output is exact. Otherwise, for every polynomial $q$ of degree less than $m$, $Qq(\widehat H)Q^*\widehat r=q(\widehat A)\widehat r$. With $p_\ell(x)=\sum_{j<\ell}x^j$,
\[
 \widehat v_\ell-\widetilde v^{\rm G}_\ell=(p_\ell-q)(\widehat A)\widehat r-Q(p_\ell-q)(\widehat H)Q^*\widehat r,
 \qquad \|\widehat v_\ell-\widetilde v^{\rm G}_\ell\|_w\le2\sup_{|x|\le\gamma}|p_\ell(x)-q(x)|\,\|\widehat r\|_w .
\]
Write $x=\gamma t$ with $t\in[-1,1]$ and expand in Chebyshev polynomials $T_k(t)$. The Poisson-kernel identity gives $\frac1{1-\gamma\cos\theta}=\frac1{\sqrt{1-\gamma^2}}(1+2\sum_{k\ge1}\beta^k\cos k\theta)$, so $f(x)=1/(1-x)$ has coefficients $c_k(f)=2\beta^k/\sqrt{1-\gamma^2}$ for $k\ge1$. Every monomial $t^j$ has nonnegative Chebyshev coefficients. Hence $0\le c_k(p_\ell)\le c_k(f)$, since $c_k(p_\ell)$ is nondecreasing in $\ell$ and converges to $c_k(f)$ by uniform convergence on $[-1,1]$. Truncating $p_\ell$ after degree $m-1$ gives
$\sup|p_\ell-q_\ell|\le\sum_{k\ge m}c_k(f)=2\beta^m/((1-\beta)\sqrt{1-\gamma^2})$ uniformly in $\ell$, which proves the first bound \citep{trefethen2013approximation}. For the second, $s=\sqrt{1-\gamma^2}$ and $1-\beta=(1+s-\gamma)/(1+s)$, and $(1-\beta)s\ge s^2/2$ is equivalent to $(1-\gamma)^2+s\ge0$. The anchored bound follows from Theorem~\ref{thm:anchored} with $k\ge\lceil3m/4\rceil$.
\end{proof}
\begin{proof}[Proof of (b)]
Every context in $\mathfrak C_m$ has $\widetilde v_\ell\in U=\operatorname{col}(D)$, since the initial output is zero. For any $U$ of dimension at most $m$, Eckart--Young gives
$\max_{\ell\le L}\operatorname{dist}_w(\widehat v_\ell,U)^2\ge\frac1L\sum_{\ell\le L}\operatorname{dist}_w(\widehat v_\ell,U)^2=\frac1L\|(I-\Pi_U)V_L\|_F^2\ge\sigma_{m+1}^2/L$.
Projecting onto the top $m$ left singular vectors gives residual at most $\|(I-\Pi)V_L\|_2=\sigma_{m+1}$ in every column. For stability, Weyl's inequality gives $|\sigma_i(V_L)-\sigma_i(V_L')|\le\|V_L-V_L'\|_F\le\sqrt L\sup_\ell\|\widehat v_\ell-\widehat v'_\ell\|_w$. The recursion $e_{\ell+1}=(\widehat r-\widehat r')+\widehat Ae_\ell+(\widehat A-\widehat A')\widehat v'_\ell$ with $\|\widehat v'_\ell\|_w\le\|\widehat r'\|_w/(1-\gamma)$ then gives $\sup_\ell\|e_\ell\|_w\le\delta_r/(1-\gamma)+\delta_A\|\widehat r'\|_w/(1-\gamma)^2$ when both models are $w$-contractive.
\end{proof}

\subsection{A Certified Adaptive Rank}
\label{app:threshold}
\begin{proposition}[Threshold certificate]
Add the declared gate $\operatorname{thr}_c(x)=\mathbf 1\{x>c\}$. In round $j$ of $w$-Arnoldi, multiply $b_{j+1}$ by $\operatorname{thr}_{\tau^2}(d_{j+1}/d_j)$, computed as $d_{j+1}\operatorname{recip}_0(d_j)$. Let $j^\star$ be the number of retained directions. If the gate fired, i.e., $\rho:=\|b_{j^\star+1}\|_w/\|b_{j^\star}\|_w\le\tau$, then under $w$-contraction
\[
 \sup_{\ell\ge0}\|\widehat v_\ell-\widetilde v^{\rm G}_\ell\|_w\le\frac{\tau}{(1-\gamma)^2}\|\widehat r\|_w .
\]
Otherwise the output is exact through depth $L$. Always $j^\star\le k_L$, and $\tau=0$ recovers the exact encoder.
\end{proposition}
\begin{proof}
With $q_{j+1}=b_{j+1}/\|b_{j+1}\|_w$, the Arnoldi relation reads $\widehat AQ=Q\widehat H+\rho\,q_{j^\star+1}e_{j^\star}^\top$, where $b_{j^\star+1}$ is the residual before gating. Let $\epsilon_i=\widehat A^i\widehat r-Q\widehat H^iQ^*\widehat r$, so that $\epsilon_0=0$ and $\epsilon_{i+1}=\widehat A\epsilon_i+\rho\,q_{j^\star+1}e_{j^\star}^\top\widehat H^iQ^*\widehat r$. Hence $\|\epsilon_{i+1}\|_w\le\gamma\|\epsilon_i\|_w+\rho\gamma^i\|\widehat r\|_w$, and induction gives $\|\epsilon_i\|_w\le i\rho\gamma^{i-1}\|\widehat r\|_w$. Summing over $i\ge1$ gives $\rho/(1-\gamma)^2$.
\end{proof}
The certificate controls the error, not the size of $j^\star$, because Arnoldi residuals can stagnate. Proposition~\ref{prop:instance}(b) gives the method-independent lower bound against which $j^\star$ can be compared.

\subsection{Numerically Optimized Time-Invariant Recurrences}
\label{app:ti-probe}
For $\widetilde v_\ell=D\sum_{j<\ell}K^js_0$ with context dimension $m$, the increment sequences $j\mapsto K^js_0$ span a space of exponential sums whose modes are the eigenvalues of $K$. We parameterize $m/2$ damped complex modes $\rho_ie^{\pm\mathrm i\theta_i}$. For fixed modes, the minimax problem over $D$ is convex, and we solve it with Lawson iterations. The modes are optimized by Nelder--Mead from 12 starts. This is a numerical search, so it certifies achievability of the reported values but not global optimality. On Instance~I with $\gamma=0.9$:

\begin{center}\small
\begin{tabular}{@{}lrrrr@{}}
\toprule
$m$ & 2 & 4 & 6 & 8\\
\midrule
Best time-invariant found & 0.535 & 0.341 & 0.251 & 0.194\\
Sample-and-hold & 1.570 & 1.345 & 1.000 & 1.000\\
Certified lower bound (any method) & 0.307 & 0.231 & 0.180 & 0.143\\
\bottomrule
\end{tabular}
\end{center}

Time-invariant recurrences with instance-optimized modes are therefore within $1.8\times$ of optimal on this instance, and better than sample-and-hold, whose first sample occurs after depth one. The suboptimality in Proposition~\ref{prop:mm-suboptimal} is thus a property of moment matching, not of looping without a clock. What remains open is whether some encoder computable in the forward pass produces time-invariant contexts with error $O(\Psi_m)$ on \emph{every} system. The depth-counter constructions give a positive answer with a simple, fixed encoder.
\section{Execution Depth}
\label{app:depth}
\label{app:depth-proof}
The time-invariant executor (\ref{eq:executor}) uses $L$ layers and emits every depth along the way. For a single-depth query we ask how few layers suffice, and how few are necessary. A \emph{polynomial executor} of depth $t$ composes $t$ maps $F(Z)=W_0Z+c_0+\sum_hP_hZ[\mathcal R_h\odot(Z^\top Q_hZ)]$ with fixed parameters that may depend on $n,L,m$ but not on $\gamma$ or the task, followed by a fixed linear readout. The class covers every executor of Appendix~\ref{app:executors} without $\operatorname{recip}_0$.
\begin{theorem}[Depth bounds]
\label{thm:depth}
\textbf{(a)} Suppose every context entry is a Laurent polynomial in $\gamma$ with exponents in $[-e,e]$, and a depth-$t$ polynomial executor outputs $\widehat v_L(q)$ for all $\gamma$ in an open interval on a task with $e_q^\top\widehat P^{L-1}\widehat r\ne0$. Then $3^te\ge L-1$. The unnormalized Krylov context (\ref{eq:krylov-context}) has $e=k_L$, whereas the $\ell_1$-balanced Krylov context used in Section~\ref{sec:experiments} and the state-level context (\ref{eq:linear-full-context}) have $e=1$ (Lemma~\ref{lem:balanced-affine}); hence $t\ge\log_3\frac{L-1}{k_L}$ in general and $t\ge\log_3(L-1)$ for the balanced and state-level contexts.
\textbf{(b)} Retaining the reduced columns of $H$, a fixed executor with two layer types (repeated squaring of $H$ and binary accumulation of $\sum_{j<L}H^jg$) outputs $\widehat v_L(q)$ exactly in $\lfloor\log_2L\rfloor+2$ layers. Its retained memory is $(n+h+2)k_L+O(n)\le2(nk_L+k_L)+O(n)$.
\end{theorem}
For single-depth queries on the balanced context, the optimal depth therefore lies between $\log_3(L-1)$ and $\log_2L+2$, which agree up to the factor $\log_23\approx1.58$ and an additive constant (a further $\log_3k_L$ for the unnormalized context); only the base of the logarithm is open. The lower bound is for polynomial executors, which excludes $\operatorname{recip}_0$, so it constrains the doubling and time-invariant executors but not the depth-counter one, whose unrolled $\operatorname{recip}_0$-free variant has depth $L$. Compression reduces memory (Theorem~\ref{thm:main}) but not depth, since the uncompressed state-level context can also be squared in $O(\log L)$ layers at $n^2$ memory. The looped executor's linear depth buys all $L$ depths in one pass.

\subsection{Structure of the Context in $\gamma$}
\begin{lemma}[Laurent structure in $\gamma$]
\label{lem:laurent}
Fix real $P,r$ with $k=\dim\mathcal K_L(P,r)\ge1$, and write $\tilde b_j,\tilde\omega_j,\tilde H$ for the unbalanced encoder outputs on $(P,r)$. Then on $(\gamma P,r)$ with $\gamma\ne0$, the outputs are $b_j=\gamma^{j-1}\tilde b_j$, $\omega_j=\gamma^{-(j-1)}\tilde\omega_j$, $H_{ij}=\gamma^{j-i+1}\tilde H_{ij}$, and $g=e_1$. Every entry of $B$, $Y$, $H$, and $g$ is a Laurent polynomial in $\gamma$ with exponents in $[-k,k]$.
\end{lemma}
\begin{proof}
Induct on rounds. If $b_i=\gamma^{i-1}\tilde b_i$ for $i\le j$, then $d_i=\gamma^{2(i-1)}\tilde d_i$, $\omega_i=\gamma^{-(i-1)}\tilde\omega_i$, and $\gamma Pb_j=\gamma^j P\tilde b_j$. Substituting into (\ref{eq:arnoldi}) gives $b_{j+1}=\gamma^j\tilde b_{j+1}$, and the gates do not depend on $\gamma$. Then $H_{ij}=\omega_i^\top\gamma Pb_j=\gamma^{j-i+1}\tilde H_{ij}$, and $g=\Omega^\top b_1=e_1$ by duality. $H$ is upper Hessenberg, since $\widehat Ab_j\in\operatorname{span}(b_1,\ldots,b_{j+1})$, so its nonzero exponents lie in $[0,k]$. The three terms of $y_j=b_j+\sum_i\omega_i(H_{ji}-\delta_{ji})$ have exponents $j-1$, $2-j$ (from $\gamma^{-(i-1)}\gamma^{i-j+1}$), and $1-j$. All lie in $[-k,k]$.
\end{proof}

\begin{lemma}[The balanced context is affine in $\gamma$]
\label{lem:balanced-affine}
With positive $\ell_1$ balancing, write $\tilde b_j,\tilde\omega_j,\tilde H,\tilde g$ for the balanced encoder outputs on $(P,r)$. On $(\gamma P,r)$ with $\gamma>0$ the outputs are $b_j=\tilde b_j$, $\omega_j=\tilde\omega_j$, $g=\tilde g$, $H=\gamma\tilde H$, and $Y=\tilde B-\tilde\Omega+\gamma\,\tilde\Omega\tilde H^\top$, and the gates do not depend on $\gamma$. Every entry of $B$, $Y$, $H$, and $g$ is therefore a polynomial of degree at most one in $\gamma$, i.e., $e=1$ in Theorem~\ref{thm:depth}(a).
\end{lemma}
\begin{proof}
Induct on rounds. The first candidate is $\widehat r$, which does not involve $\gamma$. If $b_i=\tilde b_i$ and $\omega_i=\tilde\omega_i$ for $i\le j$, the next candidate before balancing is $\gamma Pb_j-\sum_{i\le j}b_i\omega_i^\top\gamma Pb_j=\gamma\bigl(P\tilde b_j-\sum_{i\le j}\tilde b_i\tilde\omega_i^\top P\tilde b_j\bigr)$, a positive multiple of the candidate on $(P,r)$. Balancing divides by the $\ell_1$ norm, which removes the factor $\gamma$, so $b_{j+1}=\tilde b_{j+1}$; the squared norm, the dual, and the gate of round $j+1$ are functions of $b_{j+1}$ alone. Hence $H_{ij}=\omega_i^\top\gamma Pb_j=\gamma\tilde H_{ij}$, $g=\Omega^\top\widehat r=\tilde g$, and $y_j=b_j+\sum_i\omega_i(H_{ji}-\delta_{ji})$ is affine in $\gamma$. Numerically, on 30 random covered batches the balanced $B$ and $\Omega$ at $\gamma\in\{0.5,0.8,0.95\}$ agree to $7.1\times10^{-15}$ and $g$ to $1.8\times10^{-15}$, $H/\gamma$ agrees to $1.3\times10^{-15}$, and $Y$ is affine to $2.8\times10^{-14}$.
\end{proof}

\subsection{Doubling Executor}
\label{app:doubling}
Each retained Krylov token $j$ carries $b_j$, a block $m_j\in\mathbb R^h$ initialized to column $j$ of $H$, a fixed tag $p_j=e_j$, and scalars $a_j$ (initialized to $g_j$) and $c_j$ (initialized to $0$). Let $M$ denote the matrix with columns $m_j$. Six token-to-token heads and one readout head are used. The squaring head (key tag, query $m$, value $m$, output $m$, $\sigma=+$) adds $\sum_im_i(p_i^\top m_t)=M^2e_t$, and a clearing head (key tag, query tag, value $m$, output $m$, $\sigma=-$) removes the old column, so together they map $M\mapsto M^2$. The advance head (key $m$, query tag, value $a$, output $a$) adds $(Ma)_t$, so $a\mapsto a+Ma$. The accumulation heads add $(Mc)_t$ to $c$ (key $m$, query tag, value $c$), add $a_t$ to $c$ (key tag, query tag, value $a$, output $c$), and subtract $c_t$ (key tag, query tag, value $c$, output $c$, $\sigma=-$), so together $c\mapsto a+Mc$. With synchronous reads, a layer of type $\Lambda_0$ (squaring, clearing, advance) maps $(M,a,c)$ to $(M^2,a+Ma,c)$, and a layer of type $\Lambda_1$ (all six heads) maps it to $(M^2,a+Ma,a+Mc)$. Layer $s=0,\ldots,\lfloor\log_2L\rfloor$ uses $\Lambda_{\beta_s}$, where $L=\sum_s\beta_s2^s$, and a final readout head (key $b$, query $e_q$, value $c$) outputs $e_q^\top Bc$.

\subsection{Proofs}
\begin{lemma}[Degree growth]
\label{lem:degree}
Each output entry of $F(Z)=W_0Z+c_0+\sum_hP_hZ[\mathcal R_h\odot(Z^\top Q_hZ)]$ is a polynomial of degree at most $3$ in the entries of $Z$. A composition of $t$ such maps followed by a linear readout therefore has degree at most $3^t$.
\end{lemma}
\begin{proof}
$P_hZ$ is linear, $Z^\top Q_hZ$ is quadratic, and the masked product is cubic. Degrees multiply under composition.
\end{proof}

\begin{proof}[Proof of Theorem~\ref{thm:depth}(a)]
By Lemma~\ref{lem:degree}, a polynomial of degree at most $3^t$ in Laurent polynomials with exponents in $[-e,e]$ is a Laurent polynomial $O(\gamma)$ with exponents in $[-3^te,3^te]$. The target $V(\gamma)=\sum_{j<L}\gamma^je_q^\top\widehat P^j\widehat r$ is a polynomial whose coefficient of $\gamma^{L-1}$ is nonzero. The polynomial $\gamma^{3^te}(O-V)$ vanishes on an open interval, so it is identically zero. Hence $O=V$ as Laurent polynomials, and $L-1\le3^te$. For the unnormalized Krylov context, Lemma~\ref{lem:laurent} gives $e=k_L$; for the balanced context used in the experiments, Lemma~\ref{lem:balanced-affine} gives $e=1$; the query token is constant. The state-level context (\ref{eq:linear-full-context}) is affine in $\gamma$, so $e=1$. Tasks satisfying the hypothesis exist, for example the base point of Lemma~\ref{lem:basepoint} with all $\cos(\pi j/n)\neq0$.
\end{proof}

\begin{proof}[Proof of Theorem~\ref{thm:depth}(b)]
Use the heads of Appendix~\ref{app:doubling}. Since $p_i^\top p_t=\delta_{it}$ and $M$ is supported on the first $k_L$ coordinates, the squaring head gives $\sum_im_i(p_i^\top m_t)=M^2e_t$, and the advance head gives $\sum_ia_i(m_i^\top p_t)=(Ma)_t$. Let $p_x(H)=\sum_{j<x}H^j$ and $x_s=\sum_{s'<s}\beta_{s'}2^{s'}$. We claim that at the start of step $s$, $M=H^{2^s}$, $a=p_{2^s}(H)g$, and $c=p_{x_s}(H)g$. This holds at $s=0$, and it is preserved because $p_{x+y}=p_y+H^yp_x$ gives $a+Ma=p_{2^{s+1}}g$ and $a+Mc=p_{x_s+2^s}g$. After step $\lfloor\log_2L\rfloor$ we have $c=p_L(H)g$, and the readout gives $e_q^\top Bp_L(H)g=e_q^\top\sum_{j<L}\widehat A^j\widehat r$ by moment matching. The retained data are $B$ ($nk_L$ entries), the columns of $H$ ($hk_L$), $a$ and $c$ ($2k_L$), and the query ($O(n)$). Since $h\le n$, the total is at most $2(nk_L+k_L)+O(n)$. Numerically, the recurrence of Appendix~\ref{app:doubling} evaluated as matrix products has relative error at most $7.0\times10^{-15}$ for $L\le200$.
\end{proof}

\section{Learnability of the Shared Executor}
\label{app:learnability}
Algorithm~\ref{alg:td-pretraining} in Section~\ref{sec:learnability} gives the pretraining procedure. This appendix describes its calibration interface, proves the convergence and identifiability results, and relates parameter error to trajectory error.

\paragraph{Calibration interface.} The encoder remains fixed. Its first stage produces a state-level context with static blocks $I_n$ and $\widehat A_\tau^\top$. For calibration, the executor initializes its state vector to an input $u$ satisfying $\mathbb E[uu^\top]=I_n$ and resets the accumulated query output to zero. Multiplication with $Q_{\rm eff}(C,W)$ then produces the output increment $-C^\top u$ and the next state $F_\tau(C,W)u=(I_n+C^\top+\widehat A_\tau W)u$. The two terms of (\ref{eq:probe}) compare these quantities with $u$ and $\widehat A_\tau u$, respectively. Both targets are obtained from the input statistics and $u$; neither a solved value function nor a target parameter is needed to evaluate the loss.

\begin{proof}[Proof of Theorem~\ref{thm:learn}]
Let $E_C=C+I$ and $E_W=W-I$. Isotropy of $u$ gives $2\mathcal J=\|E_C\|_F^2+\mathbb E_\tau\|E_C^\top+\widehat A_\tau E_W\|_F^2$. For equally shaped arrays $a,b$, the quadratic form $\|a\|_F^2+\|a+b\|_F^2$ has coefficient matrix $\left[\begin{smallmatrix}2&1\\1&1\end{smallmatrix}\right]$, whose eigenvalues are $c_\pm=(3\pm\sqrt5)/2$. Therefore
$c_-(\|E_C\|_F^2+\mathbb E\|\widehat A_\tau E_W\|_F^2)\le2\mathcal J\le c_+(\|E_C\|_F^2+\mathbb E\|\widehat A_\tau E_W\|_F^2)$.
Since $\mathbb E\|\widehat A_\tau E_W\|_F^2=\operatorname{tr}(E_W^\top GE_W)$ with $\kappa I\preceq G\preceq\beta'I$, the Hessian satisfies $\lambda_-I\preceq\mathcal H\preceq\lambda_+I$. The unique zero is $E_C=E_W=0$, and a GD step is $e_{t+1}=(I-\eta\mathcal H)e_t$, whose eigenvalues lie in $[0,1-\eta\lambda_-]$ for $\eta\le1/\lambda_+$.
\end{proof}

\paragraph{Necessity of operator diversity.} If $G$ is singular, take $z$ in its nullspace; then $\widehat A_\tau z=0$ almost surely. The pair $C=-I$, $W=I+za^\top$ has zero loss for every $a$, so the probe objective cannot identify unexcited directions. Conversely, if discounted permutations appear with probability $p>0$, then $G\succeq p\gamma_0^2I$ even when all rewards coincide.

\paragraph{Parameter error and trajectory error.} For a fixed exact context with $K_\theta=I+B^\top C^\top B+Y^\top WB$ and $v^\theta_\ell=-C^\top B\sum_{j<\ell}K_\theta^jg$, let $\delta_C=\|C+I\|_2$, $\delta_W=\|W-I\|_2$, $\Delta=\|B\|_2^2\delta_C+\|Y\|_2\|B\|_2\delta_W$, and $b=\max\{\|K_\theta\|_2,\|H\|_2\}$. Then $\|K_\theta-H\|_2\le\Delta$ and
$\|v^\theta_\ell-\widehat v_\ell\|_2\le\|B\|_2\|g\|_2[\delta_C\sum_{j<\ell}b^j+\Delta\sum_{1\le j<\ell}jb^{j-1}]$,
by telescoping $K_\theta^j-H^j=\sum_iK_\theta^{j-1-i}(K_\theta-H)H^i$. Hence gradient descent also converges in output on every fixed finite-depth context.

\paragraph{Why an endpoint target differs.} Scaling to $C=-aI$, $W=aI$ ($0<a\le1$) gives $K_a=(1-a)I+aH$ and increments $aBK_a^jg$. When $H$ is stable, $aB(I-K_a)^{-1}g=B(I-H)^{-1}g$ for every $a$. Different unit-step rules can therefore share a limit, which is why terminal prediction and identification of the TD step are different objectives.

\begin{proof}[Proof of Proposition~\ref{prop:identify}]
Write $C=-I+E_C$ and $W=I+E_W$, so that $-C^\top=I-E_C^\top$ and $F_\tau=\widehat A_\tau+\delta F_\tau$ with $\delta F_\tau=E_C^\top+\widehat A_\tau E_W$. To first order in $(E_C,E_W)$,
\[
 \delta v^\theta_{\tau,1}=-E_C^\top\widehat r_\tau,\qquad
 \delta v^\theta_{\tau,2}=-E_C^\top(\widehat r_\tau+\widehat A_\tau\widehat r_\tau)+\delta F_\tau\widehat r_\tau=-E_C^\top\widehat A_\tau\widehat r_\tau+\widehat A_\tau E_W\widehat r_\tau .
\]
If the Jacobian applied to $(E_C,E_W)$ vanishes in $L^2(\tau)$, then $\mathbb E_\tau\|E_C^\top\widehat r_\tau\|^2=\operatorname{tr}(E_C^\top\mathbb E[\widehat r_\tau\widehat r_\tau^\top]E_C)=0$ forces $E_C=0$, and then $\mathbb E_\tau\|\widehat A_\tau E_W\widehat r_\tau\|^2=\operatorname{vec}(E_W)^\top\mathbb E[(\widehat r_\tau\widehat r_\tau^\top)\otimes(\widehat A_\tau^\top\widehat A_\tau)]\operatorname{vec}(E_W)=0$ forces $E_W=0$. Hence the Jacobian is injective, the Gauss--Newton Hessian of $\mathcal J_{\rm traj}$ at the target is positive definite, and the target, at which $\mathcal J_{\rm traj}=0$, is a strict local minimizer. For the endpoint map, on the curve $(C,W)=(-aI,aI)$ we have $-C^\top=aI$ and $F_\tau=(1-a)I+a\widehat A_\tau$, so $v^\theta_{\tau,\infty}=aI\,(I-F_\tau)^{-1}\widehat r_\tau=(I-\widehat A_\tau)^{-1}\widehat r_\tau$ for every $a$; differentiating at $a=1$ shows that the Jacobian of $(C,W)\mapsto v^\theta_{\tau,\infty}$ annihilates $(-I,I)$.
\end{proof}
The calibration objective (\ref{eq:probe}) is the same two-depth identification with an isotropic input in place of $\widehat r_\tau$: it removes the reward moments from the conditions and makes the objective exactly quadratic, which is what Theorem~\ref{thm:learn} exploits.

\section{Connections to Classical Statistical and Reduced Models}
\label{app:classical}
\paragraph{LSTD and iLSTD.} With features $x_t$, batch TD depends on the data only through $G_N=\sum_tx_t(x_t-\gamma x_{t+1})^\top$ and $b_N=\sum_tx_tR_{t+1}$. LSTD solves the projected fixed point, and recursive variants update these statistics online \citep{bradtke1996lstd,boyan2002lstd,geramifard2006ilstd}. Our first stage is the tabular, visitation-normalized instance and is not a new principle. Theorem~\ref{thm:memory-law}(4) shows that these statistics are optimal up to a factor $(n+1)/n$ among continuous encoders when $k_L=n$.
\paragraph{BEBFs, Krylov spaces, and model reduction.} Reward-initialized Krylov, Bellman-error, and model-error features span the same spaces under the conditions of \citet{parr2008models} (see also \citealp{parr2007bebf,petrik2007analysis}). The sequence $\widehat A^j\widehat r$ is the impulse response of a linear system with full-state output. Theorem~\ref{thm:main}(a) is therefore a finite-horizon minimal-realization statement \citep{ho1966realization,freund2003krylov,antoulas2005approximation}. Classical Krylov model reduction matches moments and controls $H_2$ or $H_\infty$ errors; our error is the supremum over depths of partial sums. Under this error, moment matching is not minimax (Proposition~\ref{prop:mm-suboptimal}).
\paragraph{Lumpability.} If $PF=F\overline P$ and $r=F\overline r$ for a partition matrix $F$, then $\sum_{j<\ell}(\gamma P)^jr=F\sum_{j<\ell}(\gamma\overline P)^j\overline r$ \citep{kemeny1976chains,givan2003equivalence}. A quotient restricts decoding directions to block indicators. Krylov directions are reward specific and exist without a compatible partition; the block-cycle experiments are lumpable, and the sampled experiments are not.
\paragraph{Widths.} Lemma~\ref{lem:waterfill} is a Kolmogorov-width computation for orthogonal families \citep{pinkus1985widths}. The new ingredient is the choice of block increments at the scale $1/(m(1-\gamma))$, which is what removes the $\gamma$-dependence of the constant.

\paragraph{Constructions for other in-context RL problems.}
\label{app:other-constructions}
Table~\ref{tab:context-comparison} lists only constructions that execute TD policy evaluation in the forward pass, because our retained object, the Krylov context of the TD trajectory, is a compression of that computation. Three further constructions retain other objects and are not comparable in retained memory: \citet{lin2024transformersdecisionmakers} implement LinUCB and Thompson sampling for bandits and UCB-VI for tabular MDPs from $T$ rounds of interaction with $|\mathcal A|$ actions, retaining $\Theta(Td|\mathcal A|)$; \citet{liang2026transformerspolicyimprovement} implement semi-gradient SARSA and actor--critic on $N$ transitions with $d$-dimensional state--action features, retaining $\Theta(Nd)$; and \citet{chen2025optimaldynamicregret} implement the MASTER strategy for non-stationary bandits over $T$ rounds, retaining $\Theta(Td\log T)$. Measured with the protocol of Table~\ref{tab:context-comparison} ($T=2000$, $d=24$, $|\mathcal A|=4$, padding to 2048 tokens for the last), their encoded buffers are at least $2000$ KiB, $578.41$ KiB, and $2824$ KiB, respectively; the first excludes unspecified auxiliary padding. Whether the control computations they execute admit a Krylov-type compression is open, since the relevant subspace changes with the policy.
\section{Experimental Details and Additional Results}
\label{app:experiments}
All experiments start from raw records $(S_t,R_{t+1},S_{t+1})$. The implementation stores tokens by role, with one array per named field, and evaluates every task-dependent operation as a routed selector head (\ref{eq:encoder-head}) with synchronous reads, plus $\operatorname{recip}_0$, release, the ReLU absolute value used by $\ell_1$ balancing, and the clock gate. It does not call QR, SVD, a linear solver, or a rank routine to build any compiled context. Such routines appear only in tests, in the method-free references of Figure~\ref{fig:sampled}, and in the learned time-invariant read-out of Appendix~\ref{app:exp-learned}, which rescales its recurrence by a numerically computed spectral norm. Unless stated otherwise, computation uses float64 arithmetic on a CPU. The following subsections specify the task-generation rules, reference calculations, optimization settings, and evaluation metrics. The float32 comparisons are described in Appendix~\ref{app:exp-memory}, and the learned bottleneck models use the precision specified in Appendix~\ref{app:exp-learned}.

\subsection{Identity and Bound Checks}
\label{app:exp-checks}
\begin{table}[h]
\caption{\textbf{Implementation identities and bound checks.} ``Max'' is the maximum over the stated random instances. Ratios below one mean that the bound holds with room to spare.}
\label{tab:checks}
\centering\small
\begin{tabularx}{\linewidth}{@{}X r@{}}
\toprule
Check (instances) & Max\\
\midrule
Stage-1 heads vs.\ direct counting of $(\widehat A,\widehat r)$ (60 random batches) & $2.2\times10^{-16}$\\
Exact Krylov context + TD executor vs.\ direct TD, balanced two-pass (60) & $3.8\times10^{-15}$ rel.\\
\quad same, literal unbalanced one-pass encoder (60) & $8.0\times10^{-13}$ rel.\\
Visit-weighted Galerkin execution vs.\ closed form; dual biorthogonality (60) & $2.4\times10^{-15}$; $4.4\times10^{-16}$\\
Sample-and-hold encoder + depth-counter executor vs.\ $\widehat v_{\sigma(\ell)}$ (60) & $8.9\times10^{-16}$\\
Krylov-anchored encoder + executor vs.\ $\widetilde v^{\rm G}_\ell+e_{\sigma(\ell)}$ (40) & $2.8\times10^{-15}$\\
Doubling executor (matrix recurrence of Appendix~\ref{app:doubling}) vs.\ $\widehat v_L(q)$, $L\le200$ (50) & $7.0\times10^{-15}$ rel.\\
Theorem~\ref{thm:memory-law} family: Jacobian rank of $\theta\mapsto$ trajectory for $(n,k,L)=(8,3,6),(10,5,6),(12,2,9),(7,4,4)$ & $27,55,26,28$ (full)\\
Sample-and-hold error $/\,\gamma^k((j+1)(1-\gamma))^{-1}$; $/\,e^2\Psi_m$ in $\|\cdot\|_\mu$ and $\|\cdot\|_\infty$ (300 + 120 systems) & $1.00$; $0.34$, $0.30$\\
Anchored error $/\min\{2e^2\Psi_m,2\gamma^k((j+1)(1-\gamma))^{-1}\}$; $/(2\mathcal E^{\rm G}_k)$ (120 count-symmetric) & $0.17$; $0.50$\\
Lanczos error $/\,4\beta^m((1-\beta)\sqrt{1-\gamma^2})^{-1}$, $\gamma\le0.999$ (100 count-symmetric) & $0.082$\\
Threshold certificate error $/\,\tau(1-\gamma)^{-2}$ (120) & $0.30$\\
Truncation on the constant-plus-spike ring vs.\ $\gamma^m/(1-\gamma)$ at $(\gamma,m)=(0.9,10),(0.99,10)$ & $3.4868$, $90.438$ (exact)\\
\bottomrule
\end{tabularx}
\end{table}
All entries are float64 comparisons. The first six rows run the attention-level implementation (the Stage-1 heads, the Krylov encoder, the executors, and the sample-and-hold and anchored encoders) on random covered batches against direct numerical references (counting, the TD recurrence, and closed forms). The remaining rows check the stated inequalities numerically: the doubling executor, the terminal-corrected truncation values, and the threshold certificate are evaluated by direct matrix recurrences of the corresponding formulas rather than through selector heads, and the Jacobian row is a finite-difference rank computation on the family of Appendix~\ref{app:memory-proof}, with the perturbed model kept inside the open set of positive kernels. Rank decisions use the finite-precision tolerance of Appendix~\ref{app:numerical-margin}, so ``exact'' throughout this appendix means agreement to float64 rounding, not exact arithmetic.

\subsection{Context--Accuracy Tradeoff (Figure~\ref{fig:context-budget})}
\label{app:exp-threshold}
Panels (a) and (b) use 100 tasks each with $n=24$, $\gamma=0.9$, $L=30$, and $N=960$ covered records, and report the mean and interquartile range of the normalized all-depth error $\max_{\ell\le L}\|\widehat v_\ell-\widetilde v_\ell\|_w/\|\widehat r\|_w$ with visit weights $w$, for the Krylov projection, Krylov truncation, sample-and-hold, and anchored contexts built from the raw records by the attention-level implementation. \emph{Cyclic MRPs:} 12 blocks of two states; every record from block $b$ moves to a uniformly random state of block $b+1$ modulo $12$; the reward is a random multiple in $\{0.5,\ldots,1.5\}$ of the indicator of block $0$; state labels are permuted (seed 2027). Row sums into the next block are exactly one and rewards are block-constant, so $k_L=12$ algebraically, and the balanced encoder returns $k_L=12$ on all 100 tasks. \emph{Heterogeneous reversible MRPs:} symmetric random weights with density $0.4$ plus $0.05$ on the diagonal, normalized by rows, with deterministic per-state rewards scaled to $\|r\|_\infty=1$, observed along one trajectory of $960$ steps (seed 2028; one extra visit is appended to any unvisited state). All 100 tasks have $k_L=24$. Panel (c) is the $\gamma=0.99$ panel of Figure~\ref{fig:rates}, whose instances and certificate are described in the next subsection. The mean Krylov projection error on the cyclic tasks is $0.86$ at $m=11$ and $6\times10^{-16}$ for $m\ge12$; on the reversible tasks it is $1.1\times10^{-1}$, $2.0\times10^{-4}$, $2.2\times10^{-7}$, and $1.9\times10^{-14}$ at $m=4,8,12,20$, while truncation gives $1.2$, $0.73$, $0.45$, and $0.15$ and sample-and-hold $1.1$, $1.0$, $0.12$, and $0.053$.

\begin{figure}[h]
\centering
\includegraphics[width=\linewidth]{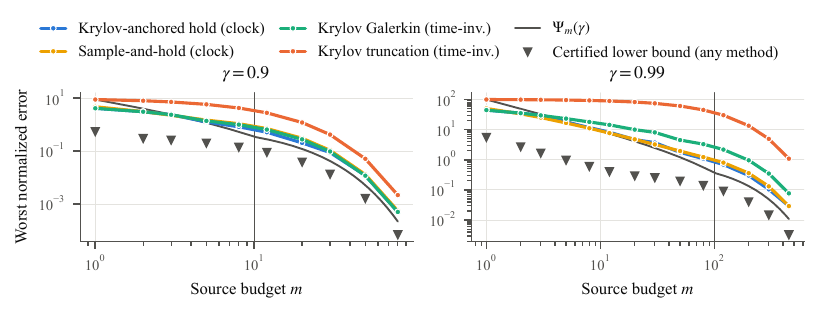}
\caption{\textbf{$\gamma$-uniform context--accuracy law.} Worst normalized all-depth error over structured ring instances, built from raw batches through attention heads, at $\gamma=0.9$ and $\gamma=0.99$. Triangles give a certified lower bound for any $m$-dimensional context; the vertical line marks $m=1/(1-\gamma)$. Sample-and-hold and the anchored hold follow $\Psi_m$, whereas Krylov truncation plateaus until the effective horizon. The right panel is reproduced in Figure~\ref{fig:context-budget}(c).}
\label{fig:rates}
\end{figure}

\subsection{The $\gamma$-Uniform Law (Figure~\ref{fig:rates})}
For $\gamma\in\{0.9,0.99\}$, the errors of sample-and-hold and the anchored hold stay within $[0.45,2.7]\times\Psi_m$ and within $21\times$ of the certified bound. Below the horizon at $\gamma=0.99$ ($2\le m\le50$), their log--log slopes are $-0.89$ and $-0.93$, against the predicted $-1$. Krylov truncation has slope $-0.14$ there and is up to $37\times$ worse, close to the predicted $1/(e(1-\gamma))\approx 36.8$. Krylov projection is up to $2.7\times$ worse, growing with $1/\sqrt{1-\gamma}$.
For each $(\gamma,m)$, the instance bank contains the following, all realized as covered raw batches with one transition per state (so $\widehat P$ is the ring and $\mu$ is uniform):
(i) Instance~I with $n=\lceil m+1/(1-\gamma^2)\rceil+2+\lfloor2/(1-\gamma)\rfloor$;
(ii) the constant-plus-spike ring of Proposition~\ref{prop:mm-suboptimal}(a) with $n=m+2$ and $\varepsilon=10^{-6}$;
(iii) rewarded-arc rings with $B=2m$ even blocks, $n=(2B+1)w+1$, and $w=\operatorname{round}(1/(c\,m(1-\gamma)))\ge2$ for $c\in\{2,4,8\}$.
Errors use depths up to $n+15/(1-\gamma)$. The certified lower bound is the largest, over the bank, of half the exact water-filling width of the orthogonal (block) increments, i.e., the smallest $\varepsilon$ with $\sum_t(1-\varepsilon^2/\|x_t\|^2)_+\le m$; this is valid for every method in $\mathfrak C_m$ by Lemma~\ref{lem:block-width}. Krylov truncation is evaluated through its exact identity with TD stopped after $m$ steps. Galerkin uses the balanced two-pass Arnoldi encoder and the TD executor; sample-and-hold and the anchored hold use their encoders and the depth-counter executor.

\begin{figure}[h]
\centering
\includegraphics[width=\linewidth]{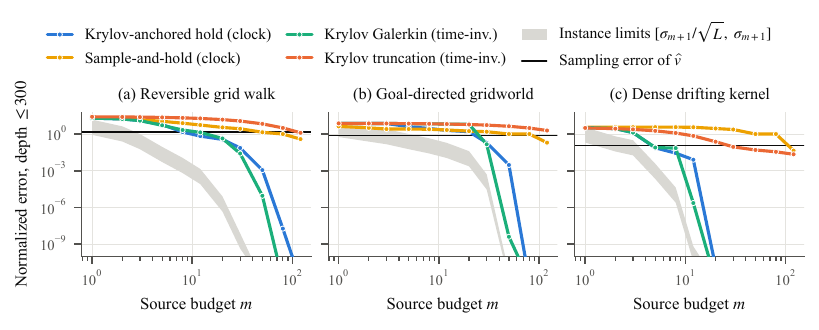}
\caption{\textbf{Sampled noisy tasks.} Mean normalized error over five raw trajectories versus the context dimension. The horizontal line is the sampling error of the empirical trajectory $\widehat v$ relative to the population TD trajectory; the band is the numerical-rank window of Proposition~\ref{prop:instance}(b).}
\label{fig:sampled}
\end{figure}

\subsection{Sampled Noisy Tasks (Figure~\ref{fig:sampled})}
\label{app:exp-accuracy}
Figure~\ref{fig:sampled} uses single trajectories with Gaussian reward noise ($\sigma=0.5$), $\gamma=0.99$, depths up to $300$, and five seeds. On all three tasks the exact rank is generically $k_L=n$, so exact compression gives nothing, but the $\varepsilon$-rank is small. On the two spectrally benign tasks the anchored hold stays within twice the projection error at context dimension $\lceil3m/4\rceil$, as Theorem~\ref{thm:anchored} predicts; on the goal task, where projection stalls, it follows sample-and-hold; neither pure method does both. It falls below the statistical error of $\widehat v$ (horizontal line) with $m=12$, $30$, and $5$, against $n=144$, $126$, and $200$. The shaded band, from Proposition~\ref{prop:instance}(b), bounds what any $m$-dimensional context can achieve on each task.

\emph{Reversible grid walk:} a $12\times12$ grid with uniform moves, where blocked moves stay in place (symmetric $P$). The mean reward is the difference of two Gaussian bumps centered at relative positions $(0.25,0.3)$ and $(0.7,0.75)$, with widths $0.12a$ and $0.15a$ and weights $1$ and $-0.6$.
\emph{Goal-directed gridworld:} a $12\times12$ grid with two interior walls, each with one gap ($n=126$ free cells). The policy is $\epsilon$-greedy ($\epsilon=0.2$) toward the goal along BFS shortest paths; the goal resets to the start cell. The reward is $-0.02$ per step and $+1$ at the goal. The chain is nonreversible.
\emph{Dense drifting kernel:} $n=200$ points on a circle; the transition from $x$ follows a Gaussian kernel of width $0.08$ centered at $x+0.05$ (dense and nonreversible); the mean reward is $\sin2\pi x+0.5\cos6\pi x$.
Each run samples a single trajectory of length $N$ ($10^5$, $10^5$, $2\times10^5$) with i.i.d.\ Gaussian reward noise of standard deviation $0.5$ (seeds 1000--1004); full coverage held in every run. With $\gamma=0.99$ and $L=300$, the error is $\max_{\ell\le300}\|\widehat v_\ell-\widetilde v_\ell\|_w/\|\widehat r\|_w$ with visit weights $w$. The sampling error is the same functional of $\widehat v_\ell-v^{\rm pop}_\ell$, where $v^{\rm pop}$ is the TD trajectory of the population model. The band plots $[\sigma_{m+1}/\sqrt L,\sigma_{m+1}]$ of the normalized $V_L$. In these runs the exact rank is $k_L=n$, while the number of singular values of the normalized $V_L$ above $10^{-2}$ is only $9$, $25$, and $5$.

\subsection{Memory Accounting (Table~\ref{tab:memory})}
\label{app:exp-memory}
\begin{table}[h]
\caption{\textbf{Retained memory (KiB, float32) on exact and sampled tasks.} Exact rows: dense-block cycles, $\gamma=0.9$, $L=100$, $N=40n$; LB is $nk_L+k_L$ (Theorem~\ref{thm:main}(c)). Approximate rows: $\gamma=0.99$, $L=300$, smallest context dimension with error $\le10^{-2}$ (Krylov projection / anchored hold). Bold marks the smallest entry in each row.}
\label{tab:memory}
\centering\small\setlength{\tabcolsep}{3.2pt}
\begin{tabular*}{\linewidth}{@{\extracolsep{\fill}}lrrrrrrr@{}}
\toprule
Task & $n$ & Raw & Statistics & Sparse TD & Answers & Ours & LB\\
\midrule
Block cycle, exact & 24 & 11.44 & 2.44 & \textbf{0.75} & 9.38 & 2.49 & 1.17\\
Block cycle, exact & 240 & 114.4 & 226.9 & 36.2 & 93.8 & \textbf{24.4} & 11.3\\
Block cycle, exact & 480 & 228.8 & 903.8 & 102.2 & 187.5 & \textbf{48.8} & 22.5\\
Dense kernel, $\epsilon=10^{-2}$ & 200 & 2345 & 157.8 & 139.0 & 234.4 & \textbf{15.7} / 17.2 & --\\
Goal gridworld, $\epsilon=10^{-2}$ & 126 & 1173 & 63.0 & \textbf{5.7} & 147.7 & 38.5 / 44.5 & --\\
\bottomrule
\end{tabular*}
\end{table}
Table~\ref{tab:memory} extends Table~\ref{tab:exp-memory-sweep} with an answer table storing $\widehat v_1,\ldots,\widehat v_L$ ($nL$ values) and with the sampled tasks of Figure~\ref{fig:sampled}. On the dense kernel, contexts accurate to $10^{-2}$, well below the statistical error of $0.22$, are $9\times$ smaller than sparse TD and $150\times$ smaller than the raw batch; on the sparse goal-directed gridworld, sparse TD remains the most compact representation.
Storage counts float32 values and int32 indices of the retained representation only, excluding shared parameters. Raw TD keeps records (12 bytes each), visit counts, the current value vector, and a query: $12N+8n+4$ bytes. Statistics keep $\widehat A$, $\widehat r$, and the value vector: $4(n^2+2n)$ bytes. Sparse TD keeps CSR row pointers, column indices and values, reward, value, and an output buffer: $4(n+1)+8\,\mathrm{nnz}+12n$ bytes. Answers keep $\widehat v_1,\ldots,\widehat v_L$: $4nL$ bytes. Krylov keeps $4(2n+1)(k+1)$ bytes, and the anchored hold keeps $4[(2n+1)(k+1)+(n+1)j]$ bytes.
Dense-block cycles have 12 blocks of size $n/12$; each record's successor is uniform in the next block; the reward is $1$ on block $0$; there are $N=40n$ records, and state labels are permuted (seeds $54000+n$). Row sums into the next block are exactly one and rewards are block-constant, so $k_L=12$ algebraically; the attention encoder retains exactly 12 tokens with error $0$ relative to TD. Approximate rows use seed 2000 and search $m\in\{1,\ldots,39\}\cup\{40,45,\ldots\}$ for the first budget with error at most $10^{-2}$.

\paragraph{Peak memory.}
The Peak column of Table~\ref{tab:exp-memory-sweep} is the peak task memory of Definition~\ref{def:memory}, the maximum over both encoding stages and execution; Table~\ref{tab:exp-peak} lists the three terms. The two encoding peaks are measured in the prototype as the live fields of all token groups plus the working storage of the heads of the current layer (every pending output and the largest score block, an upper bound on what is live at any instant). Stage~1 streams the raw records in chunks of $C=2n$ records, so its live storage is the $O(n^2)$ state tokens, one chunk of $C(2n+2)$ scalars, and the $C\times n$ score block and $n\times(n+2)$ outputs of the three heads; Stage~2 holds the state tokens, the $h=\min\{n,L\}$ candidate tokens, and the scores and outputs of its heads. The execution term is the analytic budget of one query, the retained context plus the $k_L(k_L+1)$ source--source scores; the prototype executor instead answers all $n$ coordinate queries at once and keeps the whole trajectory for the error metric, which is not what Definition~\ref{def:memory} counts. All three terms are $\Theta(n^2)$, the same order as the statistics, and the measured total is $12$--$16\times$ the statistics because the prototype stores five width-$n$ blocks per state token (identity, counts, empirical and discounted rows, and visit weights), fixed-capacity candidate arrays, and the full head outputs; a dense numerical implementation would keep two blocks. The benefit claimed is the retained context, i.e., what must persist to answer all later state and depth queries, not peak memory or speed.
\begin{table}[h]
\caption{\textbf{Retained and peak live storage} (KiB, float32) of the exact dense-block runs, with Stage~1 streamed in chunks of $2n$ records; ``chunk'' is the part of the Stage-1 peak occupied by one chunk of one-hot records, and both stage peaks include the heads' working storage. Execution is the analytic budget of one query. Total is the peak task memory of Definition~\ref{def:memory}.}
\label{tab:exp-peak}
\centering\small\setlength{\tabcolsep}{4pt}
\begin{tabular*}{\linewidth}{@{\extracolsep{\fill}}rrrrrrrr@{}}
\toprule
$n$ & Retained & Statistics & Peak, Stage 1 & \quad of which chunk & Peak, Stage 2 & Peak, execution & Total\\
\midrule
24 & 2.49 & 2.44 & 28.1 & 9.4 & 37.5 & 3.1 & 37.5\\
96 & 9.80 & 36.75 & 436.5 & 145.5 & 582.0 & 10.4 & 582.0\\
240 & 24.43 & 226.88 & 2711.3 & 903.8 & 1947.2 & 25.0 & 2711.3\\
480 & 48.80 & 903.75 & 10822.5 & 3607.5 & 5985.0 & 49.4 & 10822.5\\
\bottomrule
\end{tabular*}
\end{table}

\paragraph{Precision.}
The identity checks of Appendix~\ref{app:exp-checks} and all curves are computed in float64, whereas the storage in Table~\ref{tab:exp-memory-sweep} counts float32 values. Table~\ref{tab:exp-precision} reports the same tasks under one data type at a time. Storing the float64 context in float32 and executing in float32 reproduces the TD trajectory to between $2.0\times10^{-7}$ and $6.2\times10^{-7}$ of its largest entry, i.e., to float32 precision; running the whole pipeline in float32 (statistics heads, Krylov encoder with rank tolerance $10^{-5}$, executor, and a reciprocal that keeps the float32 type, so that no float64 intermediate is used) retains $k_L=12$ tokens on all four sizes and is accurate to between $2.0\times10^{-7}$ and $4.5\times10^{-7}$; with a different numerical linear-algebra implementation, the float32 results remain within the same order of magnitude. In float64 every storage count doubles, so the savings relative to dense statistics are unchanged, and the reproduction is accurate to between $1.6\times10^{-16}$ and $2.9\times10^{-15}$.
\begin{table}[h]
\caption{\textbf{Memory and TD error under one data type.} Retained memory of the Krylov context and total peak in float32 and float64 (KiB), and the maximum absolute error of the executed trajectory relative to $\max_{\ell,s}|\widehat v_\ell(s)|$ for the float64 pipeline, for the float64 context stored and executed in float32, and for the whole pipeline in float32.}
\label{tab:exp-precision}
\centering\small\setlength{\tabcolsep}{3pt}
\begin{tabular*}{\linewidth}{@{\extracolsep{\fill}}rrrrrrrr@{}}
\toprule
& \multicolumn{2}{c}{Retained} & \multicolumn{2}{c}{Total peak} & \multicolumn{3}{c}{Max normalized error}\\
\cmidrule(lr){2-3}\cmidrule(lr){4-5}\cmidrule(lr){6-8}
$n$ & float32 & float64 & float32 & float64 & float64 & float32 context & float32 pipeline\\
\midrule
24 & 2.49 & 4.98 & 37.5 & 75.0 & $1.6\times10^{-16}$ & $2.0\times10^{-7}$ & $2.0\times10^{-7}$\\
96 & 9.80 & 19.60 & 582.0 & 1164.0 & $1.7\times10^{-15}$ & $2.4\times10^{-7}$ & $2.7\times10^{-7}$\\
240 & 24.43 & 48.85 & 2711.3 & 5422.5 & $9.6\times10^{-16}$ & $6.2\times10^{-7}$ & $4.5\times10^{-7}$\\
480 & 48.80 & 97.60 & 10822.5 & 21645.0 & $2.9\times10^{-15}$ & $3.3\times10^{-7}$ & $4.2\times10^{-7}$\\
\bottomrule
\end{tabular*}
\end{table}

\paragraph{The lower-bound column.}
The column LB in Table~\ref{tab:memory} is the bound $nk_L+k_L$ of Theorem~\ref{thm:main}(c), which holds for every encoder that is continuous on a relatively open set of strictly positive rank-$k_L$ models (Theorem~\ref{thm:memory-law}). It is a bound on what such an encoder must retain on \emph{some} model of that class, not on what the block-cycle tasks themselves require: on these tasks every row of $\widehat P$ moves to the next block and the reward is block-constant, so $\widehat v_\ell=\sum_{j<\ell}\gamma^j\mathbf 1_{B_{-j}}$ depends on the model only through the block partition, $\mathbf 1\in\mathcal K_L$, and $\widehat P$ has zero entries; an encoder built for this family alone would retain the $n$ block labels and nothing else (Appendix~\ref{app:memory-proof}, scope of the bound). The comparison in Table~\ref{tab:memory} is therefore between the retained memory of the task-uniform encoder of Theorem~\ref{thm:main}(b), which depends on the task only through $(n,k_L)$, and the smallest memory that any continuous encoder can achieve uniformly over rank-$k_L$ dense models. The block-cycle tasks serve to instantiate $k_L=12$ exactly; the ratio $2.2\times$ is a property of the encoder at rank $12$, not a certificate of instance-level optimality on these tasks.

\subsection{Learnability Experiments}
\label{app:exp-learning}
Figure~\ref{fig:exp-parameters} in Section~\ref{sec:experiments} shows the learned parameter matrices for the three supervision signals. Here we specify the task distribution, training objectives, optimization settings, and results across initializations.

\paragraph{Tasks and initialization.}
Each of 2,400 training tasks has $n=24$ and eight observations per state ($N=192$). The family of each task is drawn uniformly from six (dense Dirichlet, sparse with both cycle neighbours and one further distinct successor, symmetric-weight, six-block, identical-row, and permutation; $410$, $394$, $405$, $410$, $384$, and $397$ tasks), with statewise rewards in $\{-1,-\frac34,\ldots,1\}$ and $\gamma\in\{0.5,0.7,0.8,0.9\}$; every generated kernel is verified to be row-stochastic, and 120 tasks are held out. The operator Gram matrix $\mathbb E_\tau[\widehat A_\tau^\top\widehat A_\tau]$ of the training bank has eigenvalues in $[0.199,0.545]$, so Assumption~\ref{ass:richness} holds with $\kappa=0.199$. Five paired initializations with entries $\mathcal N(0,0.1^2/n)$ are used for every objective, without checkpoint selection. 

\paragraph{Training objectives.}
The executor is run on the state-level context of Stage~1 with $Q_{\rm eff}(C,W)$, so that $v^\theta_{\tau,\ell}=-C^\top\sum_{j<\ell}F_\tau(C,W)^j\widehat r_\tau$. The three objectives are (\ref{eq:probe}), whose expectation over the isotropic calibration input is taken in closed form, $2\mathcal J=\|C+I\|_F^2+\mathbb E_\tau\|(C+I)^\top+\widehat A_\tau(W-I)\|_F^2$ as in the proof of Theorem~\ref{thm:learn}, so that gradient descent is the exact deterministic iteration of the theorem (the closed form is checked against direct evaluation of (\ref{eq:probe}) on random non-symmetric parameters); $\mathcal J_{\rm traj}$, normalized as $\frac1L\sum_{\ell\le L}\|v^\theta_{\tau,\ell}-\widehat v_{\tau,\ell}\|^2/\|\widehat r_\tau\|^2$ with $L=30$; and $\mathcal J^{(30)}_{\rm end}:=\mathbb E_\tau\|v^\theta_{\tau,30}-\operatorname{sg}(\widehat r_\tau+\widehat A_\tau v^\theta_{\tau,30})\|^2/\|\widehat r_\tau\|^2$, the depth-30 bootstrapped surrogate of $\mathcal J_{\rm end}$ with a stopped-gradient target (the infinite-depth $\mathcal J_{\rm end}$ of Section~\ref{sec:learnability} is not computable at finite depth, so the experiment tests the surrogate that a practitioner would minimize). 

\paragraph{Optimization and evaluation.}
Full-batch gradient descent on $\mathcal J/n$ uses step size $4$, below the certified $n/\lambda_+\approx9.17$, for 3,000 steps. Adam \citep{kingma2015adam} uses minibatches of 128 tasks, learning rate $10^{-2}$ with cosine decay to $10^{-4}$ over 6,000 steps, and gradient clipping at 10, for all three objectives. We report the normalized parameter error $\sqrt{(\|C+I\|_F^2+\|W-I\|_F^2)/(2n)}$ (equal to $1$ at initialization) and the held-out error $\max_{\ell\le30}\|v^\theta_{\tau,\ell}-\widehat v_{\tau,\ell}\|/\|\widehat r_\tau\|$ averaged over the 120 held-out tasks.

\paragraph{Results.} Gradient descent on (\ref{eq:probe}) reaches parameter error $1.0\times10^{-15}$ on all five seeds (held-out error $6.8\times10^{-15}$), at the geometric rate of Theorem~\ref{thm:learn}; Adam on the same objective reaches $3.2\times10^{-16}$. 

Adam on $\mathcal J_{\rm traj}$ reaches $7.4\times10^{-2}$ (range $6.5$--$8.4\times10^{-2}$ over seeds; held-out error $7.6\times10^{-2}$), with the diagonals of $C$ and $W$ at $-0.992$ and $0.992$ and off-diagonal entries of root-mean-square size $1.4\times10^{-2}$; continuing the same schedule to 20,000 steps on three seeds lowers the parameter error to $3.3\times10^{-3}$. The slow tail is a conditioning effect, not a failure of identification: along the scale family $(-aI,aI)$ the loss $\mathcal J_{\rm traj}$ is $0.15$ at $a=0.5$ and $5\times10^{-3}$ at $a=0.9$, a shallow valley, whereas leaving the family changes the fixed point and is penalized by every deep term, so first-order methods progress slowly along the valley; the isotropic objective (\ref{eq:probe}) has no such valley, which is why it carries the global guarantee. 

Adam on $\mathcal J^{(30)}_{\rm end}$ stops at parameter error $0.51$ (range $0.50$--$0.54$), with diagonals at $-0.81$ and $0.82$ and off-diagonal entries of size $1.0\times10^{-1}$. This is the degeneracy of Proposition~\ref{prop:identify} in action: on held-out tasks the depth-30 output is within $0.038$ of TD and its bootstrap residual is $0.011$ (TD's own residual is $0.004$), while the trajectory is wrong, with errors $0.50$, $0.42$, $0.29$, $0.17$ at depths $1$, $2$, $5$, $10$ and a maximum of $0.57$ ($0.63$ averaged over seeds); the endpoint objective fixes the fixed point of the learned update, not the update.

\subsection{Finite-Precision Rank Boundary}
\label{app:exp-stability}
On the two-state swap of Appendix~\ref{app:numerical-margin} with rewards $(1,1+\varepsilon)$, $\gamma=0.9$, and $L=10$, we compare the literal unbalanced one-pass float64 encoder (L) with the balanced two-pass encoder with tolerance $10^{-12}$ (B).
\begin{table}[h]
\caption{\textbf{Rank-boundary sensitivity.} Retained rank (L/B), largest context entry, and maximum all-state, all-depth error. At $\varepsilon=10^{-16}$ the float64 input already equals one.}
\label{tab:exp-stability}
\centering\small
\begin{tabular*}{\linewidth}{@{\extracolsep{\fill}}rrrrrr@{}}
\toprule
$\varepsilon$ & Rank L/B & Prompt L & Prompt B & Error L & Error B\\
\midrule
$0$ & 1/1 & $1.0$ & $2.0$ & $3.6\times10^{-15}$ & $0$\\
$10^{-4}$ & 2/2 & $1.1\times10^{4}$ & $2.0$ & $2.0\times10^{-12}$ & $3.6\times10^{-15}$\\
$10^{-8}$ & 2/2 & $1.1\times10^{8}$ & $2.0$ & $1.5\times10^{-8}$ & $8.9\times10^{-16}$\\
$10^{-12}$ & 2/1 & $1.1\times10^{12}$ & $2.0$ & $1.5\times10^{-12}$ & $3.1\times10^{-12}$\\
$10^{-16}$ & 1/1 & $1.0$ & $2.0$ & $3.6\times10^{-15}$ & $0$\\
\bottomrule
\end{tabular*}
\end{table}
Unbalanced duals grow like $1/\varepsilon$, whereas balancing keeps context entries bounded. The tolerance then suppresses a genuinely nonzero but negligible direction, which is exactly the behavior that the $\varepsilon$-rank of Proposition~\ref{prop:instance}(b) makes principled.

\subsection{Feature-Space Policy Evaluation (Table~\ref{tab:exp-memory-sweep})}
\label{app:exp-features}
The bottom rows of Table~\ref{tab:exp-memory-sweep} run the un-normalized TD update (\ref{eq:lfa-td}) of Theorem~\ref{thm:features} on single sampled trajectories. \emph{Boyan's chain} \citep{boyan2002lstd}: 13 states, four hat features, reward $-3$ per step ($-2$ on the last transition), $\gamma=0.95$, one trajectory of $N=20{,}000$ steps, $L=100$. \emph{Mountain car}: the standard dynamics with a fixed energy-pumping policy ($a=\operatorname{sign}(\dot x)$, $\epsilon=0.1$ random actions), reward $-1$ per step, termination at $x\ge0.5$ with zero successor features, $\gamma=0.99$, $L=200$; features are eight $8\times8$ tilings of the position--velocity box with random offsets ($d=512$, eight active tiles per state), of which $417$--$452$ occur in the trajectories of $N\in\{5{,}000,20{,}000,80{,}000,320{,}000\}$ steps. Stage~1 forms $A_N$ and $b_N$ on all $512$ coordinates by the two heads of Appendix~\ref{app:features}, streamed in chunks of $2d$ records; coordinates that never occur have zero rows and never move, and Stage~2 runs on the active ones. The step size is a public input chosen before encoding from the spectrum of $A_N$, $\alpha=c/\lambda_{\max}(\operatorname{sym}A_N)$ with the largest $c\in\{0.9,0.45,\ldots\}$ for which $|1-\alpha\lambda|<1$ for every nonzero eigenvalue $\lambda$ of $A_N$; this choice is made by the experimenter, not by the selector-head network, which receives $\alpha$ as a constant. Eigenvalue zero arises from the tilings, whose indicators sum to one, and is harmless because $b_N$ has no component along the corresponding left null vectors. Errors are $\max_{\ell\le L}\|\Phi(\theta_\ell-\widetilde\theta_\ell)\|_w/\max_\ell\|\Phi\theta_\ell\|_w$ over the visited states with uniform weights, i.e., relative to the scale of the trajectory. On Boyan's chain the exact context has $k^\phi_L=4=d$ and reproduces the trajectory to $9.4\times10^{-16}$; Krylov projection with $m=1,2,3$ has error $0.95$, $0.34$, $0.14$. On mountain car the exact encoder retains $h=\min\{d,L\}=200$ tokens at tolerance $10^{-12}$, but the trajectory matrix has at most $10$ singular values above $10^{-2}\sqrt L$, and Krylov projection reaches relative error $9.1\times10^{-3}$--$1.06\times10^{-2}$ at $m=32$ (the largest value at $N=5{,}000$), at most $3.0\times10^{-7}$ at $m=48$, and at most $6\times10^{-15}$ at $m=64$ over the four values of $N$. Sample-and-hold uses the finite-depth schedule of Proposition~\ref{prop:finite-depth}(b) with the measured contraction factor $\rho=\max_\lambda|1-\alpha\lambda|\in[0.99998,0.99999]$ of $I-\alpha A_N$; since $\lfloor1/(1-\rho)\rfloor>L$, all $m$ tokens are hold tokens at the quantiles of the mass through depth $L$, which here are nearly uniformly spaced, and the encoder keeps exactly the requested number of tokens up to $m=192$ and $200$ tokens for $m\ge200$. Its relative error is $0.31$--$0.33$ at $m=32$, $0.19$--$0.21$ at $m=64$, $0.07$--$0.08$ at $m=128$ and $m=192$, and below $10^{-15}$ once every depth is sampled (requested $m\ge200$, $200$ tokens): the increments do not decay within $L=200$, so holding misses one full increment wherever two consecutive depths are not both sampled, the short-horizon regime of Proposition~\ref{prop:finite-depth}(d). The retained memory of the dimension-$32$ context is $108$--$117$~KiB against $681$--$800$~KiB for the LSTD statistics $A_N,b_N$, and it does not depend on $N$, whereas the sparse raw prompt grows from $332$~KiB to $21$~MiB. The peak column of Table~\ref{tab:exp-memory-sweep} is measured as in Appendix~\ref{app:exp-memory}, including the heads' working storage: on mountain car it is the Stage-1 peak on all $512$ coordinates ($9{,}226$~KiB, of which one chunk of $2d$ raw records is $4{,}100$~KiB), and on Boyan's chain the Stage-2 peak ($1.1$~KiB).

\begin{figure}[h]
\centering
\includegraphics[width=\linewidth]{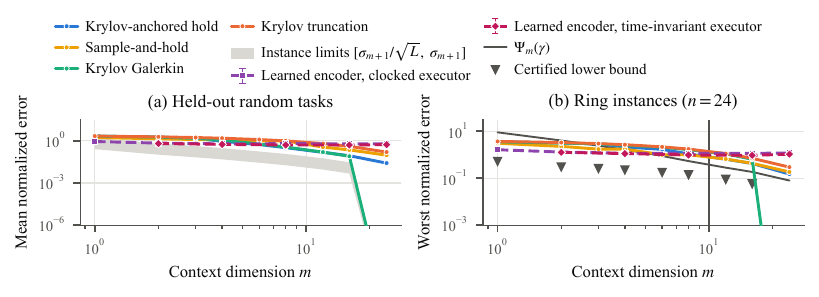}
\caption{\textbf{Learned bottleneck encoders against compiled contexts.} (a) Mean normalized error on 256 held-out tasks and (b) worst error over three ring instances, for standard Transformer encoders with a clocked or time-invariant executor (three seeds per point, 8,000 updates each; error bars are one standard deviation across seeds), the compiled contexts on the same batches, $\Psi_m$, and the certified lower bound for every $m$-dimensional context. The vertical line marks $m=1/(1-\gamma)$.}
\label{fig:learned}
\end{figure}

\subsection{Learned Bottleneck Contexts (Figure~\ref{fig:learned})}
\label{app:exp-learned}
\paragraph{Model.} The input is the output of the fixed first stage of Section~\ref{sec:exact-context}, the iterates $\widehat v_1,\ldots,\widehat v_L$. During training, they are computed by direct counting of the empirical statistics followed by the TD recurrence, which agrees with the selector-head implementation to $6.3\times10^{-15}$ on the training tasks; the heads themselves are not executed during training. The encoder reads the $L$ tokens $(\widehat v_\ell,\ell/L)\in\mathbb R^{n+1}$. They are embedded to width $d=64$ and processed by two standard pre-LayerNorm Transformer encoder layers (four softmax heads, GELU MLP of width $128$, full attention over the $L$ tokens). $m$ learned latent vectors then cross-attend to the encoded records in two rounds, each followed by self-attention among the latents and an MLP, as in Perceiver-style bottlenecks \citep{jaegle2021perceiver}; a final LayerNorm gives $z_u\in\mathbb R^d$ for $u\in[m]$. The latents are turned into a context for the executors of Definition~\ref{def:executors}. A linear map sends $z_u$ to the decoder column $d_u\in\mathbb R^n$, giving $D\in\mathbb R^{n\times m}$. In the \emph{clocked} model the source scalars are depth indexed, $s_{\ell,u}=\operatorname{MLP}(z_u,e(\ell))$ with a sinusoidal depth code $e(\ell)$, which generalizes the gates $\mathbf 1\{\ell+1=\tau_u\}$ of the depth-counter executor. In the \emph{time-invariant} model a linear readout gives $s_0\in\mathbb R^m$, a bilinear form of the latents gives $K\in\mathbb R^{m\times m}$, rescaled so that $\|K\|_2\le\gamma$ (a numerically computed spectral norm, used only in this learned read-out), and $s_{\ell+1}=Ks_\ell$, which is the recurrence of the time-invariant executor with $(D,K,s_0)$ produced by the encoder. In both cases the output is $\widetilde v_\ell=D\sum_{j<\ell}s_j$, so the model lies in $\mathfrak C_m$ and the lower bounds of Section~\ref{sec:approx-context} apply. Nothing is hand-set; each model has about $1.9\times10^5$ parameters. Two other interfaces were tried with the same encoder and read-out and are reported below: raw transition tokens $(e_{S_t},e_{S_{t+1}},R_{t+1})$, and the $n$ state-statistics tokens $(e_i,\widehat A_{i,:},\widehat r_i)$. A pilot with a looped standard Transformer block as the coefficient generator, trained through $L=40$ loops from raw records, did not leave the plateau either (error $1.9$ for every $m$, against $2.6$ for the zero predictor and $2.1$ for the best rank-one context along $\widehat r$), which is why the executor classes are used as the read-out.

\paragraph{Tasks and training.} $n=24$, $\gamma=0.9$, $L=40$, $N=192$ records per task with every state visited at least once; the empirical model is obtained by direct counting, and its trajectory is computed by the TD recurrence; Appendix~\ref{app:exp-checks} verifies these calculations against the selector-head implementation. Tasks are drawn uniformly from six families: dense Dirichlet rows, sparse rows with two to four successors, symmetric-weight (reversible) chains, four-block cycles, single $n$-cycles, and random permutations; rewards are Gaussian on all states or on one to three states, scaled to $\|\widehat r\|_\infty=1$. The target is the empirical trajectory $\widehat v_1,\ldots,\widehat v_L$ of the batch, and the loss is $\frac1L\sum_\ell\|\widehat v_\ell-\widetilde v_\ell\|_w^2/\|\widehat r\|_w^2$ with visit weights $w$. One model is trained per $(m,\text{executor type},\text{seed})$ with AdamW (learning rate $10^{-3}$, cosine decay to $10^{-5}$ after $200$ warm-up steps, weight decay $0.01$, gradient clipping at $1$) for $8{,}000$ steps of $32$ fresh tasks each, on a CPU in float32 with two threads per run, with three seeds ($0$, $1$, $2$) for every point of the grid $m\in\{1,2,4,8,16,24\}$ (clocked) and $m\in\{2,4,8,16,24\}$ (time-invariant); an earlier single-seed sweep used $4{,}000$ steps. No checkpoint or seed selection is used. Evaluation uses the metric of Section~\ref{sec:experiments}, $\max_{\ell\le L}\|\widehat v_\ell-\widetilde v_\ell\|_w/\|\widehat r\|_w$, on $256$ held-out tasks from the same distribution (seed $777$) and on three deterministic ring instances of size $n=24$ realized as covered batches: the single-reward ring and rewarded arcs of width $2$ and $3$ (Appendix~\ref{app:gamma-law-proof}). The certified lower bound in Figure~\ref{fig:learned}(b) is the water-filling width of the orthogonal block increments of these instances (Lemmas~\ref{lem:block-width} and~\ref{lem:waterfill}), valid for every context in $\mathfrak C_m$. The compiled contexts are evaluated on exactly the same batches through the attention implementation.

\paragraph{Results.} Table~\ref{tab:learned} lists the numbers behind Figure~\ref{fig:learned} as means and standard deviations over the three seeds. Four facts stand out. First, below the effective horizon the learned encoders are better than the compiled contexts on typical tasks: for $m\le8$ their held-out error ($0.50$--$0.89$) is below sample-and-hold ($1.00$--$1.72$), and for $m\le4$ below the anchored hold ($0.93$--$2.17$); on the rings they are below sample-and-hold for $m\le4$ and far below the worst-case guarantee $e^2\Psi_m$. The compiled contexts are tuned to the worst case and spend their first tokens on the earliest increments, which a learned encoder need not do. Second, the learned error does not decrease with $m$: on held-out tasks it stays at $0.61$--$0.63$ for every $m\ge4$ with the clocked executor and at $0.50$--$0.54$ for every $m\ge8$ with the time-invariant one, and on the rings at $1.14$--$1.19$ and $0.95$--$1.05$, with log--log slopes over $2\le m\le8$ of $-0.07$ and $-0.18$ on held-out tasks and $-0.09$ and $-0.19$ on the rings, against $-0.59$ for sample-and-hold and $-1.46$ for $\Psi_m$. From $m=8$ on the ring error exceeds $\Psi_m$, and at $m=24$ it exceeds the upper bound $e^2\Psi_m$ by $2.0\times$ (clocked) and $1.8\times$ (time-invariant): a learned encoder with the full budget is worse than the sample-and-hold guarantee, while the compiled contexts reach $0.19$ (SH) and $0.15$ (anchored) and the held-out error of the anchored hold is $0.026$. Third, the seed spread is small compared with these gaps: the standard deviations are $0.003$--$0.06$ on held-out tasks ($0.11$ at $m=2$ for the time-invariant model) and $0.01$--$0.11$ on the rings, so the flatness in $m$ is not a seed effect. Fourth, every learned error lies above the certified lower bound, by $3.1\times$ at $m=1$ and $16$--$19\times$ at $m=16$, as Lemma~\ref{lem:block-width} requires of any context in $\mathfrak C_m$. Per family, the learned encoders reach $0.35$--$0.55$ on the dense, reversible, and block tasks, $0.53$--$0.64$ on the sparse ones, and $0.81$--$0.94$ on the deterministic cycles and permutations, at every $m$; the compiled contexts are worse than this on the deterministic families at $m\le4$, whereas by $m=24$ the anchored hold is exact on the dense, reversible, block, and sparse families and reaches $0.03$--$0.11$ on the deterministic ones. The time-invariant executor is better than the clocked one on held-out tasks for $m\ge4$ ($0.50$--$0.56$ against $0.61$--$0.63$) and on the rings; with $(K,s_0)$ read off the latents it is a learned counterpart of the Krylov context, and its error is also flat in $m$. Doubling the training budget from the earlier $4{,}000$-step single-seed sweep (held-out errors $0.74$--$0.77$ for the clocked and $0.67$--$0.71$ for the time-invariant models at $m\ge4$) lowered every curve by $0.1$--$0.2$ but did not change its shape, and at the end of training the held-out error was still decreasing by at most $0.02$ per $1{,}000$ steps, so a larger budget would lower the curves further but, on this evidence, not make them decrease with $m$. The two other interfaces of the $4{,}000$-step pilots give $1.91$ (raw records) and $1.82$ (statistics tokens) at $m=8$ with the clocked executor, against $0.75$ from the iterates; both are close to the zero predictor ($2.58$) and to the best rank-one context along $\widehat r$ ($2.09$), so they did not leave a rank-one plateau. We state the finding plainly: a standard Transformer with a bottleneck learns to place a few tokens better than the worst-case-optimal schedules do on typical tasks, but it does not discover the Krylov or clock structure that turns additional tokens into accuracy; supplying that structure is what the compiled encoders of Sections~\ref{sec:exact-context} and~\ref{sec:approx-context} contribute.

\begin{table}[h]
\caption{\textbf{Learned bottleneck encoders} (trajectory-token interface; mean $\pm$ one standard deviation over three seeds, $8{,}000$ updates each) against the compiled sample-and-hold (SH) and anchored contexts on the same batches, the upper bound $e^2\Psi_m$ of Theorem~\ref{thm:gamma-law}, and the certified lower bound (LB) of the ring instances. Dashes: budget not trained, or no certificate.}
\label{tab:learned}
\centering\small\setlength{\tabcolsep}{3pt}
\begin{tabular*}{\linewidth}{@{\extracolsep{\fill}}rccccccccc@{}}
\toprule
 & \multicolumn{4}{c}{Held-out tasks, mean error} & \multicolumn{5}{c}{Ring instances, worst error}\\
\cmidrule(lr){2-5}\cmidrule(lr){6-10}
$m$ & clocked & time-inv. & SH & anchored & clocked & time-inv. & SH & $e^2\Psi_m$ & LB\\
\midrule
1 & $0.89\pm0.01$ & -- & $1.72$ & $2.17$ & $1.64\pm0.01$ & -- & $3.08$ & $66.50$ & $0.53$\\
2 & $0.69\pm0.00$ & $0.64\pm0.11$ & $1.39$ & $1.85$ & $1.30\pm0.01$ & $1.26\pm0.02$ & $2.28$ & $29.93$ & $0.31$\\
4 & $0.61\pm0.03$ & $0.56\pm0.04$ & $1.23$ & $0.93$ & $1.16\pm0.03$ & $1.12\pm0.11$ & $1.73$ & $12.12$ & $0.23$\\
8 & $0.63\pm0.03$ & $0.51\pm0.00$ & $1.00$ & $0.33$ & $1.14\pm0.02$ & $0.98\pm0.04$ & $1.00$ & $3.98$ & $0.14$\\
16 & $0.62\pm0.06$ & $0.50\pm0.00$ & $0.23$ & $0.087$ & $1.15\pm0.06$ & $0.95\pm0.05$ & $0.43$ & $1.37$ & $0.060$\\
24 & $0.61\pm0.01$ & $0.54\pm0.05$ & $0.096$ & $0.026$ & $1.19\pm0.02$ & $1.05\pm0.09$ & $0.19$ & $0.59$ & --\\
\bottomrule
\end{tabular*}
\end{table}

\section{Scope and Limitations}
\label{app:scope}
We collect the qualifications here rather than repeating them in each proof.
\textbf{Computational model.} Exact statements use real arithmetic and declared primitives: $\operatorname{recip}_0$, token release, and, for certified adaptive rank, a threshold gate. Exact rank selection over arbitrary reals requires a discontinuous primitive (Appendix~\ref{app:numerical-margin}). Clocked contexts treat $\gamma$ as public, although in-network schedules are possible. Appendix~\ref{app:softmax} realizes both executors with standard two-head masked softmax attention; that realization is a compilation whose encoder writes log-probability codes into the tokens, and it says nothing about what trained softmax models do. The bottleneck encoders of Appendix~\ref{app:exp-learned} are the only trained encoders in this paper; they read the iterates of the fixed first stages and are trained on our task distribution, not on natural pretraining data.
\textbf{What is optimal.} Theorem~\ref{thm:memory-law} counts reals for encoders that are continuous in the model on dense model neighborhoods; sparse models can be stored more compactly. Theorem~\ref{thm:gamma-law} is a worst-case statement with absolute but unoptimized constants (measured gaps are within $20\times$). Proposition~\ref{prop:instance} supplies instance-dependent guarantees. Proposition~\ref{prop:mm-suboptimal} concerns moment-matching encoders, not all time-invariant recurrences. The depth bounds of Theorem~\ref{thm:depth} are tight up to the base of the logarithm ($\log_23$) and additive constants; the lower bound is for polynomial executors, which excludes $\operatorname{recip}_0$, so it constrains the doubling and time-invariant executors but not the depth-counter one, whose unrolled $\operatorname{recip}_0$-free variant has depth $L$.
\textbf{Resources not reduced.} Peak memory and arithmetic are not reduced below those of the statistics or direct TD; the benefit is the retained, reusable context. For sparse models, sparse numerical TD can be smaller.
\textbf{Learning.} Only the shared executor is shown to be learnable; joint training of encoders is open.
\textbf{Setting.} The results cover fixed-policy evaluation of empirical, visitation-normalized TD. In control, the Krylov space changes with the policy, and extending the laws there is open.
\end{document}